\pdfoutput=1
\documentclass{ieeeaccess}
\usepackage{cite}
\usepackage{amsmath,amssymb,amsfonts}
\usepackage{algorithmic}
\usepackage{graphicx}
\usepackage{textcomp}

\usepackage{arydshln}

\usepackage{url}
\usepackage{xurl}

\usepackage{float}      % Provides the [H] placement specifier
\usepackage{placeins}

\usepackage{makecell} % Para quebrar células
\usepackage{tabularx} % Para controle automático de largura

\usepackage{booktabs}

\usepackage{multirow}

\usepackage{threeparttablex}
\usepackage{ragged2e}

\usepackage{xspace}

\usepackage{xcolor}
\usepackage{soul}

\sethlcolor{yellow!50}

\providecommand{\HighlightChanges}{0}
\newif\ifhighlightchanges
\ifnum\HighlightChanges=1 \highlightchangestrue\fi
\soulregister{\cite}{7}\soulregister{\ref}{7}\soulregister{\eqref}{7}
\soulregister{\textit}{1}\soulregister{\textbf}{1}\soulregister{\emph}{1}\soulregister{\textdegree}{0}
\ifhighlightchanges
  \usepackage[switch,pagewise]{lineno}
  
  \linenumbers
\fi
\newcommand{\chg}[1]{\ifhighlightchanges\hl{#1}\else#1\fi}
\newcommand{\chgfig}[1]{\ifhighlightchanges{\setlength{\fboxsep}{2pt}\hspace*{-\fboxsep}\colorbox{yellow!50}{#1}\hspace*{-\fboxsep}}\else#1\fi}% moldura nao altera a largura

\usepackage{bm}
\makeatletter
\AtBeginDocument{\DeclareMathVersion{bold}
\SetSymbolFont{operators}{bold}{T1}{times}{b}{n}
\SetSymbolFont{NewLetters}{bold}{T1}{times}{b}{it}
\SetMathAlphabet{\mathrm}{bold}{T1}{times}{b}{n}
\SetMathAlphabet{\mathit}{bold}{T1}{times}{b}{it}
\SetMathAlphabet{\mathbf}{bold}{T1}{times}{b}{n}
\SetMathAlphabet{\mathtt}{bold}{OT1}{pcr}{b}{n}
\SetSymbolFont{symbols}{bold}{OMS}{cmsy}{b}{n}
\renewcommand\boldmath{\@nomath\boldmath\mathversion{bold}}}
\makeatother

\def\BibTeX{{\rm B\kern-.05em{\sc i\kern-.025em b}\kern-.08em
    T\kern-.1667em\lower.7ex\hbox{E}\kern-.125emX}}

\usepackage{eso-pic}
\AddToShipoutPictureFG{\AtPageUpperLeft{\put(0,-8){\makebox[\paperwidth][c]{\parbox[t]{\paperwidth}{\centering\fontencoding{OT1}\fontfamily{phv}\fontseries{m}\fontshape{n}\fontsize{6.5}{7.6}\selectfont This article has been accepted for publication in IEEE Access. This is the author's version which has not been fully edited and\\ content may change prior to final publication. Citation information: DOI 10.1109/ACCESS.2026.3754642}}}}}

\begin{document}
\history{Date of submission October 7, 2026, date of current version November 9, 2026.}
\doi{10.1109/ACCESS.2026.3754642}

\title{Injecting Hallucinations in Autonomous Vehicles: A Component-Agnostic Safety Evaluation Framework}

\author{\uppercase{Alexandre Moreira Nascimento}\authorrefmark{1,2,*},
\uppercase{Gabriel Kenji Godoy Shimanuki}\authorrefmark{1,*}, 
\uppercase{Lúcio Flavio Vismari}\authorrefmark{1},
\uppercase{João Batista Camargo Jr.}\authorrefmark{1},
\uppercase{Jorge Rady de Almeida Jr.}\authorrefmark{1},
\uppercase{Paulo Sergio Cugnasca}\authorrefmark{1},
\uppercase{Anna Carolina Muller Queiroz}\authorrefmark{3},
\uppercase{Jeremy Noah Bailenson}\authorrefmark{4}}

\address[1]{Escola Politécnica da Universidade de São Paulo, São Paulo 05508-010, Brazil}
\address[2]{Center for Design Research, Department of Mechanical Engineering, Stanford, CA 94305 USA}
\address[3]{Department of Communication, University of Miami, Coral Gables, FL 33124 USA}
\address[4]{Department of Communication, Stanford University, Stanford, CA 94305 USA}
\address[*]{Authors contributed equally to this work}

\tfootnote{The Article Processing Charge for the publication of this research was funded by Coordenação de Aperfeiçoamento de Pessoal de Nível Superior - Brasil (CAPES) - Finance Code 001 (ROR identifier: 00x0ma614). For the purpose of open access, the authors have applied a Creative Commons CC BY license to any accepted manuscript version. The work of Gabriel Kenji Godoy Shimanuki was supported by Itaú Unibanco S.A., through the Itaú Scholarships Program (PBI), linked to the Centro de Ciência de Dados (C2D) at the Escola Politécnica da Universidade de São Paulo.}

\markboth
{Nascimento \headeretal: Injecting Hallucinations in Autonomous Vehicles: A Component-Agnostic Safety Evaluation Framework}
{Nascimento \headeretal: Injecting Hallucinations in Autonomous Vehicles: A Component-Agnostic Safety Evaluation Framework}

\corresp{Corresponding author: Alexandre Moreira Nascimento (e-mail: alexandremoreiranascimento@alum.mit.edu).}

\begin{abstract}
Perception failures in autonomous vehicles (AV) remain a major safety concern because they can directly compromise decision-making and contribute to unsafe behavior. To study how these failures affect safety, researchers typically inject artificial faults into real-world or simulated components and observe the resulting outcomes. However, existing fault injection (FI) studies often target a single sensor within the machine perception (MP) module, resulting in siloed frameworks that are difficult to generalize or integrate into unified simulation environments. This work addresses that limitation by reframing perception failures as hallucinations, that is, false perceptions that distort an AV's situational awareness and may trigger unsafe control actions. Since hallucinations are defined in terms of observable effects, this abstraction supports analyses that depend less on specific sensors or algorithms and instead focuses on how faults appear in the information delivered from MP to AV control. In contrast to sensor/model-level FI and robustness testing, the approach operates at the level of perception-output effects, at the interface between MP and machine control, enabling component-agnostic, system-level safety evaluation. Building on this concept, we propose a configurable, component-agnostic hallucination injection framework that induces six plausible hallucination types in an open-source simulation environment. \chg{A total of} $18{,}356$ simulations \chg{($8{,}695$ baseline and $9{,}661$ with hallucinations injected)} were executed while AVs crossed an unsignalized intersection with transversal traffic. The results statistically validate the framework and quantify the impact of each hallucination type on collisions and \chg{on the minimum distance between the AV and the nearest vehicle, used as a continuous proxy for the safety margin}. Certain hallucinations, especially missed detection, blind region, angular drift, phantom perception, and latency, significantly increase \chg{the odds of collision}, while linear drift significantly reduces minimum distances without producing a statistically significant increase in collision odds in the evaluated scenario. The proposed framework provides a scalable and statistically grounded method for evaluating AV safety under perception distortions and supports future research on fault tolerance and resilient AV design.
\end{abstract}


\titlepgskip=-21pt

\maketitle

\section{Introduction}
\label{sec:introduction}
    \PARstart{T}{he} integration of artificial intelligence (AI) into autonomous vehicles (AVs) introduces unique challenges to safety assurance. Evaluating the safety of AI-based components is difficult because commonly used metrics tend to reflect aggregate performance while masking specific failure manifestations and their consequences \cite{ren2024safetywashing}. These limitations are especially critical in machine perception (MP) systems, which rely heavily on AI techniques and directly influence safety-critical decisions \cite{van2018autonomous, ma2020artificial}. MP interprets the driving environment and provides the situational awareness required for AV motion planning and control. Failures in this process can cause unsafe AV actions, with severe consequences, as evidenced by crashes, fatalities, and recalls reported by the National Transportation Safety Board (NTSB) \cite{NTSB_HAR1903} and the media\footnote{News reports cited in this study are used solely for motivational and illustrative purposes, from a societal perspective. They are not used as technical evidence.} \cite{tung2019fatal, ArizonaTesla2023, ZooxRecall2025}. These examples show that MP systems depend on robust safety assurance, particularly because fallback mechanisms such as human intervention often exhibit delayed responses and vigilance degradation \cite{mackworth1948breakdown, lichstein2000mackworth}.

    A persistent challenge in AV safety is the divide between the AI research community and the safety engineering community \cite{nascimento2018concerns}. Advances in deep learning have enabled state-of-the-art perception, navigation, and control \cite{nascimento2019systematic}, but these advances are often driven by benchmark performance rather than system-level safety objectives \cite{bengio2024managing}. Several researchers have noted that this AI-centric mindset focuses on local robustness and narrow metrics while neglecting application-level safety assurance \cite{muhammad2020deep, rismani2023plane, abrecht2024deep, mirzarazi2024safety, miller2024critical}. Even adversarial robustness research tends to isolate specific vulnerabilities, such as perturbations of images and videos \cite{tian2018deeptest, wang2020metamorphic, bolte2019towards, kumar2020black}, control flaws \cite{sun2021corner, koren2018adaptive}, or navigation corner cases \cite{song2023critical, shimanuki2025navigating, shimanuki2025cortex}, without modeling how these local faults propagate through the AV stack. This gap illustrates the need to address robustness at the system level, that is, the ability to continue operating correctly even when subsystems or components are faulty \cite{chen2025toward, koopman2017autonomous}.    
    
    Robustness is, therefore, a key attribute for safe AVs. Koopman \cite{koopman2018practical} highlights robustness testing for AI-based systems and advocates fault injection (FI) to evaluate performance under rare or unexpected conditions. More broadly, FI is a well-established dependability assessment technique in which faults are deliberately introduced into hardware or software systems, real or simulated, to reveal vulnerabilities and evaluate safety attributes \cite{carreira1999fault}. FI supports both fault removal and fault forecasting, the core goals of dependability validation \cite{avresky2002fault}. Classical FI approaches \cite{segall1995fiat, madeira1994rifle} focused on physical hardware or software components to expose vulnerabilities, test redundancy mechanisms, and evaluate performance under stress conditions \cite{lala1994architectural}. However, for meaningful results, the injected faults must reflect realistic system behavior \cite{avresky2002fault}.

    Simulation-based FI offers a practical and scalable approach to evaluate AV design and safety. By realistically modeling both vehicle components and their surrounding environments, it enables controlled anomalies to be introduced into specific modules, such as perception sensors or system interfaces, and allows their effects on system behavior to be systematically observed under repeatable conditions \cite{carreira1999fault}.

\begin{figure}[h]
        \centering
        \includegraphics[scale=0.39]{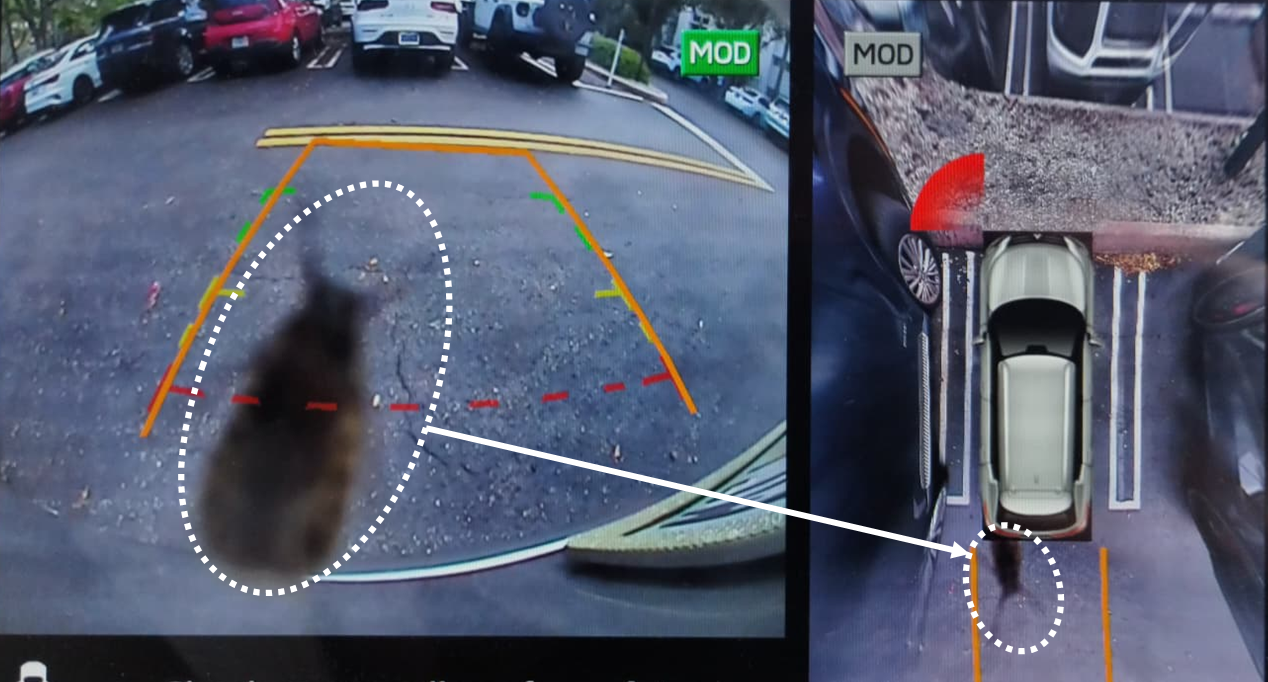}
        \caption{Hallucination produced by a \chg{360\textdegree}\ image reconstruction algorithm}
        \label{fig:real_hallucination}
    \end{figure}

    Despite advances in machine learning (ML) robustness research \cite{dey2021multilayered, wu2020testing}, few studies have investigated FI in AVs through controlled anomalies in the information provided by MP to \chg{machine control} (MC). To address this gap, this study introduces a configurable, component-agnostic hallucination injection (HI) framework that simulates six distinct types of perception-output anomalies, or hallucinations. Although the term lacks a universal definition, it has been widely associated with false but plausible outputs generated by large language and generative AI models \cite{maleki2024ai}. By analogy, in AV perception, a hallucination represents a misleading observation that the MP interprets as real. As an example of perceptual distortion (Figure~\ref{fig:real_hallucination}), a rear camera captured an unexpected obstruction, such as an insect landing on the lens. The MP software then reconstructed the \chg{360\textdegree}\ view with a giant fly apparently blocking the environment, producing a hallucination that misrepresents reality and could mislead AV motion planning. This example illustrates the importance of testing AVs against perceptual anomalies that may arise from real-world interactions, beyond traditional fault models. HI does not inject faults into the internal components of the MP system or model the full traditional $fault \rightarrow error \rightarrow failure$ chain \cite{avizienis2004basic} within the MP system itself. Instead, it shifts the evaluation focus to the MP-MC frontier, where corrupted perception outputs are delivered to decision-making modules. This enables standardized, component-agnostic testing across diverse AV architectures, regardless of the specific sensors or AI modules involved. The framework is integrated into an open-source traffic simulator, supporting interactive real-time testing at scale. We demonstrate its statistical validity through $18{,}356$ simulations in an unsignalized intersection scenario, revealing vulnerabilities that may be missed by conventional testing and offering new perspectives on AV safety under hallucinations.

    Existing FI and robustness evaluation approaches for ML commonly operate at the level of fault sources, sensor and model perturbations, or internal ML mechanisms, although some virtual-testing approaches also inject perception errors on MP outputs by modifying perceived world-state representations. While these methods are used for analyzing component-level vulnerabilities, they provide limited support for studying how diverse fault mechanisms manifest as distorted perception outputs and how these distortions propagate to planning and control decisions. In contrast, the HI framework proposed in this work operates at the perception--control interface, abstracting away MP fault sources and implementation details to evaluate the safety impact of perceptual distortions on AV behavior. This enables a component-agnostic, system-level safety evaluation that complements existing FI and robustness testing methods.

    The remainder of this paper is organized as follows. Section~\ref{sec:related-work} reviews related work on FI in AVs and on evaluating AV behavior under corrupted perception outputs delivered by MP systems. Section~\ref{sec:methodology} presents the design of the proposed HI framework, details the modeling of perception hallucination types, and explains the systematic evaluation methodology. Section~\ref{sec:results} reports the experimental results. Section~\ref{sec:discussion} discusses the implications for AV safety assessment. Finally, Section~\ref{sec:conclusions} concludes the paper.

\section{Theoretical Background} \label{sec:related-work}
    
    \subsection{FI Frameworks for Autonomous Systems}

        FI is a well-established technique for evaluating the dependability of computing systems by introducing faults and observing system behavior under fault conditions \cite{ziade2004survey}. FI research focuses on low-level hardware abstraction layers, including pin-level injection \cite{madeira1994rifle}, chip-level testing under radiation \cite{karlsson2002using}, and fault modeling through simulation \cite{civera2001exploiting}. At the software level, FI frameworks inject faults into CPUs, memory, and I/O subsystems using traps and event triggers \cite{kanawati1992ferrari, tsai1996approach, segall1995fiat, carreira1998xception, han1995doctor, benso1998exfi, aidemark2001goofi}. These tools supported robustness testing, redundancy evaluation, and fault diagnosis in safety-critical domains such as aerospace and robotics \cite{christensen2008fault,vardanega1995development}. A common assumption in this literature is that faults can be represented at the level of individual components, which is appropriate for many physical or deterministic failure modes but does not fully capture the interface-level perception anomalies considered in this work.

        % \cite{kanawati1992ferrari}, FTAPE \cite{tsai1996approach}, FIAT \cite{segall1995fiat}, Xception \cite{carreira1998xception}, DOCTOR \cite{han1995doctor}, EXFI \cite{benso1998exfi}, and GOOFI \cite{aidemark2001goofi} targeted CPUs, memory, and I/O subsystems, injecting faults using traps or event triggers. These tools enabled robustness testing, redundancy evaluation, and early fault diagnosis with minimal hardware overhead, particularly in safety-critical domains like aerospace, embedded systems, and robotics \cite{christensen2008fault, vardanega1995development}. However, these approaches relied on the assumption that faults could be accurately represented and injected at the level of individual components, which restricted them to physical or deterministic failure modes.

        Building on these foundations, \chg{system-level} FI has been adapted to AVs as a way to validate robustness in rare but safety-critical conditions \cite{koopman2018practical}. Frameworks such as AVFI and Kayotee simulate sensor failures and inject perturbations into neural-network inputs to study their effects on decision-making and control \cite{jha2018avfi,jha2019kayotee}. Multi-agent platforms also model agent interactions and fault-aware behaviors in dynamic environments \cite{garrido2020fault}. These approaches enable controlled experiments, but they often remain tied to specific modules, sensors, neural-network inputs, or AV-stack implementations, which can limit portability across simulators and AV architectures. In contrast, the HI framework proposed in this work injects controlled anomalies into the perception information delivered to MC, without requiring assumptions about the specific sensor modality, ML model, or MP implementation.

        % Building on these foundations, later research has begun adapting FI for AVs, which present new challenges due to their AI-driven, highly integrated architectures. FI at the system-level has gained traction as a way to validate the robustness of AV in rare but safety-critical scenarios \cite{koopman2018practical}. Frameworks such as AVFI \cite{jha2018avfi} and Kayotee \cite{jha2019kayotee} simulate sensor failures and inject perturbations into the input of neural network to study their downstream effects on control and decision-making. Multi-agent platforms, such as presented in \cite{garrido2020fault}, model agent interactions, and fault-aware behaviors in dynamic environments. However, these approaches remain tied to specific modules or components and fall short in capturing the breadth of consequences such failures can have at the perception abstraction level, which limits their applicability for extensive safety validation.

        Another group of studies examines MP robustness at the model level. Corruption benchmarks evaluate vision models under structured input distortions \cite{elgharbawy2016generic,hendrycks2019benchmarking}. Other works inject faults into neural-network parameters \cite{liu2017fault,breier2018practical,rakin2021t,laskar2022tensorfi+} or analyze internal behaviors using coverage criteria and logic inconsistencies \cite{pei2017deepxplore,chernikova2019self,boloor2020attacking}. These studies are useful for component diagnosis and robustness tuning, but they are typically isolated from the full AV stack and do not directly address how corrupted perception information affects motion planning and control.

        % Another group of studies has focused on improving MP robustness through adversarial testing at the module-level. Benchmarks have explored structured corruptions in vision models \cite{elgharbawy2016generic, hendrycks2019benchmarking}, and tools such as AV-Fuzzer \cite{li2020av} and ontology-based generators \cite{tahir2021intersection} have been used to create dangerous edge-case scenarios. Other works target vulnerabilities in internal mechanisms of ML systems by injecting faults into neural network parameters \cite{liu2017fault, breier2018practical, rakin2021t, laskar2022tensorfi+} or exploring neuron-level behaviors through coverage criteria and logic inconsistencies \cite{pei2017deepxplore, chernikova2019self, boloor2020attacking}. However, these studies operate at low abstraction levels of ML, typically isolated from the entire AV stack, and offer little support to understand how such faults manifest as hazardous behaviors when propagated through perception, planning, and controlling pipelines.

        Recent efforts move closer to the perception--control interface focusing on perception-level fault injection and error modeling. A first category injects faults into sensor streams or sensor-aligned pipeline inputs to stress perception pipelines. Examples include multisensor-fusion corruption benchmarks \cite{gao2023benchmarking}, physically plausible cross-sensor insertions \cite{gao2024multitest}, LiDAR point-cloud transformations \cite{guo2022lirtest}, and systematic camera and LiDAR FI with search-guided testing, as in FADE \cite{10.1145/3728910}. Camera failure libraries also model realistic RGB failure modes and evaluate their effects on detectors and driving agents \cite{secci2020failures,ceccarelli2022rgb}. These methods increase fidelity at the sensing layer, but they tend to remain coupled to particular modalities and pipeline implementations.

        A second category reduces dependence on detailed sensor physics by modifying the perceived world state in simulation, such as object lists, and injecting detection and parameter errors. Early work shows how perception errors can be injected by bypassing built-in sensing and feeding an altered object representation to the AV \cite{piazzoni2020modeling}. The Perception Error Model (PEM) formulation later defines perception error models as virtual components for studying how perception errors affect safety without modeling the sensors \cite{piazzoni2023pem}. This approach is closely related to the present work because it modifies the perceived world state rather than the physical sensor data. However, some PEM formulations compute errors from objects that already exist in the simulated ground truth. As a result, they can model errors such as missed detections or incorrect object parameters, but they may not naturally generate false-positive objects, that is, hallucinated objects with no corresponding real object in the scene \cite{piazzoni2023pem}. The proposed HI framework also accommodates such hallucinations and configurable combinations of error types, enabling the evaluation of a broader range of perception failures and their system-level safety consequences. Related surrogate-perception approaches for efficient simulation can make similar simplifications, which improves scalability but can restrict the distortions that can be studied \cite{sadeghi2021step}. Beyond manually specified models, sensor-agnostic statistical and generative approaches learn time-correlated error processes from production data, including Hidden Markov Model-based error models \cite{zec2018statistical} and recurrent generative adversarial network models that capture long-term temporal dependencies \cite{arnelid2019recurrent}. Uncertainty-aware perception work further motivates evaluations that account for uncertainty propagation through the stack \cite{feng2021review}. These perception-error and world-state modification approaches are closely related to perception-insufficiency injection for virtual testing, since they emulate the information delivered by perception without requiring full sensor-realistic modeling.

        Although \chg{they} do not propose FI frameworks, several studies address complementary robustness and resilience mechanisms for AV perception. HydraFusion adapts fusion according to environmental context \cite{malawade2022hydrafusion}. Hybrid redundancy approaches combine hardware and analytical redundancy for real-time fault detection and isolation \cite{jin2024hybrid}. Fault-source classification aims to diagnose the origin of sensor faults online to support remediation \cite{hou2023fault}. These studies improve operational robustness, but they are designed primarily for runtime adaptation, detection, or diagnosis rather than for portable, repeatable FI-based safety-sensitivity evaluation.

        A further related category investigates adversarial manipulations that cause misdetections or mislocalizations. Examples include LiDAR spoofing and adversarial examples aimed at affecting driving decisions \cite{cao2019adversarial}, physical-world attacks against multisensor-fusion perception designed to induce end-to-end collisions \cite{cao2021invisible}, and trajectory perturbation attacks that distort LiDAR perception through motion-compensation dependencies \cite{li2021fooling}. Although these works are not primarily designed as virtual safety-evaluation tools, they show that different perturbation mechanisms can produce similar perception-output anomalies, such as false objects, missed objects, or incorrect object locations, which may affect MC.

        These tradeoffs motivate an interface-level injection abstraction that evaluates how perceptual distortions in the information delivered by MP affect motion planning and control. Such an abstraction can complement perception-error and virtual-testing approaches by focusing on the safety consequences of distorted situational awareness without requiring a specific sensor model, calibrated error distribution, or perception pipeline.        

        % \input{Tabs/hi_unified_positioning}

        % \input{Tabs/fi_comparison}

        % Recent research efforts have aimed to close this gap by focusing on perception-level failures. However, they often remain tightly coupled to specific sensors or data fusion strategies. For example, FADE \cite{10.1145/3728910} injects faults directly into camera and LiDAR data to model real-world sensor degradations. PEM \cite{piazzoni2023pem} proposes error models to abstract failure perceptions but derives these from predefined sensor setups, which limits their generalization. HydraFusion \cite{malawade2022hydrafusion} improves robustness by dynamically adjusting the fusion based on environmental context. Jin \textit{et al.} \cite{jin2024hybrid} focus on real-time fault detection and isolation by combining hardware and analytical redundancy. In contrast, Hou \textit{et al.} \cite{hou2023fault} aim to classify the source of sensor faults in real-time, to allow diagnosis and remediation. Despite their contributions, these studies primarily address the origin of faults (e.g., sensor degradation, network faults, or fusion inconsistencies) rather than their observable consequences on situational awareness and decision-making.

    % \subsection{Faults in AV MP Systems}
    \subsection{Failures Produced by MP Systems}

\begin{figure}[h]
            \centering
            \includegraphics[scale=0.505]{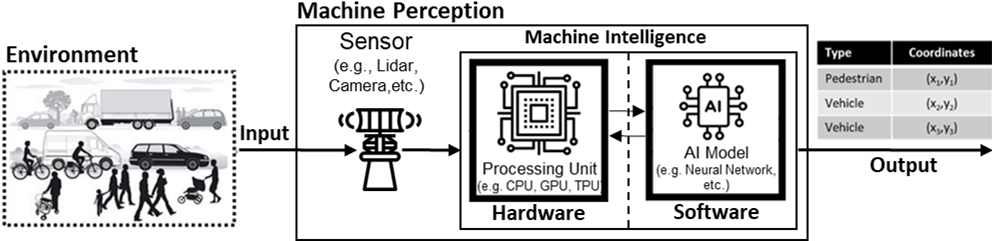}
            \caption{MP system model}
            \label{fig:fig-3}
        \end{figure}

        MP systems provide situational awareness to AVs. They interpret sensor data to detect and classify critical elements in the environment, such as vehicles, pedestrians, and signals, providing the spatial and semantic information needed for decision-making. Figure~\ref{fig:fig-3} illustrates a typical MP system, showing its main components and an abstraction of its inputs and outputs. Although a common configuration includes a camera, GPU, and ML model, perception pipelines may also integrate other sensors, such as LiDAR and radar \cite{vargas2021overview}.       
    
        % INSERT FIGURE 30

\begin{figure}[h]
            \centering
            \includegraphics[scale=0.58]{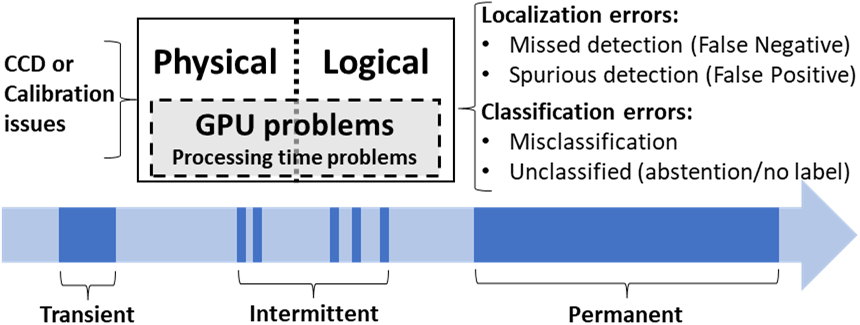}
            \caption{Taxonomy of dimensions and persistence of MP faults}
            \label{fig:fig-5}
        \end{figure}

        Regardless of the sensor modality, MP remains vulnerable to faults that can compromise safety \cite{van2018autonomous}. Faults can be classified by persistence (Figure~\ref{fig:fig-5}) \cite{avizienis2004basic}. Persistence categories include \textbf{permanent faults}, which persist until the faulty component is repaired or replaced, such as a processor damaged by overheating or firmware corruption that causes consistent malfunction \cite{avizienis2004basic}. They also include \textbf{transient faults}, which are temporary and difficult to reproduce, such as a bit flip in sensor data caused by electromagnetic interference \cite{avizienis2004basic}. In addition, \textbf{intermittent faults} \cite{qi2008no} occur at irregular intervals while the system otherwise operates normally, such as an unstable LiDAR connection that fails under specific thermal conditions. Following the AV functional model adopted in \cite{nascimento2019systematic}, faults can affect MP, MC (broadly responsible for motion planning and decision-making), or machine actuator (MA), responsible for executing control commands. This study focuses on MP-related faults whose observable effects appear as corrupted perception information delivered to MC, where motion planning and decision-making are performed.

\begin{figure}[h]
            \centering
            \includegraphics[scale=0.52]{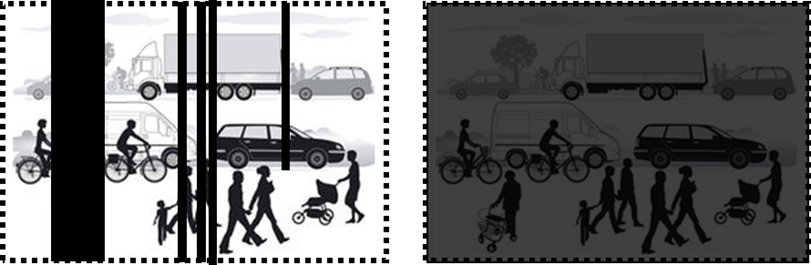}
            \caption{Illustration of observable consequences of CCD partial fault (left) and complete fault (right)}
            \label{fig:fig-8}
        \end{figure}

        Faults can also be classified by dimension \cite{avizienis2004basic}, such as hardware or software (Figure~\ref{fig:fig-5}). Hardware faults may include camera calibration drift and damage to the charge-coupled device (CCD). For example, damage to a CCD can lead to partial image corruption (Figure~\ref{fig:fig-8}, left) \cite{hainaut_ccd_artifacts} or complete frame loss (Figure~\ref{fig:fig-8}, right) \cite{hainaut_ccd_artifacts}. Both can compromise scene classification \cite{su2019one}. Beyond these examples, other fault types can jeopardize the utility of CCD sensors and LiDARs \cite{jin2023pla}, including black pixels \cite{oxinst_ccd_blemishes}, which are commonly caused by contamination of sensor material, and traps \cite{oxinst_ccd_blemishes}, which appear as dark columns produced by a charge-transfer barrier.

        % INSERT FIGURE 32

        % INSERT FIGURE 35

    \subsection{Hallucination Injection in AV Systems}

        According to Avizienis \textit{et al.} \cite{avizienis2004basic}, a failure occurs when the service delivered by a system deviates from its intended correct service because of an underlying fault. Thus, a faulty component in an MP system may lead to an MP failure, which in turn may produce perceptual distortion. In other words, an MP failure may generate a perceptual experience without a corresponding environmental stimulus. In an AV system, such MP failures can propagate to MC because they may be interpreted with the same force and clarity as genuine perceptions. Consequently, MC may adapt its behavior to distorted situational awareness, which can compromise vehicle safety.

        From a psychological perspective, perceptual experiences that occur without corresponding environmental stimuli and are perceived with the same force and clarity as genuine perceptions are known as hallucinations \cite{edition2013diagnostic}. They can manifest across sensory modalities, including auditory, visual, tactile, olfactory, and gustatory domains \cite{edition2013diagnostic, waters2018auditory}. Hallucinations can alter an individual's perception of reality, affecting situational awareness and, consequently, behavior and reactions \cite{edition2013diagnostic, waters2018auditory}. 

        Inspired by the psychological domain, this work defines MP hallucinations as MP failures that manifest as perceptual experiences occurring without corresponding environmental stimuli and perceived with the same force and clarity as genuine perceptions. Because hallucinated outputs from MP systems can mislead motion planning and control, they represent a class of AV subsystem faults that may pose a significant risk to safety, yet are often overlooked by conventional fault models. In this work, hallucinations are restricted to perception-output distortions that alter the situational awareness provided to the control module, explicitly excluding sensor-noise modeling, sensor-failure modeling, and internal ML failure mechanisms within perception pipelines. This definition establishes the level of abstraction and scope of interest adopted here for system-level safety evaluation in the presence of MP failures, regardless of which component is faulty and which fault mechanism is involved.

        % However, this novel concept requires a framework for injecting hallucinations into perception rather than focusing narrowly on component-level faults. To the best of our knowledge, prior work has not framed these distortions as a configurable hallucination taxonomy injected specifically at the perception--control interface for component-agnostic downstream safety evaluation. Therefore, in an effort to bridge literature gaps and introduce the proposed novel perspective of analysis, the present work adopts a component-agnostic approach, focusing on injecting and exploring the effects of hallucinations into the AV MP, rather than explicitly modeling fault sources. This abstraction enables faults to be injected at the level of behavioral symptoms, allowing for unified evaluation across AV platforms regardless of technical architecture, components, or sensor configuration. By decoupling failure modeling from implementation details, this approach facilitates standardized and interoperable safety testing in simulations, offering new insights into the propagation and emergent system behavior resulting from MP failures. Moreover, it can accelerate AV safety research by supporting anticipated AV evaluations against novel and potential hallucination types prior to research that uncovers their root causes and fault mechanisms.

        However, this concept requires a framework for injecting hallucinations into perception rather than focusing narrowly on component-level faults. To the best of our knowledge, prior work has not framed these distortions as a configurable hallucination taxonomy injected specifically at the perception--control interface for component-agnostic safety evaluation. Therefore, the present work adopts a component-agnostic approach that focuses on injecting and exploring hallucination effects in the information delivered from MP to MC, rather than explicitly modeling fault sources. This abstraction represents faults at the level of behavioral symptoms, enabling unified evaluation across AV platforms regardless of technical architecture, components, or sensor configuration. By decoupling failure modeling from implementation details, the approach supports standardized and interoperable safety testing in simulation while preserving focus on how perceptual distortions affect AV behavior. Moreover, it can support the evaluation of potential hallucination types before their root causes and fault mechanisms are fully understood.        

        This shift from modeling fault causation to analyzing observable effects addresses a broader need for system-level safety assessment beyond individual modules. Rather than focusing on the failure of a particular component, this perspective emphasizes how the system behaves when perception is degraded. By representing failures as parameterizable hallucinations, the approach enables reproducible and quantitative evaluation across diverse scenarios, supporting a more systematic and generalizable understanding of AV safety. Therefore, the present work adopts hallucination injection (HI), defined here as the controlled injection of perception-output distortions into the information delivered from MP to MC. Rather than explicitly modeling fault sources, HI focuses on exploring how such distortions affect AV behavior.

    % \subsection{Types of Hallucinations in AV Systems}

        Although the concept of hallucinations in the AV domain is introduced in this study to raise the level of abstraction of the analysis and avoid focusing on component-level fault mechanisms and models, plausible hallucinations should still be grounded in known or plausible MP failures and their underlying fault mechanisms. For this reason, each hallucination type defined here is briefly associated with plausible underlying faults or components that may produce it.

\begin{figure}[h]
            \centering
            \includegraphics[scale=0.52]{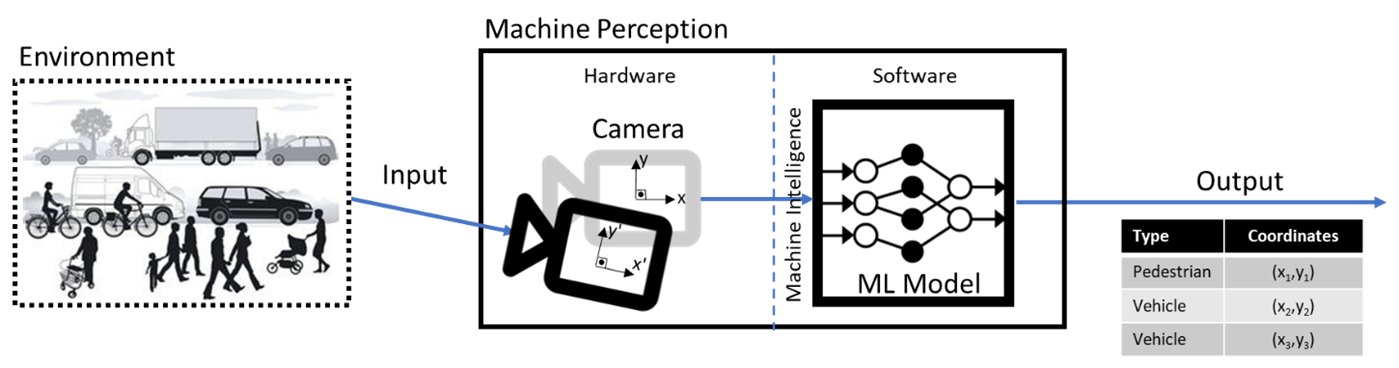}
            \caption{Camera calibration drift}
            \label{fig:fig-6}
        \end{figure}

        Hallucinations can occur across different sensing modalities, including cameras, LiDAR, radar, and sensor-fusion systems, and may arise from hardware, software, or environmental interactions. For example, calibration drift, that is, the gradual deviation of sensor outputs over time, can result from environmental factors such as temperature variation or mechanical shock \cite{talluru2014calibration, halkon2021establishing}, shifting coordinate frames and degrading object location (Figure~\ref{fig:fig-6}). Such distortions can alter the sensor reference system (${O, x, y, z}$ to ${O, x', y', z'}$), leading to spatial misinterpretation (Figure~\ref{fig:fig-6}).

        % INSERT FIGURE 33
        
\begin{figure}[h]
            \centering
            \includegraphics[scale=1]{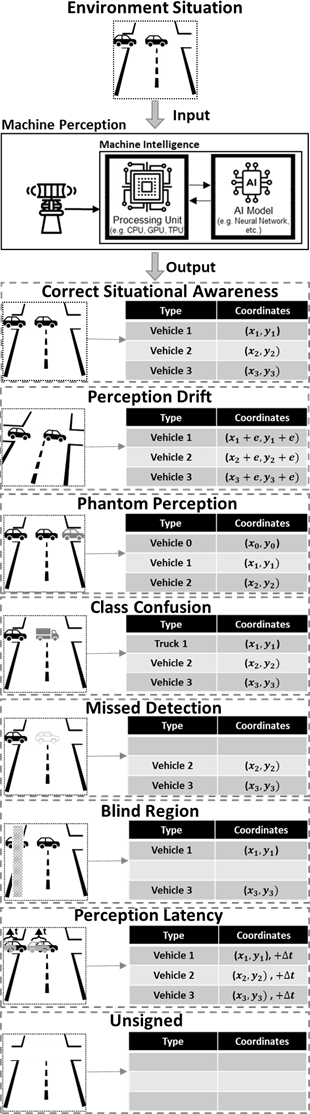}
            \caption{Effect of distinct MP hallucinations on the AV situational awareness}
            \label{fig:fig-9}
        \end{figure}

        Specific hallucinations, such as recognition latency, missed detections, false classifications, and spatial drift, can emerge from different causes across hardware and software layers \cite{avizienis2004basic} (Figure~\ref{fig:fig-9}). For example, \textbf{Perception Linear Drift} and \textbf{Perception Angular Drift} can cause spatial location errors by displacing detected objects and affecting subsequent reasoning and planning \cite{kuutti2018survey}. Class confusion, such as misidentifying a pedestrian as a cyclist or vehicle, as well as spurious detections, can manifest as \textbf{Phantom Perception} (false positives), in which the system detects or classifies objects that are not present or assigns them incorrect labels, thereby compromising situational awareness and potentially leading to unsafe responses \cite{janai2020computer}. \textbf{Missed Detections} (false negatives) may result from poor generalization in ML models, degraded input quality, or synchronization errors between system components \cite{hoiem2012diagnosing, janai2020computer}. In contrast, a \textbf{Blind Region} refers to an entire spatial area that the perception system fails to observe, rather than to a single missed entity, such as when sensor mounting constraints or field-of-view limitations systematically obscure part of the environment. \textbf{Perception Latency} can result from overloaded processing units, including GPUs subject to scheduling delays, memory bottlenecks, or load spikes \cite{amert2017gpu, hsieh2016transparent, yang2018avoiding}, as well as from inefficient software pipelines or timing problems in sensor data transmission \cite{xu2020preview}. \textbf{Unsigned} \chg{hallucinations}, such as temporal instability and adversarial vulnerabilities, further illustrate how hallucinations can arise from both hardware- and software-related causes \cite{su2019one, goodfellow2014explaining}.

        % INSERT FIGURE 36

\section{Methodology}\label{sec:methodology}

    Injecting faults into safety-critical systems under real traffic conditions would be excessively risky, especially when evaluating how AVs react to hallucinated perception outputs and the resulting safety implications. However, other scientific fields have successfully studied the behavioral consequences of perceptual modifications under safer experimental conditions, which provided inspiration for the present investigation. In particular, psychological and behavioral research has used virtual environments (VEs) to systematically manipulate a user’s sensory reality. Researchers can alter visual, auditory, and haptic feedback to create perceptual experiences that would be dangerous, counterintuitive, expensive, or impossible to control in the physical world \cite{bailenson2018experience}. This capability enables direct investigation of how specific perceptual modifications affect human cognition, emotion, and behavior \cite{box2022future}. An illustrative example is avatar embodiment with visually altered body features, such as an arm appearing to be made of stone, which made participants feel heavier and stiffer, move more slowly, and exhibit changes in motor cortical excitability \cite{buetler2022tricking}. This level of experimental control over the perception of self and environment provides a useful analogy for exploring causal links between perception and behavior.

    Driving simulators represent a specialized class of VEs that extend these principles to transportation research. Such platforms allow researchers to study driver perception, emotion, and decision-making under controlled and repeatable conditions. Previous studies have investigated how onboard voice interfaces influence driver affect and safety outcomes \cite{nass2005improving, harris2011emotion}. Others have modeled accident risk based on visual cues, such as facial expressions \cite{jabon2010facial}. Additional work has examined behavioral adaptation in the presence of partially autonomous driving systems \cite{lee2014partially, koo2015did}. These examples suggest that simulators are useful tools for investigating how driver perception and behavior are influenced by advanced vehicle systems.    
    
    A critical factor in the success of such studies is the plausibility of the virtual scenario. A plausible illusion is the cognitive interpretation that events occurring in the VE are real, not in the sense of photorealism, but in the sense that they are happening and remain coherent with the rules of the environment \cite{slater2009place}. For a participant to react authentically to a perceptual modification, the scenario must be accepted as a credible sequence of events, even if it is fantastical. Without plausibility, the user may disengage or respond to the artificiality of the setup rather than the intended stimulus. Therefore, plausibility is something that the user continuously and unconsciously evaluates while immersed in a VE.    
    
    In the present study, inspired by analogous investigations in the Behavioral Sciences, a VE was used to evaluate the proposed HI module under controlled and repeatable conditions. The analogy with human-subject VE research motivates the use of plausible simulated scenarios, since the injected hallucinations should appear as coherent events within the simulated driving environment rather than as arbitrary artifacts. In this work, plausibility is therefore treated as a design consideration for the simulation setup, not as a separate variable evaluated by the AV. The remaining methodological choices, procedures, protocols, and materials are described in the following subsections.

    \subsection{Experimental Use Case}\label{sub-sec:experimental_use_case}

\begin{figure}[h]
            \centering
            \includegraphics[scale=0.69]{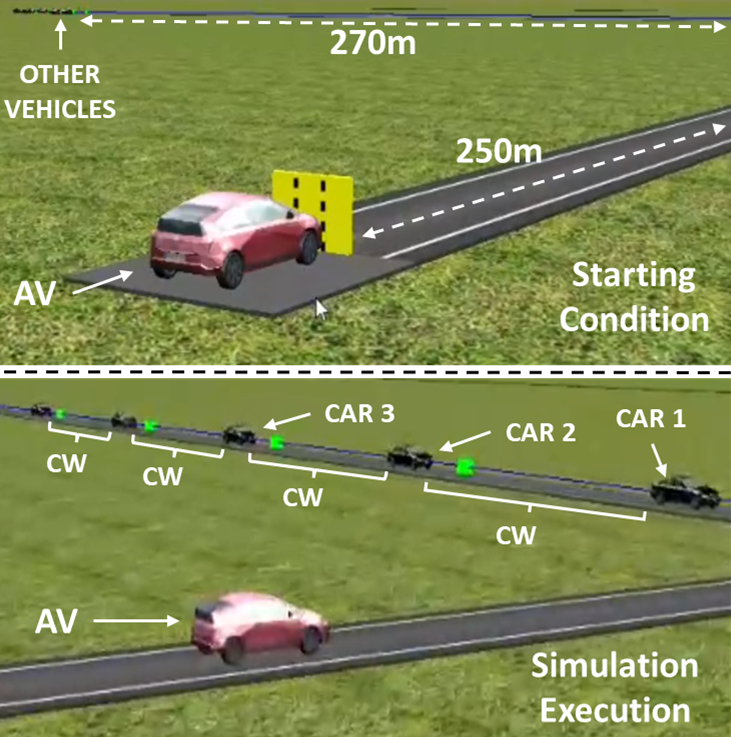}
            \caption{Snapshots of starting condition (top) and execution of simulation (bottom)}
            \label{fig:fig-23}
        \end{figure}

        A simulated intersection scenario (Figure~\ref{fig:fig-23}) was used as the experimental testbed because of its high accident potential and relevance to safety evaluation research \cite{swanson2019statistics}. Intersections are critical points in the road network, accounting for about one quarter of traffic fatalities and nearly half of all injuries in the U.S. \cite{FHWAIntersectionSafety}. Their complexity makes them a relevant setting for assessing fault tolerance in AVs.

        To complement general intersection crash statistics, this study draws on an exploratory analysis of NHTSA Standing General Order (SGO) crash reports for Automated Driving Systems (ADS) and Advanced Driver Assistance Systems (ADAS) vehicles from 2021 to 2025 \cite{SGO_21_25}. In these reports, intersections appear among the most frequent roadway types and become the most prevalent category after filtering to non-stationary pre-crash movements and cases alleging injuries (Appendix~\ref{app:sgo}). These SGO results are used only to support scenario relevance, since they reflect reported incidents and are not exposure-normalized.        
        
        In the simulated scenario, the AV must cross a one-way street intersected by five other vehicles that have the right of way. These vehicles start approximately 270 meters from the intersection and run from rest to a maximum speed of 54 km/h. The AV starts 250 meters from the intersection and must identify a safe time window, referred to as a candidate window (CW), to cross without violating traffic rules. CWs are defined as predicted time-space gaps in traffic during which the AV can safely cross the intersection, as computed by the control module.        
        
        Each simulation run ends when the AV either (1) crosses safely, (2) causes a collision, or (3) halts indefinitely because no safe crossing opportunity is available. Any simulation error, such as a simulator crash or runtime failure, is treated as an invalid execution, and the corresponding log is discarded. The condition is then repeated until the configured number of valid executions is reached. This procedure ensures that the analysis includes only fully valid executions.

        % INSERT FIGURE 56

    % \subsection{Simulation Environment}\label{sub-sec:simulation_environment}

        % All simulations were conducted using the real-time component of the USP54 framework \cite{vismari2018simulation}, a research simulation environment previously used in AV safety studies \cite{vismari2018simulation, molina2018enhancing, naufal20172}. USP54 integrates established open-source tools \cite{balkus2022survey}, including VEINS \cite{sommer2019veins}, SUMO \cite{krajzewicz2012recent}, OMNET++ \cite{varga2001discrete}, and OpenDS \cite{math2013opends}, to support real-time and accelerated simulation of complex traffic scenarios with vehicle-to-everything (V2X) communication. To support the objectives of the present study, additional implementations were developed on top of this environment to enable HI experiments and extend the scope of the original framework.  

    \subsection{Simulation Environment}\label{sub-sec:simulation_environment}

        All simulations were conducted using an extended version of the real-time simulation module of the simulation-based safety analysis framework \cite{vismari2018simulation}, a research environment previously used in AV safety studies \cite{vismari2018simulation, molina2018enhancing, naufal20172}. The original framework combines real-time and fast-time simulation approaches and integrates established open-source tools \cite{balkus2022survey}, including VEINS \cite{sommer2019veins}, SUMO \cite{krajzewicz2012recent}, OMNeT++ \cite{varga2001discrete}, and OpenDS \cite{math2013opends}, to support the analysis of AV behavior in complex traffic scenarios with vehicle-to-everything (V2X) communication. To support the objectives of the present study, additional implementations were developed on top of this environment to enable HI experiments and extend the scope of the original framework.

\begin{figure}[h]
            \centering
            \includegraphics[scale=0.62]{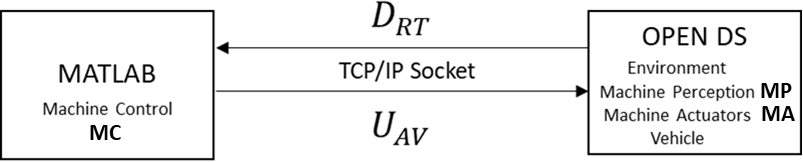}
            \caption{System components distribution and integration across OpenDS and MATLAB}
            \label{fig:fig-11}
        \end{figure}

        Figure~\ref{fig:fig-11} illustrates the real-time simulation (ReTS) feature of the simulation framework. Traffic scenarios are executed in OpenDS, which handles environmental elements, vehicle behavior, and perception modules. The MP system and actuators are also implemented in OpenDS. The control algorithms are executed in MATLAB, which \chg{receives} real-time updates from OpenDS through matrix $D_{RT}$, containing the position coordinates of each vehicle, and \chg{returns} control commands through matrix $U_{AV}$, which specifies the AV throttle, brake, and steering-wheel angle. Communication between OpenDS and MATLAB is handled via a socket interface.

    \subsection{Hallucination Injection (HI) Module} \label{sub-sec:hallucination_injection_module}

\begin{figure}[h]
            \centering
            \includegraphics[scale=0.65]{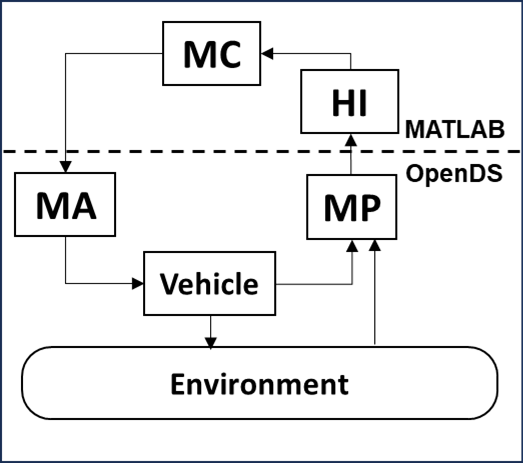}
            \caption{Updated ReTS simulation environment with HI module}
            \label{fig:fig-16}
        % \end{figure*}
        \end{figure}

        To support this study, the proposed HI module was implemented (Figure~\ref{fig:fig-16}). This module enables the simulation of different hallucination scenarios through the configurable properties summarized in Table~\ref{tab:investigated_components}. The HI module was implemented in MATLAB. The following subsections describe each HI property implemented as an HI variable.

        \setlength{\tabcolsep}{1.5pt}
% \begin{table*}[h]
\begin{table}[h]
\caption{HI module's properties}
\label{tab:investigated_components}
\centering
\begin{threeparttable}
% \begin{tabularx}{\textwidth}{|>{\centering\arraybackslash}m{1.7cm}|m{6cm}|>{\centering\arraybackslash}m{0.9cm}|}
% \begin{tabularx}{\textwidth}{>{\centering\arraybackslash}m{1.7cm} m{5.6cm} >{\centering\arraybackslash}m{0.9cm}}
\begin{tabularx}{\textwidth}{
    >{\centering\arraybackslash\scriptsize}m{1.4cm} % Property
    >{\centering\arraybackslash\scriptsize}m{2.8cm} % Variable
    >{\scriptsize}m{3.3cm}                             % Description
    >{\centering\arraybackslash\scriptsize}m{0.5cm} % Type*
}
% \begin{tabularx}{\textwidth}{|m{2.5cm}|m{2.8cm}|m{2.8cm}|}
 % \toprule
 \cmidrule(){1-4}
 \textbf{Property} & \textbf{Variable} & \textbf{Description} & \textbf{Type*} \\ 
 % \hline
\cmidrule(){1-4}
   Active & \textit{ModuleActivation} &  Setting the HI module to ON activates the injection of the configured signal, while setting it to OFF deactivates it & C \\
 
 \cmidrule(){1-4}
  Type & \textit{HallucinationType} & Type of hallucination injected into the AV during a specific simulation execution & C \\
 % \hline
 \cmidrule(){1-4}

   Domain & \textit{AffectedDomain} & \chg{Domain} \chg{affected} by the hallucination injected during a specific simulation execution & C \\
 
 \cmidrule(){1-4}

   Configuration & \textit{HallucinationConfiguration} & Configuration parameters of the hallucination injected into the AV during a specific simulation execution & C \\
 \cmidrule(){1-4}

   Probability & \textit{HallucinationProbability} & Probability of the hallucination occurrence during a specific simulation execution & N \\
 \cmidrule(){1-4}

   Persistence & \textit{HallucinationPersistence} & Hallucination Persistence configured for a specific simulation execution & C \\
 \cmidrule(){1-4}

\end{tabularx}

% \end{table*}
\smallskip
\scriptsize
\begin{tablenotes}
    \RaggedRight
    % \item[]* Type: C = Categoric, N = Numeric, B = Binary
    \item[]* Type: C = \chg{Categorical,} N = Numeric
\end{tablenotes}
\end{threeparttable}
\end{table}

        \begin{figure*}[h]
            \centering
            \includegraphics[scale=0.672]{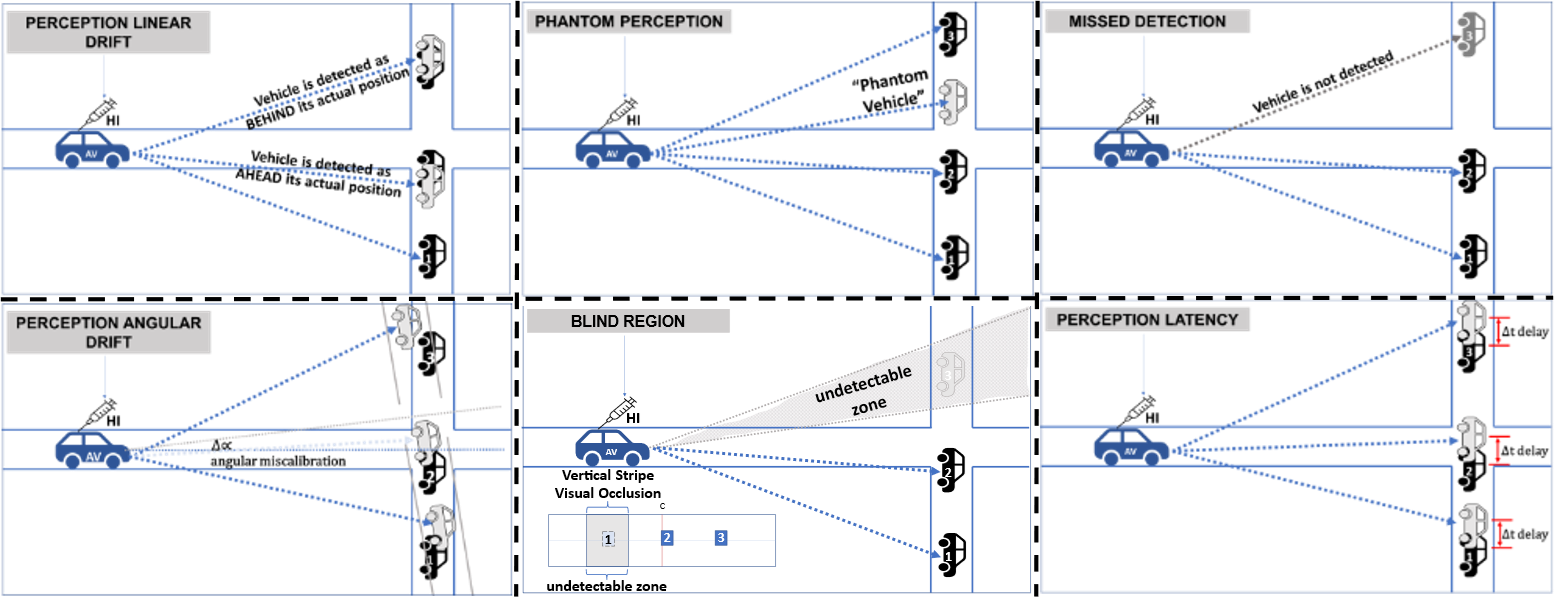}
            \caption{Hallucination types implemented in the HI framework}
            \label{fig:failure_modes}
        \end{figure*}

        As shown in Figure~\ref{fig:fig-16}, the perception output delivered to MC is treated as the injection interface in this study. In the experiments, this output is derived from simulation ground truth and modified to emulate hallucination-like effects without explicitly modeling sensor physics. This design supports controlled analysis of how perception-output distortions affect MC decisions.

        The \textit{ModuleActivation} variable is a categorical indicator that specifies whether a hallucination is injected during a given simulation run. It takes the values \textit{ON} or \textit{OFF}, allowing straightforward comparison between simulations with and without HI.        
        
        The categorical variable \textit{HallucinationType} represents six representative hallucination types that can be injected into the MP system. Table~\ref{tab:failuretypes_labels} lists all six implemented hallucination types, and Figure~\ref{fig:failure_modes} illustrates them.

\setlength{\tabcolsep}{1pt}
% \begin{table*}[h]
\begin{table}[h]
\caption{\textit{HallucinationType} categories}
\label{tab:failuretypes_labels}
\centering
% \begin{tabularx}{\textwidth}{|m{2.5cm}|m{9cm}|m{5cm}|}
% \begin{tabularx}{\textwidth}{|m{2.5cm}|m{2.8cm}|m{2.8cm}|}
% \begin{tabularx}{\textwidth}{|m{3cm}|m{5.7cm}|}
% \begin{tabularx}{\textwidth}{|m{3.1cm}|m{5.6cm}|}
% \begin{tabularx}{\textwidth}{|m{0.7cm}|m{2.7cm}|m{5.2cm}|}
% \begin{tabularx}{\textwidth}{m{0.7cm} m{2.7cm} m{5.2cm}}
\begin{tabularx}{\textwidth}{m{1.0cm} m{2.7cm} m{4.7cm}}
 % \toprule
 \cmidrule(){1-3}
 \textbf{Id} & \textbf{Category} & \textbf{Description} \\ 
 % \hline
 % \cmidrule(){1-3}
 % T00 & \textit{None} & Benchmark \\
 % \hline
 \cmidrule(){1-3}
 LinDrift & \textit{PerceptionLinearDrift} & Perturbs the perceived coordinates of a detected vehicle around its true position \\
 \cmidrule(){1-3}

 AngDrift & \textit{PerceptionAngularDrift} & Introduces an angular deviation in the detected vehicle position \\
 \cmidrule(){1-3}

 Phant & \textit{PhantomPerception} & Non-existent vehicles are detected (false positives) \\
 \cmidrule(){1-3}

 Missed & \textit{MissedDetection} & Existing vehicles are not detected (false negatives) \\
 \cmidrule(){1-3}

 Blind & \textit{BlindRegion} & Occluded regions in field-of-view that may obstruct object detection \\
 \cmidrule(){1-3}

 Latency & \textit{PerceptionLatency} & Delays are introduced into the perception mechanisms, leading to delayed situational awareness sent to MC \\
 \cmidrule(){1-3}
\end{tabularx}
% \end{table*}
\end{table}

        The \textit{PerceptionLinearDrift} \chg{hallucination} perturbs the perceived coordinates of detected objects with respect to their true positions, introducing a \chg{proportional} positional offset. \chg{Specifically, for each detected crossing vehicle $i$, the true longitudinal distance to the intersection, $d_i$, is replaced by the perceived distance $d_i' = d_i(1+u)$, where $u \sim U(-1,1)$. The value of $u$ is sampled once per simulation run and remains fixed throughout the run, and the same sampled value is applied to all detected crossing vehicles.} This distortion can cause inappropriate vehicle behavior, such as sudden braking or failure to respond properly to hazards \cite{zhao2023improving, kochanthara2024safety, van2018autonomous, yeong2021sensor}. The \textit{PerceptionAngularDrift} hallucination simulates angular miscalibration by applying a coordinate transformation to detected object centroids based on a configured angular offset. The MP system then processes misaligned data without mathematical compensation, which may degrade decision quality \cite{yeong2021sensor}. As a result, the compromised information delivered to MC can lead to incorrect assessments and unsafe decisions. The \textit{PhantomPerception} hallucination introduces fictitious objects into the MP output, prompting reactions to non-existent entities that may lead to unnecessary or unsafe maneuvers. This tests the system's ability to reject spurious detections. The \textit{MissedDetection} hallucination suppresses data from actual vehicles, simulating missed detections that may lead to unsafe decisions because the path is incorrectly assumed to be clear \cite{koopman2016challenges}. \chg{The} \textit{BlindRegion} hallucination simulates artifacts that obscure part of the visual field, degrading object detection. Previous work shows that such distortions can significantly affect \chg{artificial neural network (ANN)-based} perception \cite{weng2018evaluating, geng2022robust}, increasing the risk of false positives and false negatives. By introducing structured occlusions into simulated perception inputs, it becomes possible to assess the resilience of AV perception systems and to investigate strategies for mitigating such image-quality problems. The \textit{PerceptionLatency} hallucination simulates latency in the MP system by freezing detection data for a configurable number of cycles while new inputs are buffered. After this interval, the buffered data are replayed sequentially, emulating delayed perception of the environment. This reflects processing or transmission delays that impair real-time decision-making and may lead to outdated situational awareness and increased navigation risk \cite{rajkumar2010cyber}. Previous work has addressed such delays through edge computing to reduce latency \cite{premsankar2018edge}, whereas \cite{xu2020preview} emphasizes the need for efficient architectures to ensure safety in dynamic conditions. By implementing \textit{PerceptionLatency}, this study enables controlled analysis of timing anomalies in autonomous systems.

        The categorical variable \textit{AffectedDomain} (Table~\ref{tab:dimensionaffected_labels}) is a derived analysis label that groups hallucination types according to the aspect of perception information they distort. It is not an independent injection parameter. Rather, it supports aggregate analysis of the simulation results by allowing safety effects to be compared across broader perception-output domains. The \textit{ObjectPosition} \chg{category} groups hallucinations that affect the perceived coordinates or localization of detected objects, including \textit{PerceptionLinearDrift} and \textit{PerceptionAngularDrift}. The \chg{\textit{ObjectRecognition}} \chg{category} groups hallucinations that affect object existence or detection, including \textit{PhantomPerception}, \textit{MissedDetection}, and \textit{BlindRegion}. The \textit{InformationTiming} \chg{category} groups hallucinations that affect the temporal freshness of situational awareness, such as \textit{PerceptionLatency}. 
        
        % It should be noted that no unsigned hallucination cases were implemented in the current version of the HI framework. Although these cases are important, they are left for future work because they are closely related to cyberattacks and belong to a distinct and already well-developed research domain that is outside the scope of the present study.

        % INSERT TABLE 10
        % \setlength{\tabcolsep}{1pt}
% % \begin{table*}[h]
% \begin{table}[h]
% \caption{Description of \texttt{DimensionAffected} labels}
% \label{tab:dimensionaffected_labels}
% \centering
% % \begin{tabularx}{\textwidth}{|m{2.5cm}|m{9cm}|m{5cm}|}
% % \begin{tabularx}{\textwidth}{|m{2.5cm}|m{2.8cm}|m{2.8cm}|}
% \begin{tabularx}{\textwidth}{|m{3cm}|m{5.7cm}|}
%  % \toprule
%  \cmidrule(){1-2}
%  \textbf{Label} & \textbf{Description} \\ 
%  % \hline
%  \cmidrule(){1-2}
%  D00\_None & Benchmark \\
%  % \hline
%  \cmidrule(){1-2}
%  D01\_ObjectPosition & Fault injection affects the coordinate dimension \\
%  \cmidrule(){1-2}

%  D02\_ObjectRecognition & Fault injection affects object recognition \\
%  \cmidrule(){1-2}

%  D03\_InformationTiming & Fault injection affects timing information \\
%  \cmidrule(){1-2}
% \end{tabularx}
% % \end{table*}
% \end{table}

\setlength{\tabcolsep}{1pt}
% \begin{table*}[h]
\begin{table}[h]
\caption{\textit{AffectedDomain} categories}
\label{tab:dimensionaffected_labels}
\centering
% \begin{tabularx}{\textwidth}{|m{2.5cm}|m{9cm}|m{5cm}|}
% \begin{tabularx}{\textwidth}{|m{2.5cm}|m{2.8cm}|m{2.8cm}|}
% \begin{tabularx}{\textwidth}{|m{3cm}|m{5.7cm}|}
% \begin{tabularx}{\textwidth}{|m{0.8cm}|m{2.2cm}|m{5.6cm}|}
% \begin{tabularx}{\textwidth}{m{0.8cm} m{2.2cm} m{5.6cm}}
\begin{tabularx}{\textwidth}{m{0.8cm} m{2.2cm} m{5.3cm}}
 % \toprule
 \cmidrule(){1-3}
 \textbf{Id} & \textbf{Category} & \textbf{Perception output domain} \\ 
 % \hline
 % \cmidrule(){1-3}
 % D00 & \textit{None} & Benchmark \\
 % \hline
 \cmidrule(){1-3}
 Pos & \textit{ObjectPosition} & Perceived coordinates or localization of detected objects \\
 \cmidrule(){1-3}

 Rec & \textit{ObjectRecognition} & Object existence, detection, or recognition \\
 \cmidrule(){1-3}

 Time & \textit{InformationTiming} & Temporal freshness of perception information\\
 \cmidrule(){1-3}
\end{tabularx}
% \end{table*}
\end{table}

        The categorical variable \textit{HallucinationConfiguration} (Table~\ref{tab:failureconfiguration_labels}) defines the configuration parameters associated with a hallucination injected during a simulation run. Different configurations are associated with each \textit{Hallucination Type}. \chg{When} no additional configuration is required for a given \textit{Hallucination Type}, the value \textit{None} is assigned. \textit{PerceptionLinearDrift} is uniquely identified by the \textit{Location} category. Hallucination types \textit{MissedDetection} and \textit{PhantomPerception} require specification of which car is affected or used for phantom perception. The identifier \chg{$x$} corresponds to cars 1, 2 and 3, respectively. For \textit{PerceptionAngularDrift}, the configuration defines the camera rotation angle relative to the AV longitudinal axis, with standard values of 5\textdegree, 10\textdegree, 20\textdegree, and 25\textdegree, either to the left (L) or to the right (R), resulting in categories such as \textit{ANG05L} \chg{(5\textdegree}\ rotation to the left) and \textit{ANG25R} \chg{(25\textdegree}\ rotation to the right). \chg{Similarly, the \textit{BlindRegion} hallucination is configured by the angular width of the occluded region in an angular coordinate system centered on the AV. In this study, widths of 40\textdegree, 50\textdegree, and 60\textdegree\ are used, producing categories such as \textit{BLIND50L} (a 50\textdegree-wide occlusion on the left side). The center of the occluded region is drawn uniformly between 0\textdegree\ (the AV heading) and 90\textdegree\ to the left at the beginning of each run. Crossing vehicles located within the occluded region are removed from the perception output.} The occlusion stripes are restricted to the left side because the vehicles approach from that direction, as illustrated in the Blind part of Figure~\ref{fig:failure_modes}. In the case of \textit{PerceptionLatency}, the configuration corresponds to the number of cycles by which information is delayed in the AV system. Two standardized values were used: 20 and 40 execution cycles, denoted by \textit{LAT20} and \textit{LAT40}, respectively. \chg{An execution cycle corresponds to one iteration of the MATLAB perception--control loop. On average, each cycle lasted $14.5$\,ms. Therefore, \textit{LAT20} and \textit{LAT40} delayed the perception information by approximately $0.29$\,s and $0.58$\,s, respectively.} These predefined configurations reduce the dimensionality of the variable space and facilitate pairwise analyses and framework validation.
        
        % Future studies may expand the range of possible configurations beyond the present validation-oriented scope.

        % INSERT TABLE 11
        % \setlength{\tabcolsep}{1pt}
% % \begin{table*}[h]
% \begin{table}[h]
% \caption{Description of \textit{FailureConfiguration} labels}
% \label{tab:failureconfiguration_labels}
% \centering
% % \begin{tabularx}{\textwidth}{|m{2.5cm}|m{9cm}|m{5cm}|}
% % \begin{tabularx}{\textwidth}{|m{2.5cm}|m{2.8cm}|m{2.8cm}|}
% \begin{tabularx}{\textwidth}{|m{2cm}|m{6.7cm}|}
%  % \toprule
%  \cmidrule(){1-2}
%  \textbf{Label} & \textbf{Description} \\ 
%  % \hline
%  \cmidrule(){1-2}
%  C00\_None & Benchmark \\
%  % \hline
%  \cmidrule(){1-2}
%  C01\_Car\_\texttt{X} & Fault injections targeting individual cars in the line $(\texttt{X} = 1, 2, or 3)$ \\
%  \cmidrule(){1-2}

%  C02\_Angle\_\texttt{Y}o\texttt{D} & Angle deviation settings: $\texttt{Y} = \{5, 10, 20, 25, 40, 50, 60\}$, $\texttt{D}$ = \{L (left) or R (right)\} \\
%  \cmidrule(){1-2}

%  C03\_Periods\_\texttt{Z} & Delay configurations with $\texttt{Z} = \{20, 40\}$ simulation cycle periods \\
%  \cmidrule(){1-2}
% \end{tabularx}
% % \end{table*}
% \end{table}

\setlength{\tabcolsep}{1pt}
\begin{table}[h]
\caption{\textit{HallucinationConfiguration} categories}
\label{tab:failureconfiguration_labels}
\centering
\begin{tabularx}{\textwidth}{m{4.2cm} m{4.2cm}}
 \cmidrule(){1-2}
 \textbf{ID \& Category} & \textbf{Description} \\ 

 \cmidrule(){1-2}
 \footnotesize{Location} & \footnotesize{A systemic linear drift affecting the perceived position of all detected \chg{vehicles}} \footnotesize{\chg{(proportional error of up to 100\% of each vehicle's distance to the intersection)}} \\

 % \cmidrule(){1-2}
 % NONE & Benchmark (no hallucination configuration) \\
 \cmidrule(){1-2}
 \footnotesize{Car1, Car2, Car3} & Selects \chg{the} crossing car whose perceived attributes will be altered in the AV MP \\
 \cmidrule(){1-2}
 % \scriptsize{ANG05L, ANG10L, ANG20L, ANG25L} & \scriptsize{Camera rotated $Y \in \{5,10,20,25\}$ degrees to the left} \\
 % \cdashline{1-2}
 % \scriptsize{ANG05R, ANG10R, ANG20R, ANG25R} & \scriptsize{Camera rotated $Y \in \{5,10,20,25\}$ degrees to the right} \\
 \footnotesize{Ang05L, Ang10L, Ang20L, Ang25L, Ang05R, Ang10R, Ang20R, Ang25R} & Sets the camera miscalibration rotating 5\textdegree, 10\textdegree, 20\textdegree, \chg{or 25\textdegree}\ to the left (L) or right (R) side of the AV, considering \chg{0\textdegree}\ the \chg{axis} parallel to the AV trajectory \\
 
 \cmidrule(){1-2}
 % \scriptsize{BLIND40L, BLIND50L, BLIND60L} & \scriptsize{Visual occlusion at $Y \in \{40,50,60\}$ degrees to the left} \\
 % \cdashline{1-2}
 % \scriptsize{BLIND40R, BLIND50R, BLIND60R} & \scriptsize{Visual occlusion at $Y \in \{40,50,60\}$ degrees to the left} \\
 \footnotesize{Blind40L, Blind50L, Blind60L} & \chg{Sets the angular width of the occlusion stripe (40\textdegree, 50\textdegree, or 60\textdegree) on the left side of the AV. Its center is drawn uniformly between 0\textdegree\ and 90\textdegree\ to the left in each run.} 0\textdegree\ is parallel to the AV trajectory\\
 \cmidrule(){1-2}
 \footnotesize{Lat20, Lat40} & Selects the delay to be introduced between the simulation status and the MP in number of simulation cycles (20 or 40\chg{; approximately 0.29~s and 0.58~s}) \\
 \cmidrule(){1-2}
\end{tabularx}
\end{table}

% \setlength{\tabcolsep}{1pt}
% % \begin{table*}[h]
% \begin{table}[h]
% \caption{Description of \textit{HallucinationConfiguration} labels}
% \label{tab:failureconfiguration_labels}
% \centering
% % \begin{tabularx}{\textwidth}{|m{2.5cm}|m{9cm}|m{5cm}|}
% % \begin{tabularx}{\textwidth}{|m{2.5cm}|m{2.8cm}|m{2.8cm}|}
% % \begin{tabularx}{\textwidth}{|m{2cm}|m{6.7cm}|}
% % \begin{tabularx}{\textwidth}{|m{1.3cm}|m{1.4cm}|m{5.7cm}|}
% \begin{tabularx}{\textwidth}{m{1.3cm} m{1.4cm} m{5.7cm}}
%  % \toprule
%  \cmidrule(){1-3}
%  \textbf{Label} & \textbf{Name} & \textbf{Description} \\ 
%  % \hline
%  \cmidrule(){1-3}
%  C00 & \textit{None} & Benchmark \\
%  % \hline
%  \cmidrule(){1-3}
%  C01\_00CX & \textit{Car\_{X}} & Hallucinations targeting individual cars in the line X = \{1, 2, 3\} \\
%  \cmidrule(){1-3}

%  C02\_AYD & \textit{Angle\_{Y}\textdegree{D}} & \parbox{5.7cm}{Angle deviation settings:\\Y = $\{05, 10, 20, 25, 40, 50, 60\}$,\\D = \{L (left) or R (right)\}} \\
%  \cmidrule(){1-3}

%  C03\_0PZ & \textit{Periods\_{Z}} & Delay configurations with $\texttt{Z} = \{20, 40\}$ simulation cycle periods \\
%  \cmidrule(){1-3}
% \end{tabularx}
% % \end{table*}
% \end{table}

        The \textit{HallucinationProbability} variable defines the \chg{activation probability of a hallucination}, with values ranging from $0$ to $1$. \chg{Whether it is evaluated at each simulation cycle or once per run depends on \textit{HallucinationPersistence}, as described below.} In the simulations conducted for this study, probabilities of $1\%$, $5\%$, $10\%$, $25\%$, and $50\%$ were used to represent a broad spectrum of hallucination likelihoods, from rare to frequent events.

        % The \textit{HallucinationPersistence} variable is categorical and represents how the hallucination manifests during a simulation run. When no hallucination is injected, the simulation is categorized as \textit{Baseline}. If the hallucination is intermittent, the category \textit{Intermittent} is used, whereas a permanently injected hallucination is denoted by \textit{Permanent}.

        The \textit{HallucinationPersistence} variable is categorical and represents the temporal behavior of the injected hallucination during a simulation run. When no hallucination is injected, the simulation is categorized as \textit{Baseline}. In the \textit{Intermittent} category, a random draw is performed at each simulation cycle, and the hallucination is activated for that cycle according to the configured \textit{HallucinationProbability}. Therefore, intermittent hallucinations may occur sporadically throughout the run. \chg{In the \textit{Permanent} category, a single random draw is performed at the beginning of each run, and the hallucination is activated according to the configured \textit{HallucinationProbability}. When activated, the hallucination remains active until the end of the run. Otherwise, the run proceeds without the hallucination.}

    \subsection{Machine Perception (MP) Module}  \label{sub-sec:machine_perception_module}

        % To achieve the goals of this study, it was not necessary to replicate the complexity of physical sensors or to implement AI-based MP models. Rather than that, it focused on establishing a controlled and transparent representation of the MP for safety analysis. To this end, we adopted the MP module developed in \cite{molina2018enhancing}, using a bypass strategy in which the MP module directly accessed ground truth data from the simulation environment, including the real-time positions and velocities of all vehicles. In its default configuration, this perception sensor has an unlimited range and 360\textdegree coverage, effectively providing the complete state of the environment. To reproduce the directional constraints of real perception systems, this information was subsequently filtered through a virtual field of view (FOV) abstraction (Figure~\ref{fig:fig-37}). The resulting perception layer produced structured outputs equivalent to those illustrated in Figure~\ref{fig:fig-37}, allowing systematic HI and controlled experimentation. By abstracting the complexities of sensor noise and uncertainty, a reliable baseline was established to support the safety implications of hallucinations. At the same time, it preserved a foundation for integrating more sophisticated MP models in future research.

        To support a controlled and interpretable safety evaluation of perception-induced failures, this work does not attempt to replicate sensor physics (e.g., photometry, point-cloud generation) nor does it benchmark a specific AI-based perception model. Instead, a perception surrogate is implemented at the perception--control interface to provide a \chg{controlled} baseline and to enable systematic, parameterized distortion of perception outputs for HI.

\begin{figure}[h]
            \centering
            \includegraphics[scale=0.72]{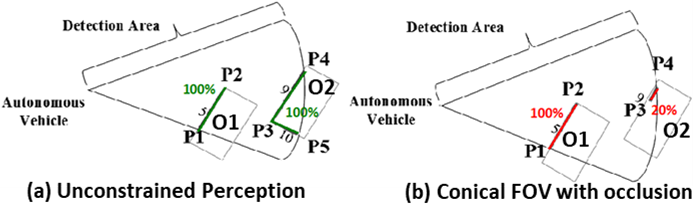}
            % \caption{Virtual field-of-view (FOV) filtering applied to simulator ground truth to produce the perception output used for HI. Adapted from~\cite{molina2018enhancing}}
            \caption{Virtual field-of-view (FOV) filtering applied to simulator ground truth. 
            (a) Object-surface segments, such as P1--P2, P3--P4, and P4--P5, visible within the AV detection area. 
            (b) Effect of occlusion on the visible portion of object surfaces. Adapted from~\cite{molina2018enhancing}.}
            \label{fig:fig-37}
        % \end{figure*}
        \end{figure}

        The MP module from~\cite{molina2018enhancing} is adopted together with a virtual field-of-view (FOV) filter to produce a more realistic perception output from simulator ground truth (Fig.~\ref{fig:fig-37}). The FOV filter restricts the objects available to MP to those located within the AV detection area and accounts for visibility constraints such as partial occlusion. As illustrated in Fig.~\ref{fig:fig-37}, object surfaces represented in the simulator ground truth are converted into visible segments according to the AV field of view; when occlusion occurs, only the visible portion of an object is retained. In the system integration shown in Fig.~\ref{fig:fig-11}, OpenDS provides the simulated environment and vehicle state, whereas MATLAB executes the MP, HI, and MC modules. \chg{The sensor-model module responsible for introducing uncertainty and error into the perception readings was enabled in all experimental conditions, including the baseline. In the baseline condition, the HI module was disabled. In HI-enabled conditions, the configured hallucination was additionally applied at the perception--control interface. Thus, the additional safety effects associated with HI can be isolated from the uncertainty already present in the perception process.}

        HI is then applied on top of this MP output. Depending on the experimental condition, HI can remove objects (missed detections), add objects (phantoms), suppress detections within a configured blind region of the FOV, perturb object states (e.g., drift or offset), and introduce delay or persistence to emulate latency and temporal inconsistency. These effects are parameterized by the HI settings (type, configuration, probability, and persistence) defined in Section~\ref{sub-sec:hallucination_injection_module} and Table~\ref{tab:failuretypes_labels}. The HI module acts at the interface between MP and MC, injecting distortions into the information sent from MP to MC rather than directly modifying the internal operation of the MP component. This design supports portability of the HI framework to alternative, potentially more realistic MP and MC implementations.  

        \chg{Using simulator ground truth as the source for the perception model, while retaining the same sensor uncertainty and error process across experimental conditions, provides a controlled basis for comparison. The baseline therefore does not represent perfect perception.} This choice does not claim sensor realism, but it isolates how perception-output distortions propagate into control decisions and safety outcomes.
        
    \subsection{Machine Control (MC) Module}

        This study uses a simplified MC module. The module receives data from MP to obtain situational awareness and continuously estimates the future occupancy of the intersection by other vehicles in order to identify CWs during which the AV can cross safely (Figure~\ref{fig:fig-23}). This approach follows the classical gap-acceptance concept in traffic engineering, in which drivers decide whether to accept a temporal gap before performing a maneuver \cite{pollatschek2002decision}. Each CW represents a predicted interval during which no other vehicle is expected to enter the intersection. Thus, MC was designed under the paradigm of autonomous intersection management, in which vehicles exploit time windows to achieve efficient and coordinated intersection passage \cite{dresner2008multiagent}.

        For each CW, the MC assesses whether the AV can reach the center of the window without violating speed limits. Infeasible CWs are discarded. Among the feasible ones, the closest valid window is selected, and the MC module computes the longitudinal control commands, namely throttle $\alpha_{th}$ and brake $\alpha_{br}$. As shown in Fig.~\ref{fig:fig-11}, these commands are generated in MATLAB and applied to the AV model in OpenDS, allowing the AV arrival time to be adjusted to match the selected CW. This strategy is consistent with reservation-based trajectory optimization methods, in which conflict-free intersection entries are planned by reserving spatio-temporal resources \cite{levin2016paradoxes}. This MC method was chosen because it offers greater interpretability and predictability than black-box models, which facilitates causal analysis of HI effects.

       The MC adopted in this study is intentionally simplified to isolate the effect of perceptual distortions on decision-making. The HI framework is independent of the specific control architecture and does not assume a particular planning or control strategy. The same injection mechanisms can therefore be applied to other MC stacks, including optimization-based planners and learning-based autonomous driving systems. In optimization-based planners (e.g., trajectory optimization and model predictive control), hallucinated perceptual distortions alter the state and constraint estimates used to compute trajectories. In learning-based systems (e.g., imitation learning, reinforcement learning, or end-to-end neural policies), the same distortions affect the perceived inputs used for action selection. In both cases, the HI framework operates at the perception--control interface, and the induced distortions affect planning and control decisions regardless of whether the control logic is model-based or data-driven.

    \subsection{Research Questions and Investigated Hypotheses} \label{sub-sec:reserach_questions_investigated_hypotheses}
        
        A set of research questions (RQs) was formulated to evaluate the HI module and quantify its implications for AV safety. Each RQ was mapped to a corresponding hypothesis, and each hypothesis was further decomposed into fine-grained testable hypotheses. For clarity and ease of reference, the hypotheses discussed below are consolidated in Table~\ref{tab:hypothesis_hierarchy}.

\setlength{\tabcolsep}{5pt}
\begin{table}[h]
\caption{Summary of granular hypotheses}
\label{tab:hypothesis_hierarchy}
\centering
\begin{threeparttable}
\begin{tabularx}{\columnwidth}{c l c c}
\toprule
\textbf{Hypothesis} & \textbf{Independent Variable} & \makecell{\textbf{Expected}\\\textbf{Influence}} & \makecell{\textbf{Dependent}\\\textbf{Variable}} \\
\midrule
$H_{1.1}$ & \chg{\textit{ModuleActivation}} & $\uparrow$$\downarrow$ & \multirow{6}{*}{\makecell{\textit{Accident}\\\chg{\textit{Occurrence}}}} \\
$H_{2.1}$ & \textit{HallucinationType} & $\uparrow$$\downarrow$ & \\
$H_{3.1}$ & \textit{AffectedDomain} & $\uparrow$$\downarrow$ & \\
$H_{4.1}$ & \textit{HallucinationConfiguration} & $\uparrow$$\downarrow$ & \\
$H_{5.1}$ & \textit{HallucinationProbability} & $\uparrow$ & \\
$H_{6.1}$ & \textit{HallucinationPersistence} & $\uparrow$$\downarrow$ & \\
\midrule
% --- Segundo Bloco de Hipóteses ---
$H_{1.2}$ & \chg{\textit{ModuleActivation}} & $\uparrow$$\downarrow$ & \multirow{6}{*}{\makecell{\textit{Minimum}\\\textit{Distance}}} \\
$H_{2.2}$ & \textit{HallucinationType} & $\uparrow$$\downarrow$ & \\
$H_{3.2}$ & \textit{AffectedDomain} & $\uparrow$$\downarrow$ & \\
$H_{4.2}$ & \textit{HallucinationConfiguration} & $\uparrow$$\downarrow$ & \\
$H_{5.2}$ & \textit{HallucinationProbability} & $\downarrow$ & \\
$H_{6.2}$ & \textit{HallucinationPersistence} & $\uparrow$$\downarrow$ & \\
\bottomrule
\end{tabularx}

\smallskip
\scriptsize
\begin{tablenotes}
    \RaggedRight
    % Nota atualizada para explicar a nova seta
    \item[]Note: An up arrow ($\uparrow$) indicates the independent variable is expected to increase the dependent variable. A down arrow ($\downarrow$) indicates the value is expected to decrease (e.g., a smaller minimum distance). A double arrow ($\uparrow$$\downarrow$) indicates that the direction of the influence (increase or decrease) is not known a priori.
\end{tablenotes}
\end{threeparttable}
\end{table}

        \subsubsection*{RQ1) Do hallucinations increase safety risk to AV operators?}

            The first hypothesis ($H_1$) states that \textit{hallucinations increase safety risk \chg{to} AV operators}. This reflects the primary assumption that the introduction of hallucinations increases the likelihood of accidents relative to the baseline (non-failure) condition. Although the literature uses diverse metrics \cite{westhofen2023critically} to operationalize safety risk, this study adopts two proxy variables: (i) accident occurrence\chg{, where an accident is a collision between the AV and another vehicle,} and (ii) minimum distance to the nearest vehicle crossing the intersection. An accident (collision) provides an unambiguous oracle for critical safety failure, whereas minimum distance is widely used as a continuous proxy for \chg{safety-margin erosion, indicating how close the AV comes to a collision} \cite{tian2022mosat, ebadi2021efficient, shimanuki2025cortex, kaufmann2021critical, kluck2023empirical}. These metrics are also used in the fitness functions of search-based testing frameworks to uncover safety-critical scenarios systematically \cite{tian2022mosat, shimanuki2025cortex, kluck2023empirical}. By decomposing safety into quantifiable and testable components, the proposed methodology is consistent with the literature on rigorous, evidence-based AV safety evaluation in diverse operational scenarios \cite{svensson2006estimating, kalra2016driving}. Thus, $H_1$ was decomposed into:

            \begin{itemize}
                \item \bm{$H_{1.1}$} Injected hallucinations influence the likelihood of an accident occurring;
                \item \bm{$H_{1.2}$} Injected hallucinations influence the minimum distance between the AV and the nearest vehicle.
            \end{itemize}

        \subsubsection*{RQ2) How do distinct types of hallucinations affect system safety?}

            The second hypothesis ($H_2$) asserts that \textit{hallucination type significantly influences safety risk}. This hypothesis evaluates whether different hallucination types produce statistically distinguishable safety outcomes. Previous work argues that ignoring distinctions among failure types can mask critical risk variations and compromise system safety \cite{borgovini1993failure}. Thus, $H_2$ was defined as:

            \begin{itemize}
                \item \bm{$H_{2.1}$}: Hallucination type influences the likelihood of an accident occurring;
                \item \bm{$H_{2.2}$}: Hallucination type influences the minimum distance between the AV and the nearest vehicle.
            \end{itemize}

        \subsubsection*{RQ3) Do hallucinations targeting distinct perception domains affect system safety differently?}

            MP spans multiple domains, including object position, object recognition, and information timing. Each supports distinct safety-critical functions. The third hypothesis ($H_3$) holds that \textit{the perception domain targeted by a hallucination influences system safety}. For example, object-position hallucinations distort trajectories, while object-recognition hallucinations can suppress yielding behavior. $H_3$ was decomposed into the following components:

            \begin{itemize}
                \item \bm{$H_{3.1}$}: The perception domain targeted by hallucinations influences the likelihood of an accident occurring;
                \item \bm{$H_{3.2}$}: The perception domain targeted by hallucinations influences the minimum distance between the AV and the nearest vehicle.
            \end{itemize}

        \subsubsection*{RQ4) How do distinct hallucination configurations influence system safety?}

            The HI module was designed to parameterize hallucinations by type and configuration. Whereas $H_2$ focuses on hallucination type, the fourth hypothesis ($H_4$) states that \textit{hallucination configuration significantly influences system safety}. Therefore, $H_4$ evaluates whether different configurations lead to measurable differences in accident likelihood and in the proximity between the AV and crossing vehicles, thereby supporting validation of the HI module. Thus, $H_4$ was decomposed into:

            \begin{itemize}
                \item \bm{$H_{4.1}$} Hallucination configuration influences the likelihood of an accident occurring;
                \item \bm{$H_{4.2}$} Hallucination configuration influences the minimum distance between the AV and the nearest vehicle.
            \end{itemize}

        \subsubsection*{RQ5) How does the probability of hallucination occurrence impact system safety?}

            Higher hallucination probabilities are expected to introduce more disturbances into MC, which may induce more frequent unsafe control reactions, affecting the AV safety. The fifth hypothesis ($H_5$) asserts that \textit{higher hallucination probability increases safety risk}. Then, $H_5$ was decomposed into:

            \begin{itemize}
                \item \bm{$H_{5.1}$} Higher hallucination probabilities increase the likelihood of an accident occurring;
                \item \bm{$H_{5.2}$} Higher hallucination probabilities reduce the minimum distance between the AV and the nearest vehicle.
            \end{itemize}

        \subsubsection*{RQ6) How does hallucination persistence over time affect system safety?}

            Finally, the sixth hypothesis ($H_6$) states that \textit{hallucination persistence (permanent vs. intermittent) significantly influences system safety outcomes}, including accident likelihood and minimum distance. Persistence determines duration: intermittent hallucinations may allow recovery, whereas permanent hallucinations may accumulate risk over time. Reliability studies identify persistence as a critical factor in safety assessment \cite{qi2008no}. $H_6$ was decomposed into the following components:

            \begin{itemize}
                \item \bm{$H_{6.1}$} Hallucination persistence influences the likelihood of an accident occurring;
                \item \bm{$H_{6.2}$} Hallucination persistence influences the minimum distance between the AV and the nearest vehicle.
            \end{itemize}

    \subsection{Experimental Procedures} \label{sub-sec:experimental_procedures}

\begin{figure}[h]
            \centering
            \includegraphics[scale=0.475]{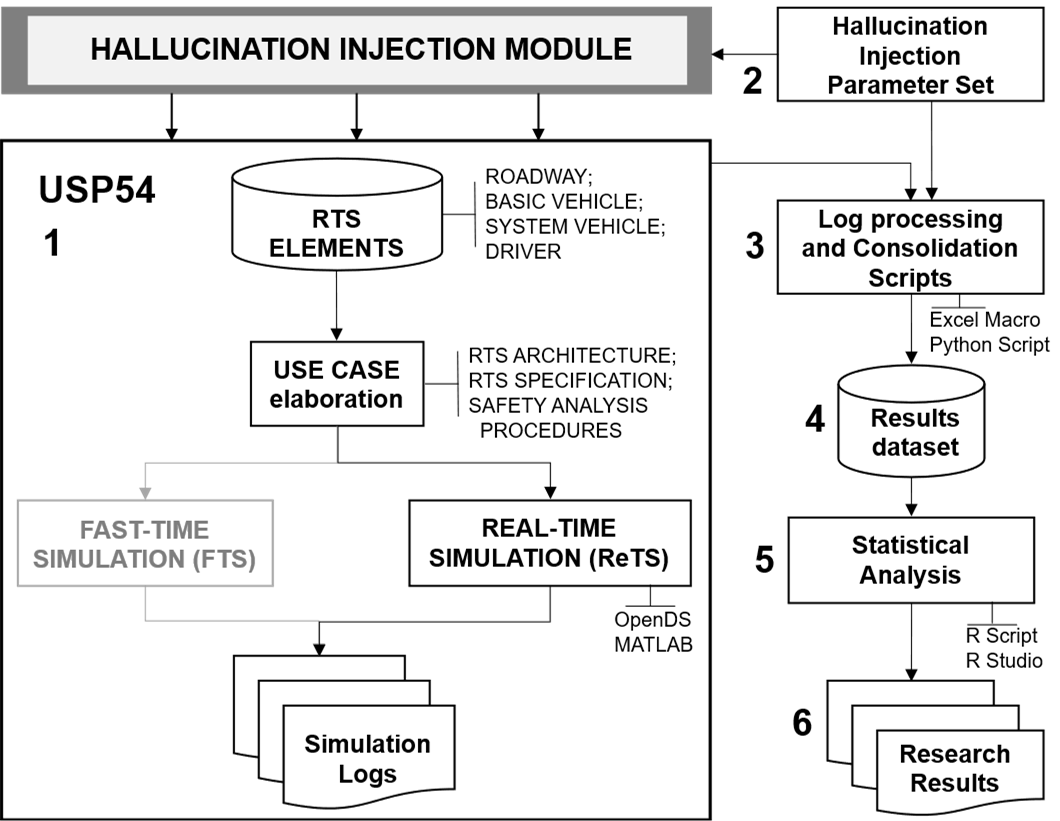}
            \caption{Elements supporting the experimental procedures}
            \label{fig:fig-24}
        \end{figure}

        The proposed HI module was evaluated using the ReTS environment (Figure~\ref{fig:fig-24}, element 1). A total of 201 distinct experimental conditions were designed. They included one baseline condition (HI module off) and 200 conditions covering all combinations of HI type, configuration, probability, and persistence defined by element 2 in Figure~\ref{fig:fig-24} (see Section \ref{sub-sec:hallucination_injection_module}). Approximately 50 simulations were executed for each of the $200$ HI conditions, totaling around $10{,}000$ runs, and approximately $9{,}000$ runs were executed for the baseline configuration.
        
        \chg{Because the sensor-model uncertainty and error module remained enabled in both baseline and HI-enabled runs, repeated baseline runs varied under the same sensor-reading conditions used in the HI experiments. These runs provided the reference distribution against which the additional effect of hallucination injection was evaluated.}

        During simulation execution, log files recorded the measurements collected for each run (Table \ref{tab:information_collected}). After invalid \chg{simulation} logs were discarded (see Section~\ref{sub-sec:experimental_use_case} for criteria), the results of $18{,}356$ valid log files ($8{,}695$ HI OFF and $9{,}661$ HI ON) were consolidated with scripts (element 3) into a dataset (element 4, Table~\ref{tab:dataset_fields_summary}) used to investigate the study hypotheses (Section~\ref{sub-sec:reserach_questions_investigated_hypotheses}). Each dataset entry corresponds to an individual simulation and includes the HI configuration used and the observed safety outcomes (Accident flag and Minimum Distance between vehicles). Finally, this dataset was imported into RStudio (element 5), where the statistical analyses detailed in Section \ref{sub-sec:analysis_procedures} were performed to test the hypotheses and generate the research results (element 6).

        % INSERT FIGURE 87

        % INSERT TABLE 17
        \setlength{\tabcolsep}{1pt}
% \begin{table*}[h]
\begin{table}[h]
\caption{Simulation log content}
\label{tab:information_collected}
\centering
% \begin{tabularx}{\textwidth}{|m{2.5cm}|m{9cm}|m{5cm}|}
% \begin{tabularx}{\textwidth}{|m{2.5cm}|m{2.8cm}|m{2.8cm}|}
% \begin{tabularx}{\textwidth}{|>{\centering\arraybackslash}m{2.5cm}|m{6.2cm}|}
% \begin{tabularx}{\textwidth}{>{\centering\arraybackslash}m{2.2cm} m{6.2cm}}
\begin{tabularx}{\textwidth}{>{\centering\arraybackslash}m{1.8cm} m{6.6cm}}
 % \toprule
 \cmidrule(){1-2}
 \textbf{Metric (unit)} & \textbf{Description} \\
\cmidrule(){1-2}
% Time (ms) & Simulation time from start of execution \\
Time (ms) & Timestamp \\
\cmidrule(){1-2}
$x_{av} [m]$, $z_{av} [m]$, $v_{av} [km/h]$ & Position and speed of AV at the moment indicated by the timestamp \\
\cmidrule(){1-2}
% Steering Wheel Position $\sigma(t)$ [-1,1] & Time-varying signal representing steering wheel actuation \\
Steering Wheel Position $\sigma(t)$ & Position of the steering wheel ranging from -1 to 1 commanded by the MC \\
\cmidrule(){1-2}
% Throttle Pedal Position $\alpha_{th}(t)$ [0,1] & Time-varying signal representing throttle pedal pressure \\
Throttle Pedal Position $\alpha_{th}(t)$ & Position of throttle pedal pressure ranging from 0 to 1. 0 indicates no actuation and 1 represents full pressure \\
\cmidrule(){1-2}
% Brake Pedal Position $\alpha_{br}(t)$ [0,1] & Time-varying signal representing brake pedal pressure \\
Brake Pedal Position $\alpha_{br}(t)$ & Position of brake pedal pressure ranging from 0 to 1. 0 indicates no actuation and 1 represents full pressure \\
\cmidrule(){1-2}
% $x_{n} [m]$, $z_{n} [m]$, $v_{n} [km/h]$ & Position and instantaneous speed of vehicle $n$, where $n=1$ to $5$ \\
$x_{n} [m]$, $z_{n} [m]$, $v_{n} [km/h]$ & Position and speed of each vehicle $n$ at the transversal street at the moment indicated by the timestamp; $n=1$ to $5$\\
\cmidrule(){1-2}
\end{tabularx}
\end{table}

        % \setlength{\tabcolsep}{1pt}
% \begin{table}[h]
% \caption{Dataset features used to verify the hypotheses $H_1$ to $H_6$}
% \label{tab:dataset_fields_summary}
% \centering
% % \begin{tabularx}{\textwidth}{|p{3.8cm}|p{4.9cm}|}
% % \begin{tabularx}{\textwidth}{p{3.8cm} p{4.9cm}}
% \begin{tabularx}{\textwidth}{p{3.4cm} p{4.9cm}}
% \cmidrule(){1-2}
% \textbf{Variable(s)} & \textbf{Description} \\
% \cmidrule(){1-2}
% Accident & Indicates whether a collision occurred in a given simulation execution \\
% \cmidrule(){1-2}
% MinimalDistance & Minimal distance between AV and nearest crossing vehicle in the intersection \\
% \cmidrule(){1-2}
% HallucinationInjected, HallucinationType, AffectedDomain, HallucinationConfiguration, HallucinationProbability, HallucinationPersistence & Characteristics and context of the injected hallucination during each simulation configuration \\
% \cmidrule(){1-2}
% \end{tabularx}
% \end{table}

\begin{table}[h]
\centering
\begin{threeparttable}
\caption{Dataset Features Used to Verify Hypotheses $H_1$ to $H_6$}
\label{tab:dataset_fields_summary}
\begin{tabularx}{\columnwidth}{l X}
\toprule
\textbf{Variable} & \textbf{Description} \\
\midrule
\textit{Accident} & Indicates whether a collision occurred in a given simulation run. \\
\addlinespace
\textit{MinimalDistance} & The minimum distance between the AV and the nearest crossing vehicle at the intersection. \\
\addlinespace
HI Variables\tnote{*} & Define the characteristics and context of the injected hallucination for each run. \\
\bottomrule
\end{tabularx}
\begin{tablenotes}
    \item[*] Includes \textit{ModuleActivation}, \textit{HallucinationType}, \textit{AffectedDomain}, \textit{HallucinationConfiguration}, \textit{HallucinationProbability}, and \textit{HallucinationPersistence}.
\end{tablenotes}
\end{threeparttable}
\end{table}

% \setlength{\tabcolsep}{1pt}
% \begin{table}[h]
% \caption{Dataset fields consolidated per tested configuration}
% \label{tab:dataset_fields_summary}
% \centering
% \begin{tabularx}{\textwidth}{|p{3.8cm}|p{4.9cm}|}
% \cmidrule(){1-2}
% \textbf{Variable(s)} & \textbf{Description} \\
% \cmidrule(){1-2}
% AccidentRate & Accident rate across all executions with a unique set of simulation configurations \\
% \cmidrule(){1-2}
% ShortestDistanceVehicles & Minimal distance between AV and nearest crossing vehicle in the intersection \\
% \cmidrule(){1-2}
% FailureType, DimensionAffected, FailureConfiguration, FailureProbability, Scenario & Characteristics and context of the injected failure during each simulation configuration \\
% \cmidrule(){1-2}
% DiversityArchitecture, NumberSensors, VotingCriteria & Settings related to the AV's sensor diversity and fusion strategy \\
% \cmidrule(){1-2}
% SpeedMean, SpeedSD & Mean and standard deviation of the AV's speed during the simulation \\
% \cmidrule(){1-2}
% MsgCount & Number of messages sent by Machine Control to actuators \\
% \cmidrule(){1-2}
% GasPedalCount, GasPedalSum, BrakePedalCount, BrakePedalSum & Frequency and intensity of gas and brake pedal activations commanded by Machine Control \\
% \cmidrule(){1-2}
% EmmergencyStopRate & Flag indicating whether an emergency stop was commanded \\
% \cmidrule(){1-2}
% FlipCount & Number of times control switched between gas and brake pedals \\
% \cmidrule(){1-2}
% \end{tabularx}
% \end{table}

        For study replication, Appendix~\ref{app:repro_guideline} provides a general guideline and a checklist of the settings used throughout the experiments.

    \subsection{Analysis Procedures} \label{sub-sec:analysis_procedures}
    
\begin{figure}[h]
            \centering
            \chgfig{\includegraphics[width=239pt]{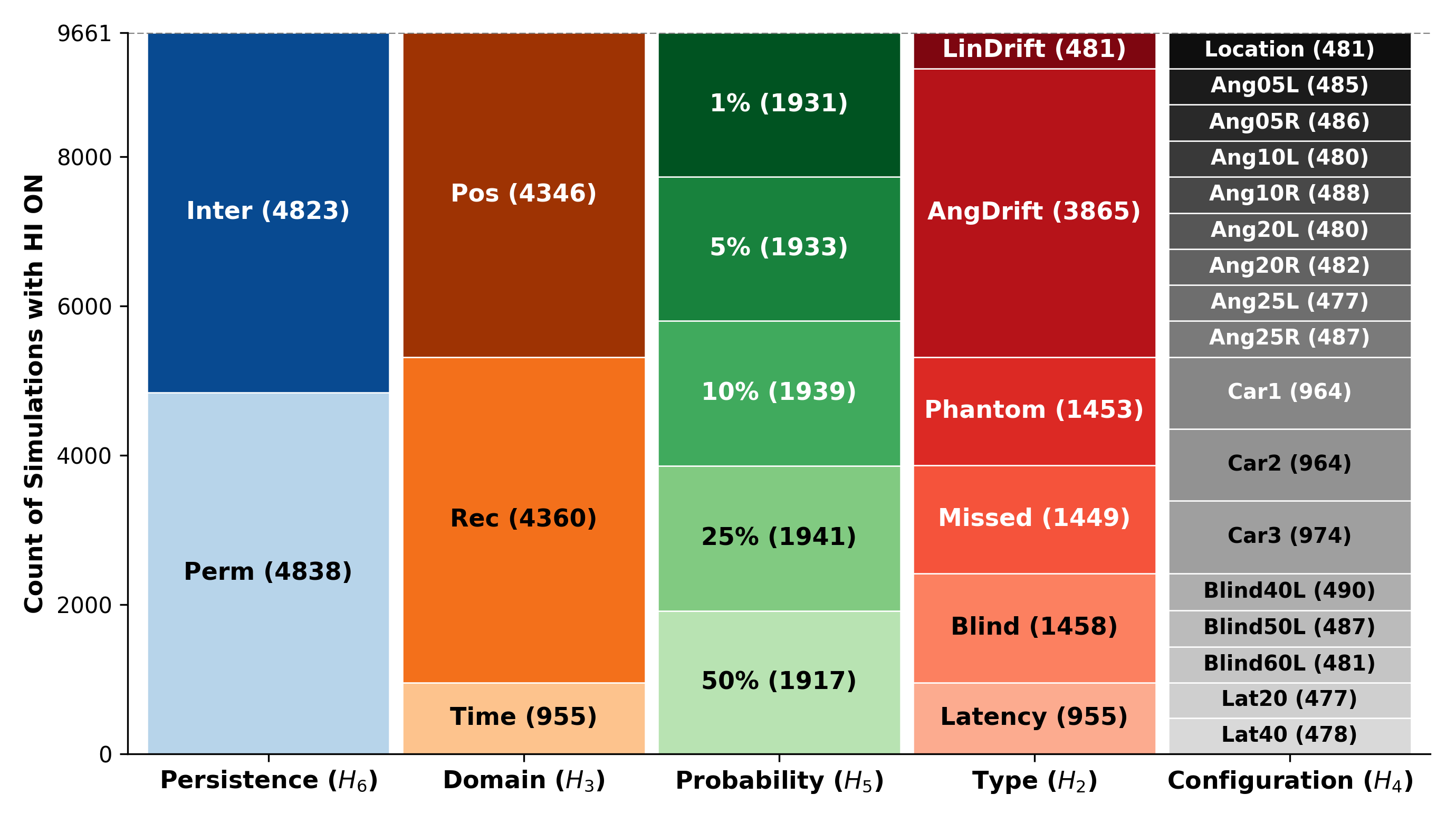}}
            \caption{Breakdown of sample size by category of each HI property}
            \label{fig:N_amostral}
        \end{figure}

        A combination of statistical approaches was used to test the hypotheses. An adequate sample size is fundamental for investigating each hypothesis. Although the total sample size was $18{,}356$, which ensured a robust evaluation of $H_1$, an analysis was performed to assess the representativeness of each condition category evaluated in $H_2$ to $H_6$. Figure~\ref{fig:N_amostral} illustrates the number of valid simulations consolidated for each category of each HI-module variable when HI was ON. A robust sample size ($N \geq 450$) was achieved for all conditions evaluated in $H_2$ to $H_6$, enabling rigorous comparison with the baseline.
        
        To evaluate the impact on accident probability (collision occurrence), logistic regression with binomial distribution was used, since Accident is a binary variable. This model is appropriate for binary outcomes because it estimates how predictors affect the odds of a collision occurring \cite{montgomery2021introduction}. Thus, logistic regression was used to test the hypotheses $H_{1.1}$, $H_{2.1}$, $H_{3.1}$, $H_{4.1}$, $H_{5.1}$, $H_{6.1}$.

        % Thus, logistic regression was used to test the hypotheses of $H_{1.1}$, $H_{1.2}$, $H_{2.1}$, $H_{2.2}$, $H_{3.1}$, $H_{3.2}$, $H_{4.1}$, $H_{4.2}$, $H_{5.1}$, $H_{5.2}$, $H_{6.1}$, $H_{6.2}$.

        In contrast, for the continuous Minimum Distance variable, a standard linear regression model was applied. This approach provides clear estimates of the direction and magnitude of association, allowing direct evaluation of how different factors influence the \chg{minimum distance} between vehicles \cite{montgomery2021introduction}. Thus, linear regression was used to test the hypotheses $H_{1.2}$, $H_{2.2}$, $H_{3.2}$, $H_{4.2}$, $H_{5.2}$, $H_{6.2}$.

        % Thus, linear regression was used to test the hypotheses of $H_{1.1}$, $H_{1.2}$, $H_{2.1}$, $H_{2.2}$, $H_{3.1}$, $H_{3.2}$, $H_{4.1}$, $H_{4.2}$, $H_{5.1}$, $H_{5.2}$, $H_{6.1}$, $H_{6.2}$.

        To evaluate the impact of categorical predictors, an Analysis of Variance (ANOVA) was performed on the models to assess the overall significance of each factor on safety risk \cite{o2012practical}. This approach is more informative than examining individual regression coefficients alone, which are limited to comparisons against a reference category.

        All statistical analyses were performed in RStudio, using established libraries for regression modeling, ANOVA, and hypothesis testing \cite{r2021studio}. In all analyses, the significance threshold was set at \chg{$\alpha = 0.05$,} consistent with standard practice in safety and reliability research \cite{o2012practical, kutner2025applied}.

    \subsection{Experimental Hardware Infrastructure} \label{sub-sec:experimental_hardware_infrastructure}

        The experiments were executed on two Windows 10 laptops. Simulation runs and log processing were performed on an Acer laptop equipped with an Intel Core i7 processor, 8 GB of RAM, and an NVIDIA GeForce MX150 GPU. Statistical analyses in RStudio were performed on a second Windows 10 laptop.
        
        % The experiments were performed using a consumer-grade laptop equipped with an Intel Core i7 processor, 8 GB of RAM, and NVIDIA GPU GeForce MX150. The same computer was used to execute Python scripts to process the logs. Another similar laptop was used to run R Studio to perform the statistical analysis.

% \section{Results} \label{sec:results}
\section{Analysis of Results and Hypothesis Evaluation} \label{sec:results}

    This section evaluates the effects of HI on AV safety according to the hypotheses defined in Section~\ref{sub-sec:reserach_questions_investigated_hypotheses}. The results are organized by hypothesis, beginning with the overall effect of HI and then examining the effects of individual HI properties on accident likelihood and minimum distance.

    \textbf{The effect of HI on system safety} (\bm{$H_1$}) (Tables \ref{tab:anova_hi_accident_probability} and \ref{tab:anova_hi_minimum_distance}, row 1, and Tables \ref{tab:odds_ratio_hallucination_injected} and \ref{tab:regression_hallucination_injected}). Injecting hallucinations had a significant effect on AV safety. First, accident likelihood increased significantly when hallucinations were injected (HI \textit{ON}) compared to the baseline (HI \textit{OFF}) (\chg{LR} $\chi^2 = 126.7, p < 0.001$, Table~\ref{tab:anova_hi_accident_probability}). The estimated odds of an accident were $3.09$ times the baseline odds ($p < 0.001$) when HI was \chg{\textit{ON},} \chg{compared} \chg{with} HI \chg{\textit{OFF},} demonstrating the effectiveness of the HI module in stressing the system (Table \ref{tab:odds_ratio_hallucination_injected}). Thus, $H_{1.1}$ was accepted.
    
    Moreover, the minimum distance between the AV and the \chg{nearest} vehicle during the crossing was significantly affected by HI \chg{($F(1, 18354) = 6{,}989$}, $p<.001$, partial $\eta^2=0.28$, Table~\ref{tab:anova_hi_minimum_distance}). Based on the linear model, the baseline minimum distance was estimated at $8.62$\,m, whereas HI reduced this distance by $1.60$\,m, resulting in an estimated distance of $7.02$\,m. This corresponds to an $18.6\%$ reduction relative to the baseline (Table~\ref{tab:regression_hallucination_injected}). Therefore, $H_1$ was accepted, since $H_{1.1}$ and $H_{1.2}$ were both accepted. 

    % FROM APPENDIX SUPPLEMENTARY STATISTIC TABLES - ANOVA
    \setlength{\tabcolsep}{5pt} % Ajuste o espaçamento entre as colunas conforme necessário
% \begin{table}[h]
\begin{table}[H]
\caption{ANOVA results for the effects of HI properties on \chg{Accident Occurrence}}
\label{tab:anova_hi_accident_probability}
\centering
\scriptsize
\begin{threeparttable}
\begin{tabular}{lccc}
\cmidrule{1-4}
HI Properties & LR $\chi^2$ & Df & p \\
\cmidrule{1-4}
$H_{1.1}$: Module Activation & 126.7 & 1 & < .001 \\
$H_{2.1}$: Hallucination Type & 186.29 & 6 & < .001 \\
\chg{$H_{3.1}$: Affected Domain} & \chg{158.15} & \chg{3} & \chg{< .001} \\
$H_{4.1}$: Hallucination Configuration & 369.72 & 17 & < .001 \\
$H_{5.1}$: Hallucination Probability & 385.43 & 5 & < .001 \\
$H_{6.1}$: Hallucination Persistence & 148.59 & 2 & < .001 \\
\cmidrule{1-4}
\end{tabular}
\smallskip
\scriptsize
\begin{tablenotes}
\item[]Note: LR $\chi^2$: Likelihood Ratio Chi-Squared, Df: Degrees of freedom.
\end{tablenotes}
\end{threeparttable}
\end{table}

\setlength{\tabcolsep}{2pt} % Minimized column spacing
\begin{table}[H]
\caption{ANOVA results for the effects of HI properties on Minimum Distance}
\label{tab:anova_hi_minimum_distance}
\centering
\scriptsize
\begin{threeparttable}
\begin{tabular}{l r c r r c c c}
\cmidrule{1-8}
Predictor & SS & df & MS & F & p & $\eta_p^2$ & 90\% CI [$\eta_p^2$]\\
\cmidrule{1-8}
$H_{1.2}$: Module Activation  & 11,785 & 1  & 11,785 & 6,989 & <.001 & 0.28 & [0.27, 0.28] \\
$H_{2.2}$: Hallucination Type  & 12,206 & 6  & 2,034 & 1,223 & <.001 & 0.29 & [0.28, 0.29] \\
\chg{$H_{3.2}$: Affected Domain} & \chg{12,057} & \chg{3} & \chg{4,019} & \chg{2,404} & \chg{<.001} & \chg{0.28} & \chg{[0.27, 0.29]} \\
$H_{4.2}$: Hallucination Configuration  & 12,292 & 17 & 723  & 436  & <.001 & 0.29 & [0.28, 0.30] \\
$H_{5.2}$: Hallucination Probability & 11,856 & 5  & 2,371 & 1,409 & <.001 & 0.28 & [0.27, 0.29] \\
$H_{6.2}$: Hallucination Persistence & 11,936 & 2  & 5,968 & 3,556 & <.001 & 0.28 & [0.27, 0.29] \\
\cmidrule{1-8}
\end{tabular}
\smallskip
\scriptsize
\begin{tablenotes}
\item[]Note: SS: Sum of Squares, df: Degrees of Freedom, MS: Mean Square, F: F-statistic, p: p-value, \chg{$\eta_p^2$}: partial $\eta^2$, \chg{$\eta_p^2$} 90\%: partial $\eta^2$ 90\% CI [LL, UL], CI: Confidence Interval, LL: Lower Limit, UL: Upper Limit
\end{tablenotes}
\end{threeparttable}
\end{table}

    % FROM APPENDIX SUPPLEMENTARY STATISTIC TABLES - ODDS RATIO
    \setlength{\tabcolsep}{2.5pt}
% \begin{table}[h]
\begin{table}[H]
\caption{Odds ratio results for predictor Module Activation (Hypothesis $H_{1.1}$)}
\label{tab:odds_ratio_hallucination_injected}
\centering
\scriptsize
\begin{threeparttable}
\begin{tabular}{lccccc}
\cmidrule{1-6}
Parameter & Odds Ratio & SE & 95\% CI & z & p \\
\cmidrule{1-6}
(Intercept) & 0.01 & $1.24 \times 10^{-3}$ & [0.01, 0.02] & -45.62 & < .001 \\
\chg{ModuleActivation [ON]} & 3.09 & 0.34 & [2.51, 3.85] & 10.39 & < .001 \\
\cmidrule{1-6}
\end{tabular}
\smallskip
\scriptsize
\begin{tablenotes}
\item[]Note: SE: Standard Error, CI: Confidence Interval. The intercept is the exponentiated logistic-regression intercept and represents the estimated baseline odds of an accident when HI is \textit{OFF}. The odds ratio for \textit{\chg{ModuleActivation [ON]}} represents the multiplicative change in accident odds when HI is \textit{ON} relative to HI \textit{OFF}.
\end{tablenotes}
\end{threeparttable}
\end{table}

    % FROM APPENDIX SUPPLEMENTARY STATISTIC TABLES - MINIMUM DISTANCE
    \setlength{\tabcolsep}{3pt}
% \begin{table}[h]
\begin{table}[H]
\caption{Linear model results for predictor Module Activation (Hypothesis $H_{1.2}$)}
\label{tab:regression_hallucination_injected}
\centering
\scriptsize
\begin{threeparttable}
\begin{tabular}{lccccc}
\cmidrule{1-6}
Parameter & Coefficient & SE & 95\% CI & t(18354) & p \\
\cmidrule{1-6}
(Intercept) & 8.62 & 0.01 & [8.59, 8.65] & 619.12 & < .001 \\
\chg{ModuleActivation [ON]} & -1.60 & 0.02 & [-1.64, -1.57] & -83.60 & < .001 \\
\cmidrule{1-6}
\end{tabular}
\smallskip
\scriptsize
\begin{tablenotes}
\item[]Note: SE: Standard Error, CI: Confidence Interval. The intercept represents the estimated minimum distance in the baseline condition, when HI is \textit{OFF}. The coefficient for \textit{\chg{ModuleActivation [ON]}} represents the estimated change relative to this baseline.
\end{tablenotes}
\end{threeparttable}
\end{table}

    \textbf{The effect of hallucination type on system safety} (\bm{$H_2$}) (Tables \ref{tab:anova_hi_accident_probability} and \ref{tab:anova_hi_minimum_distance} row 2, and Tables \ref{tab:odds_ratio_hallucination_type} and \ref{tab:regression_hallucination_type}). At a higher level of abstraction, the HI property Type, considered as a single construct, significantly affected AV safety when hallucinations were injected. Specifically, it significantly affected accident likelihood (\chg{LR} $\chi^2 = 186.29, p < 0.001$, Table~\ref{tab:anova_hi_accident_probability}) ($H_{2.1}$ accepted) and the minimum distance between the AV and the \chg{nearest} vehicle during the crossing \chg{($F(6, 18349) = 1{,}223$}, $p<.001$, partial $\eta^2=0.29$, Table \ref{tab:anova_hi_minimum_distance}) ($H_{2.2}$ accepted). Therefore, $H_2$ was accepted.

\begin{figure*}[h]
        \centering
        \chgfig{\includegraphics[width=\textwidth]{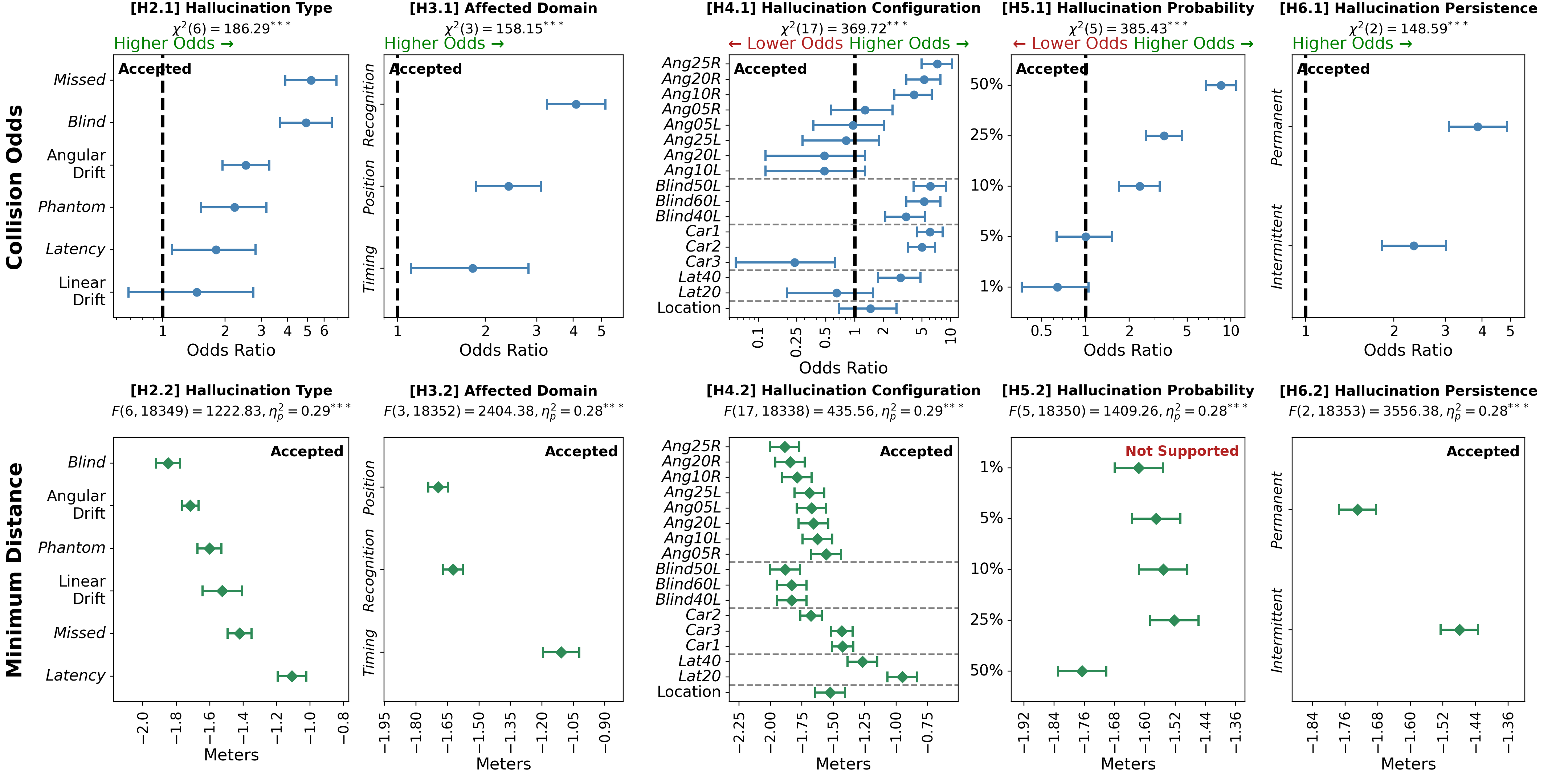}}
        \caption{Testing results of supporting hypotheses \chg{($H_2$--$H_6$)}. Five HI properties (Type, Domain, Configuration, Probability, Persistence) were tested on two variables. The top row shows odds ratios from logistic regression for \chg{\textit{Collision Odds}}, and the bottom row shows linear regression coefficients ($\beta$) for \textit{Minimum Distance}, where negative values indicate a smaller \chg{minimum distance}. Error bars show $95\%$ confidence intervals}

        \begin{minipage}{\textwidth} % Set the width of the note to match the image
            \footnotesize Note. For details on $H_4$, see Table~\ref{tab:failureconfiguration_labels}. $^{***}$ denotes significance level $<0.001$. 
        \end{minipage}
        \label{fig:hypothesis_summary}
    \end{figure*}

    However, not all hallucination types were equally dangerous. Regarding accident likelihood, as shown by chart $H_{2.1}$ in Figure \ref{fig:hypothesis_summary} and \chg{Table~\ref{tab:odds_ratio_hallucination_type}}, \textit{Missed Detection} ($OR = 5.20$, $p < 0.001$) and \textit{Blind Region} ($OR = 4.92$, $p < 0.001$) were the most critical hallucination types, increasing the odds of a collision approximately five times. They were followed by \textit{Perception Angular Drift} ($OR = 2.52$, $p < 0.001$), \textit{Phantom Perception} ($OR = 2.23$, $p < 0.001$), and \textit{Perception Latency} ($OR = 1.81$, $p = 0.012$), which increased the odds of collision by approximately $2.5$, $2.2$, and $1.8$ times, respectively. In contrast, \textit{Perception Linear Drift}, considered as a single category, was the only hallucination type whose effect on \chg{the odds of an accident} was not statistically significant ($p = 0.278$), although \chg{its point estimate corresponded to} $46\%$ \chg{higher} odds of collision. \chg{The wide 95\% confidence interval of this estimate (chart $H_{2.1}$ in Figure~\ref{fig:hypothesis_summary}) indicates that its effect on collision odds could not be precisely estimated in the tested scenario.}

    % FROM APPENDIX SUPPLEMENTARY STATISTIC TABLES - ODDS RATIO
    \setlength{\tabcolsep}{3pt}
% \begin{table}[h]
\begin{table}[H]
% \caption{$H_{2.1}$ - Odds Ratio of Failure Type}
\caption{Odds ratio results for predictor Hallucination Type (Hypothesis $H_{2.1}$)}
\label{tab:odds_ratio_hallucination_type}
\centering
\scriptsize
\begin{threeparttable}
\begin{tabular}{lccccc}
\cmidrule{1-6}
Parameter & Odds Ratio & SE & 95\% CI & z & p \\
\cmidrule{1-6}
(Intercept) & 0.01 & $1.24 \times 10^{-3}$ & [0.01, 0.02] & -45.62 & < .001 \\
Linear Drift & 1.46 & 0.51 & [0.68, 2.74] & 1.08 & 0.278 \\
Phantom & 2.23 & 0.41 & [1.53, 3.17] & 4.33 & < .001 \\
Missed Detection & 5.20 & 0.75 & [3.91, 6.88] & 11.47 & < .001 \\
Latency & 1.81 & 0.43 & [1.11, 2.81] & 2.51 & 0.012 \\
Angular Drift & 2.52 & 0.33 & [1.94, 3.27] & 7.00 & < .001 \\
Blind Region & 4.92 & 0.72 & [3.69, 6.54] & 10.96 & < .001 \\
\cmidrule{1-6}
\end{tabular}
\smallskip
\scriptsize
\begin{tablenotes}
    \RaggedRight
    % \item[]* Type: C = Categoric, N = Numeric, B = Binary
    \item[]Note: SE: Standard Error, CI: Confidence Interval
\end{tablenotes}
\end{threeparttable}
\end{table}

    Regarding the minimum distance between the AV and the \chg{nearest} vehicle while crossing the transversal street, as shown in the bottom-row Type chart $H_{2.2}$ of Figure \ref{fig:hypothesis_summary} and Table \ref{tab:regression_hallucination_type}, the \textit{Blind Region} ($p < 0.001$) and \textit{Perception Angular Drift} ($p < 0.001$) caused the largest reductions in \chg{minimum distance}, corresponding to average decreases of \chg{$1.85$\,m} and \chg{$1.72$\,m,} respectively. They were followed by significant reductions of \chg{$1.60$\,m} and \chg{$1.52$\,m} caused by \textit{Phantom Perception} ($p < 0.001$) and \textit{Perception Linear Drift} ($p < 0.001$), respectively. That result shows that \textit{Perception Linear Drift} also significantly stressed system safety. Finally, \textit{Missed Detection} ($p < 0.001$) and \textit{Perception Latency} ($p < 0.001$) caused significant reductions in minimum distance of $1.42$\,m and $1.11$\,m, respectively.

    % FROM APPENDIX SUPPLEMENTARY STATISTIC TABLES - MINIMUM DISTANCE
    \setlength{\tabcolsep}{3pt}
% \begin{table}[h]
\begin{table}[H]
\caption{Linear model results for predictor Hallucination Type (Hypothesis $H_{2.2}$)}
\label{tab:regression_hallucination_type}
\centering
\scriptsize
\begin{threeparttable}
\begin{tabular}{lccccc}
\cmidrule{1-6}
Parameter & Coefficient & SE & 95\% CI & t(18349) & p \\
\cmidrule{1-6}
(Intercept) & 8.62 & 0.01 & [8.59, 8.65] & 623.29 & < .001 \\
Linear Drift & -1.52 & 0.06 & [-1.64, -1.41] & -25.24 & < .001 \\
Phantom & -1.60 & 0.04 & [-1.67, -1.53] & -43.80 & < .001 \\
Missed Detection & -1.42 & 0.04 & [-1.49, -1.35] & -38.83 & < .001 \\
Latency & -1.11 & 0.04 & [-1.19, -1.02] & -25.21 & < .001 \\
Angular Drift & -1.72 & 0.02 & [-1.76, -1.67] & -68.80 & < .001 \\
Blind Region & -1.85 & 0.04 & [-1.92, -1.78] & -50.63 & < .001 \\
\cmidrule{1-6}
\end{tabular}
\smallskip
\scriptsize
\begin{tablenotes}
\item[]Note: SE: Standard Error, CI: Confidence Interval, $t(18349)$ = t-statistic with 18349 degrees of freedom
\end{tablenotes}
\end{threeparttable}
\end{table}

    \textbf{The effect of the hallucination-affected domain on system safety} (\bm{$H_3$}) (Tables \ref{tab:anova_hi_accident_probability} and \ref{tab:anova_hi_minimum_distance} row 3, and Tables \ref{tab:odds_ratio_affected_domain} and \ref{tab:regression_affected_domain}). The domain affected by the hallucination had a significant impact on system safety when analyzed as a single HI property. Specifically, it significantly affected accident likelihood (\chg{LR} \chg{$\chi^2 = 158.15, p < 0.001$}, Table~\ref{tab:anova_hi_accident_probability}) ($H_{3.1}$ accepted) and the minimum distance between the AV and the \chg{nearest} vehicle during the crossing (\chg{$F(3, 18352) = 2{,}404$}, $p<.001$, partial $\eta^2=0.28$, Table \ref{tab:anova_hi_minimum_distance}) ($H_{3.2}$ accepted). Therefore, $H_3$ was accepted.

    The analysis by domain category showed that all targeted domains had a significant impact on system safety. The object-recognition ($p < 0.001$), object-position ($p < 0.001$), and information-timing ($p = 0.012$) domains significantly increased the odds of an accident by \chg{$4.09$, $2.40$}, and $1.81$ times, respectively (chart $H_{3.1}$ in Figure~\ref{fig:hypothesis_summary} and Table \ref{tab:odds_ratio_affected_domain}). Moreover, the object-position ($p < 0.001$), object-recognition ($p < 0.001$), and information-timing ($p < 0.001$) domains significantly reduced the minimum distance between the AV and the \chg{nearest} vehicle during the crossing by \chg{$1.69$\,m, $1.62$\,m}, and \chg{$1.11$\,m} on average, respectively (chart $H_{3.2}$ in Figure \ref{fig:hypothesis_summary} and Table \ref{tab:regression_affected_domain}).
    
    % FROM APPENDIX SUPPLEMENTARY STATISTIC TABLES - ODDS RATIO
    \setlength{\tabcolsep}{3pt}
% \begin{table}[h]
\begin{table}[H]
\caption{Odds ratio results for predictor Affected Domain (Hypothesis $H_{3.1}$)}
\label{tab:odds_ratio_affected_domain}
\centering
\scriptsize
\begin{threeparttable}
\begin{tabular}{lccccc}
\cmidrule{1-6}
Parameter & Odds Ratio & SE & 95\% CI & z & p \\
\cmidrule{1-6}
(Intercept) & 0.01 & $1.24 \times 10^{-3}$ & [0.01, 0.02] & -45.62 & < .001 \\
\chg{Position} & \chg{2.40} & \chg{0.31} & \chg{[1.86, 3.10]} & \chg{6.74} & \chg{< .001} \\
\chg{Recognition} & \chg{4.09} & \chg{0.48} & \chg{[3.26, 5.17]} & \chg{11.99} & \chg{< .001} \\
Timing & 1.81 & 0.43 & [1.11, 2.81] & 2.51 & 0.012 \\
\cmidrule{1-6}
\end{tabular}
\smallskip
\scriptsize
\begin{tablenotes}
\item[]Note: SE: Standard Error, CI: Confidence Interval
\end{tablenotes}
\end{threeparttable}
\end{table}

    % FROM APPENDIX SUPPLEMENTARY STATISTIC TABLES - MINIMUM DISTANCE
    \setlength{\tabcolsep}{3pt}
% \begin{table}[h]
\begin{table}[H]
\caption{Linear model results for predictor Affected Domain (Hypothesis $H_{3.2}$)}
\label{tab:regression_affected_domain}
\centering
\scriptsize
\begin{threeparttable}
\begin{tabular}{lccccc}
\cmidrule{1-6}
Parameter & Coefficient & SE & 95\% CI & t(18352) & p \\
\cmidrule{1-6}
\chg{(Intercept)} & \chg{8.62} & \chg{0.01} & \chg{[8.59, 8.65]} & \chg{621.82} & \chg{< .001} \\
\chg{Position} & \chg{-1.69} & \chg{0.02} & \chg{[-1.74, -1.65]} & \chg{-70.55} & \chg{< .001} \\
\chg{Recognition} & \chg{-1.62} & \chg{0.02} & \chg{[-1.67, -1.58]} & \chg{-67.69} & \chg{< .001} \\
\chg{Timing} & \chg{-1.11} & \chg{0.04} & \chg{[-1.19, -1.02]} & \chg{-25.15} & \chg{< .001} \\
\cmidrule{1-6}
\end{tabular}
\smallskip
\scriptsize
\begin{tablenotes}
\item[]Note: SE: Standard Error, CI: Confidence Interval, $t(18352)$ = t-statistic with 18352 degrees of freedom
\end{tablenotes}
\end{threeparttable}
\end{table}

% \setlength{\tabcolsep}{2pt}
% \begin{table}[h]
% \caption{Regression analysis of affected domain over accident rate and minimum distance}
% \label{tab:regression_affected_domain}
% \centering

% \begin{threeparttable}
% % \small
% \scriptsize
% \begin{tabular}{clccccc}
% \cmidrule{1-7}
% Variable & Parameter & Coefficient & SE & 95\% CI [LL, UL] & $t(197)$ & $p$ \\
% % \hline
% \cmidrule{1-7}
% \multirow{4}{*}{\rotatebox[origin=c]{90}{\parbox{1cm}{\centering Accident Rate}}}& (Intercept) & $6.54e^{-16}$ & 0.08 & [-0.15, 0.15] & $8.70e^{-15}$ & $>0.999$ \\
% & ObjectPosition   & 0.04 & 0.08 & [-0.11, 0.19] & 0.50 & 0.620 \\
% & ObjectRecognition & 0.04 & 0.08 & [-0.10, 0.19] & 0.59 & 0.558 \\
% & InformationTiming & 0.02 & 0.08 & [-0.13, 0.17] & 0.29 & 0.771 \\
% % \hline
% \cmidrule{1-7}

% \multirow{4}{*}{\rotatebox[origin=c]{90}{\parbox{1cm}{\centering Minimum Distance}}}& (Intercept) & 7.80 & 0.46 & [6.90, 8.70] & 17.05 & $< 0.001$ \\
% & ObjectPosition   & -0.93 & 0.46 & [-1.83, -0.02] & -2.02 & 0.045 \\
% & ObjectRecognition & -0.74 & 0.46 & [-1.65, 0.17] & -1.61 & 0.109 \\
% & InformationTiming & -0.37 & 0.47 & [-1.29, 0.56] & -0.78 & 0.436 \\

% \cmidrule{1-7}
% \end{tabular}
% \smallskip
% \scriptsize
% \begin{tablenotes}
%     \RaggedRight
%     % \item[]* Type: C = Categoric, N = Numeric, B = Binary
%     \item[]Note: SE = standard error; 95\% CI = 95\% confidence interval [lower limit, upper limit]; $t(197)$ = t-statistic with 197 degrees of freedom; $p$ = significance level for the hypothesis test that the coefficient equals zero
% \end{tablenotes}
% \end{threeparttable}
% \end{table}

    \textbf{The effect of hallucination configuration on system safety} (\bm{$H_4$}) (Tables \ref{tab:anova_hi_accident_probability} and \ref{tab:anova_hi_minimum_distance}, row 4, and Tables \ref{tab:odds_ratio_hallucination_config} and \ref{tab:regression_hallucination_configuration}). The HI property Configuration had a significant impact on system safety when analyzed as a single construct. Specifically, it significantly affected accident likelihood (\chg{LR} $\chi^2 = 369.72, p < 0.001$, Table~\ref{tab:anova_hi_accident_probability}) ($H_{4.1}$ accepted) and the minimum distance between the AV and the \chg{nearest} vehicle during the crossing ($F(17, 18338) = 436$, $p<0.001$, partial $\eta^2=0.29$, Table~\ref{tab:anova_hi_minimum_distance}) ($H_{4.2}$ accepted). Compared with the baseline (HI OFF), the estimated odds of an accident ranged from $0.24$ ($p = 0.014$\chg{; a reduction relative to the baseline}), when \chg{car 3 was the target of the \textit{Missed Detection} or \textit{Phantom Perception} hallucination}, to $7.23$ ($p < 0.001$), when a 25\textdegree\ angular drift to the right was applied to miscalibrate MP perception (Table~\ref{tab:odds_ratio_hallucination_config}). Likewise, the minimum distance between the AV and the \chg{nearest} vehicle was reduced by values ranging from $0.95$\,m to $1.89$\,m on average relative to the baseline, depending on the configuration (Table~\ref{tab:regression_hallucination_configuration}). Therefore, $H_4$ was accepted. Although Configuration is an abstract construct that encompasses distinct settings associated with different hallucination types, its significant effect on safety supports the internal consistency of the HI properties in stressing the AV system.

    % FROM APPENDIX SUPPLEMENTARY STATISTIC TABLES - ODDS RATIO
    \setlength{\tabcolsep}{3pt}
% \begin{table}[h]
\begin{table}[H]
\caption{Odds ratio results for predictor Hallucination Configuration (Hypothesis $H_{4.1}$)}
\label{tab:odds_ratio_hallucination_config}
\centering
\scriptsize
\begin{threeparttable}
\begin{tabular}{lccccc}
\cmidrule{1-6}
Parameter & Odds Ratio & SE & 95\% CI & z & p \\
\cmidrule{1-6}
% (Intercept) & 0.01 & 1.22e-03 & [0.01, 0.02] & -47.15 & < .001 \\
% Car1 & 5.95 & 0.91 & [4.39, 8.02] & 11.61 & < .001 \\
% Car2 & 4.88 & 0.79 & [3.53, 6.68] & 9.75 & < .001 \\
% Car3 & 0.23 & 0.14 & [0.06, 0.61] & -2.50 & 0.012 \\
% Ang05L & 0.94 & 0.39 & [0.37, 1.96] & -0.15 & 0.878 \\
% Ang05R & 1.25 & 0.46 & [0.56, 2.41] & 0.61 & 0.541 \\
% Ang10L & 0.47 & 0.28 & [0.12, 1.25] & -1.29 & 0.199 \\
% Ang10R & 4.04 & 0.91 & [2.55, 6.17] & 6.21 & < .001 \\
% Ang20L & 0.47 & 0.28 & [0.12, 1.25] & -1.29 & 0.199 \\
% Ang20R & 5.14 & 1.06 & [3.37, 7.62] & 7.91 & < .001 \\
% Ang25L & 0.79 & 0.36 & [0.28, 1.76] & -0.51 & 0.613 \\
% Ang25R & 7.06 & 1.31 & [4.86, 10.08] & 10.54 & < .001 \\
% Blind40L & 3.35 & 0.81 & [2.03, 5.26] & 5.02 & < .001 \\
% Blind50L & 5.97 & 1.17 & [4.02, 8.68] & 9.12 & < .001 \\
% Blind60L & 5.16 & 1.07 & [3.38, 7.63] & 7.92 & < .001 \\
% Lat20 & 0.63 & 0.32 & [0.19, 1.51] & -0.90 & 0.370 \\
% Lat40 & 2.93 & 0.75 & [1.71, 4.72] & 4.18 & < .001 \\

(Intercept) & 0.01 & $1.24 \times 10^{-3}$ & [0.01, 0.02] & -45.62 & < .001 \\
Location    & 1.46 & 0.51     & [0.68, 2.74] & 1.08   & 0.278  \\
Car1        & 6.09 & 0.95     & [4.47, 8.25] & 11.60  & < .001 \\
Car2        & 5.00 & 0.82     & [3.60, 6.87] & 9.77   & < .001 \\
Car3        & 0.24 & 0.14     & [0.06, 0.63] & -2.46  & 0.014  \\
Ang05L      & 0.96 & 0.40     & [0.37, 2.01] & -0.10  & 0.923  \\
Ang05R      & 1.28 & 0.47     & [0.57, 2.48] & 0.67   & 0.500  \\
Ang10L      & 0.48 & 0.28     & [0.12, 1.28] & -1.24  & 0.214  \\
Ang10R      & 4.14 & 0.94     & [2.60, 6.34] & 6.28   & < .001 \\
Ang20L      & 0.48 & 0.28     & [0.12, 1.28] & -1.24  & 0.214  \\
Ang20R      & 5.27 & 1.10     & [3.45, 7.83] & 7.96   & < .001 \\
Ang25L      & 0.81 & 0.37     & [0.29, 1.80] & -0.45  & 0.650  \\
Ang25R      & 7.23 & 1.36     & [4.96, 10.36]& 10.56  & < .001 \\
Blind40L    & 3.43 & 0.83     & [2.08, 5.40] & 5.08   & < .001 \\
Blind50L    & 6.12 & 1.21     & [4.10, 8.92] & 9.17   & < .001 \\
Blind60L    & 5.28 & 1.10     & [3.45, 7.85] & 7.97   & < .001 \\
Lat20       & 0.65 & 0.33     & [0.20, 1.55] & -0.85  & 0.396  \\
Lat40       & 3.00 & 0.77     & [1.75, 4.85] & 4.25   & < .001 \\
\cmidrule{1-6}
\end{tabular}
\smallskip
\scriptsize
\begin{tablenotes}
\item[]Note: SE: Standard Error, CI: Confidence Interval
\end{tablenotes}
\end{threeparttable}
\end{table}

    % FROM APPENDIX SUPPLEMENTARY STATISTIC TABLES - MINIMUM DISTANCE
    \setlength{\tabcolsep}{3pt}
% \begin{table}[h]
\begin{table}[H]
\caption{Linear model results for predictor Hallucination Configuration (Hypothesis $H_{4.2}$)}
\label{tab:regression_hallucination_configuration}
\centering
\scriptsize
\begin{threeparttable}
\begin{tabular}{lccccc}
\cmidrule{1-6}
Parameter & Coefficient & SE & 95\% CI & \chg{t(18338)} & p \\
\cmidrule{1-6}
% (Intercept) & 8.54 & 0.01 & [8.51, 8.57] & 624.31 & < .001 \\
% Car1 & -1.35 & 0.04 & [-1.43, -1.26] & -30.33 & < .001 \\
% Car2 & -1.60 & 0.04 & [-1.68, -1.51] & -35.99 & < .001 \\
% Car3 & -1.35 & 0.04 & [-1.44, -1.27] & -30.61 & < .001 \\
% Ang05L & -1.59 & 0.06 & [-1.71, -1.47] & -26.09 & < .001 \\
% Ang05R & -1.48 & 0.06 & [-1.60, -1.36] & -24.20 & < .001 \\
% Ang10L & -1.55 & 0.06 & [-1.67, -1.43] & -25.20 & < .001 \\
% Ang10R & -1.71 & 0.06 & [-1.83, -1.59] & -28.07 & < .001 \\
% Ang20L & -1.58 & 0.06 & [-1.70, -1.46] & -25.70 & < .001 \\
% Ang20R & -1.76 & 0.06 & [-1.88, -1.64] & -28.80 & < .001 \\
% Ang25L & -1.61 & 0.06 & [-1.73, -1.49] & -26.16 & < .001 \\
% Ang25R & -1.81 & 0.06 & [-1.93, -1.69] & -29.66 & < .001 \\
% Blind40L & -1.75 & 0.06 & [-1.87, -1.63] & -28.79 & < .001 \\
% Blind50L & -1.80 & 0.06 & [-1.92, -1.68] & -29.59 & < .001 \\
% Blind60L & -1.75 & 0.06 & [-1.87, -1.63] & -28.57 & < .001 \\
% Lat20 & -0.87 & 0.06 & [-0.99, -0.75] & -14.11 & < .001 \\
% Lat40 & -1.19 & 0.06 & [-1.31, -1.07] & -19.33 & < .001 \\

(Intercept) & 8.62 & 0.01 & [8.59, 8.65] & 623.98 & < .001 \\
Location    & -1.52 & 0.06 & [-1.64, -1.41] & -25.27 & < .001 \\
Car1        & -1.43 & 0.04 & [-1.51, -1.34] & -32.60 & < .001 \\
Car2        & -1.68 & 0.04 & [-1.76, -1.59] & -38.34 & < .001 \\
Car3        & -1.43 & 0.04 & [-1.52, -1.35] & -32.89 & < .001 \\
Ang05L      & -1.67 & 0.06 & [-1.79, -1.56] & -27.84 & < .001 \\
Ang05R      & -1.56 & 0.06 & [-1.67, -1.44] & -25.92 & < .001 \\
Ang10L      & -1.63 & 0.06 & [-1.74, -1.51] & -26.93 & < .001 \\
Ang10R      & -1.79 & 0.06 & [-1.91, -1.67] & -29.84 & < .001 \\
Ang20L      & -1.66 & 0.06 & [-1.78, -1.54] & -27.43 & < .001 \\
Ang20R      & -1.84 & 0.06 & [-1.96, -1.73] & -30.59 & < .001 \\
Ang25L      & -1.69 & 0.06 & [-1.81, -1.57] & -27.89 & < .001 \\
Ang25R      & -1.89 & 0.06 & [-2.00, -1.77] & -31.46 & < .001 \\
Blind40L    & -1.83 & 0.06 & [-1.95, -1.71] & -30.59 & < .001 \\
Blind50L    & -1.88 & 0.06 & [-2.00, -1.77] & -31.39 & < .001 \\
Blind60L    & -1.83 & 0.06 & [-1.95, -1.71] & -30.34 & < .001 \\
Lat20       & -0.95 & 0.06 & [-1.07, -0.83] & -15.65 & < .001 \\
Lat40       & -1.27 & 0.06 & [-1.39, -1.15] & -20.96 & < .001 \\
\cmidrule{1-6}
\end{tabular}
\smallskip
\scriptsize
\begin{tablenotes}
\item[]Note: SE: Standard Error, CI: Confidence Interval, $t(18338)$ = t-statistic with 18338 degrees of freedom
\end{tablenotes}
\end{threeparttable}
\end{table}

    A segmented analysis identified the most damaging configurations within each hallucination type in terms of system safety. Angular drifts to the right produced higher \chg{odds of an accident} than angular drifts to the left (chart $H_{4.1}$ in Figure~\ref{fig:hypothesis_summary} and \chg{Table~\ref{tab:odds_ratio_hallucination_config})}. As shown in Figure~\ref{fig:hypothesis_summary} ($H_{4.1}$), larger rightward angular drifts were associated with higher accident likelihood. They also reduced the minimum distance between the AV and the \chg{nearest} vehicle during the crossing (Figure \ref{fig:hypothesis_summary}, $H_{4.2}$). In other words, the greater the rightward angular drift, the smaller the minimum distance.
    
    On the other hand, for leftward angular drift, no direct relationship between drift magnitude and accident likelihood could be identified. \chg{In fact, none of the leftward configurations significantly changed the odds of an accident relative to the baseline ($OR$ between $0.48$ and $0.96$, all $p > 0.2$). The largest estimate was observed for ANG05L ($OR = 0.96, p = 0.923$), followed by ANG25L ($OR = 0.81, p = 0.650$).} A similar pattern was observed for minimum distance, as shown in Table \ref{tab:regression_hallucination_configuration}. The largest leftward drift (25\textdegree) caused the highest minimum distance reduction \chg{($1.69$\,m,} $p < 0.001$), followed by the smallest leftward drift (5\textdegree; \chg{$1.67$\,m,} $p < 0.001$), while \chg{20\textdegree}\ \chg{($1.66$\,m,} $p < 0.001$) and \chg{10\textdegree}\ \chg{($1.63$\,m,} $p < 0.001$) produced the smallest reductions. These results appear to be related to characteristics of the simulated use case, such as the direction of traffic flow on the transversal street, vehicle speed, and other road-layout features. A similar effect, likely also linked to the characteristics of the use case, was observed for \textit{Blind Region}. \chg{Blind strips with widths of} 40\textdegree, 50\textdegree, and 60\textdegree\ increased the odds of an accident by $3.43$ ($p < 0.001$), $6.12$ ($p < 0.001$), and $5.28$ ($p < 0.001$), respectively, and reduced the minimum distance by \chg{$1.83$\,m} ($p < 0.001$), \chg{$1.88$\,m} ($p < 0.001$), and \chg{$1.83$\,m} ($p < 0.001$), respectively. Thus, no direct relationship could be established between blind-strip \chg{width} and the safety metrics.    
            
    Furthermore, changing which vehicle was targeted by the \textit{Missed Detection} and \textit{Phantom Perception} hallucinations on the transversal street also \chg{significantly affected the odds of an accident} and reduced minimum distance. Targeting the first (car 1) \chg{and} second (car 2) vehicle increased the odds of an accident by $6.09$ ($p < 0.001$) \chg{and} $5.00$ ($p < 0.001$), respectively, \chg{whereas targeting the third vehicle (car 3) reduced them to $0.24$ times the baseline odds ($p = 0.014$)} (Table \ref{tab:odds_ratio_hallucination_config}). \chg{All three targets reduced the minimum distance, by} \chg{$1.43$\,m} ($p < 0.001$), \chg{$1.68$\,m} ($p < 0.001$), and \chg{$1.43$\,m} ($p < 0.001$) \chg{for car 1, car 2, and car 3}, respectively (Table \ref{tab:regression_hallucination_configuration}). This result is also likely related to the characteristics of the simulated use case, since changing the initial speed or position of the vehicles would probably affect these findings.

    When the hallucination delayed the MP information reaching MC, the results followed intuition more closely for the larger delay. A delay of $20$ simulation cycles did not significantly affect \chg{the odds of an accident} ($OR = 0.65$, $p = 0.396$), whereas a delay of $40$ simulation cycles increased the odds of an accident to $3.00$ times the baseline odds ($p < 0.001$) (Table~\ref{tab:odds_ratio_hallucination_config}). Both latency configurations significantly reduced the minimum distance between the AV and the \chg{nearest} vehicle, by $0.95$\,m for $20$ cycles and $1.27$\,m for $40$ cycles ($p < 0.001$ for both; Table~\ref{tab:regression_hallucination_configuration}). Although the odds-ratio estimate for $40$ cycles was higher than for $20$ cycles, this comparison should be interpreted descriptively because the model coefficients are reported relative to the baseline condition. Therefore, $H_4$ was accepted at the factor level, since hallucination configuration significantly affected both accident likelihood ($H_{4.1}$) and minimum distance ($H_{4.2}$) overall. However, the individual configuration coefficients show that these effects were not uniform across all configurations.

    \textbf{The effect of hallucination probability on system safety} (\bm{$H_5$}) (Tables \ref{tab:anova_hi_accident_probability} and \ref{tab:anova_hi_minimum_distance}, row 5, and Tables \ref{tab:odds_ratio_hallucination_prob} and \ref{tab:regression_hallucination_probability}). Hallucination probability had a significant \chg{overall} effect on \chg{both} accident likelihood (\chg{LR} $\chi^2 = 385.43, p < 0.001$, Table \ref{tab:anova_hi_accident_probability}) and minimum distance \chg{($F(5, 18350) = 1{,}409$}, $p<.001$, partial $\eta^2=0.28$, Table \ref{tab:anova_hi_minimum_distance}). \chg{Higher hallucination probabilities were associated with progressively higher odds of an accident (Figure~\ref{fig:hypothesis_summary}, $H_{5.1}$, and Table~\ref{tab:odds_ratio_hallucination_prob}). The $1\%$ and $5\%$ probabilities did not significantly change the odds relative to the baseline ($OR = 0.64$, $p = 0.097$, and $OR = 1.00$, $p = 0.985$), whereas the $50\%$ probability increased them to $8.53$ times the baseline odds ($p < 0.001$). Thus, $H_{5.1}$ was accepted. In contrast, the minimum-distance coefficients did not show the monotonic decrease predicted by $H_{5.2}$. The estimated reductions relative to the baseline were $1.62$\,m, $1.57$\,m, $1.55$\,m, $1.52$\,m, and $1.77$\,m for probabilities of $1\%$, $5\%$, $10\%$, $25\%$, and $50\%$, respectively (Figure~\ref{fig:hypothesis_summary}, $H_{5.2}$, and Table~\ref{tab:regression_hallucination_probability}). Therefore, although every tested probability significantly reduced the minimum distance, the hypothesized directional relationship was not observed, and $H_{5.2}$ was not supported. Since $H_{5.1}$ was accepted and $H_{5.2}$ was not supported, $H_5$ was partially accepted.}

    % FROM APPENDIX SUPPLEMENTARY STATISTIC TABLES - ODDS RATIO
    \setlength{\tabcolsep}{3pt}
% \begin{table}[h]
\begin{table}[H]
\caption{Odds ratio results for predictor Hallucination Probability (Hypothesis $H_{5.1}$)}
\label{tab:odds_ratio_hallucination_prob}
\centering
\scriptsize
\begin{threeparttable}
\begin{tabular}{lccccc}
\cmidrule{1-6}
Parameter & Odds Ratio & SE & 95\% CI & z & p \\
\cmidrule{1-6}
(Intercept) & 0.01 & $1.24 \times 10^{-3}$ & [0.01, 0.02] & -45.62 & < .001 \\
1\% & 0.64 & 0.17 & [0.36, 1.05] & -1.66 & 0.097 \\
5\% & 1.00 & 0.22 & [0.64, 1.53] & 0.02 & 0.985 \\
10\% & 2.36 & 0.39 & [1.70, 3.24] & 5.25 & < .001 \\
25\% & 3.47 & 0.51 & [2.59, 4.61] & 8.48 & < .001 \\
50\% & 8.53 & 1.04 & [6.73, 10.86] & 17.60 & < .001 \\
\cmidrule{1-6}
\end{tabular}
\smallskip
\scriptsize
\begin{tablenotes}
\item[]Note: SE: Standard Error, CI: Confidence Interval
\end{tablenotes}
\end{threeparttable}
\end{table}

    % FROM APPENDIX SUPPLEMENTARY STATISTIC TABLES - MINIMUM DISTANCE
    \setlength{\tabcolsep}{3pt}
% \begin{table}[h]
\begin{table}[H]
\caption{Linear model results for predictor Hallucination Probability (Hypothesis $H_{5.2}$)}
\label{tab:regression_hallucination_probability}
\centering
\scriptsize
\begin{threeparttable}
\begin{tabular}{lccccc}
\cmidrule{1-6}
Parameter & Coefficient & SE & 95\% CI & t(18350) & p \\
\cmidrule{1-6}
(Intercept) & 8.62 & 0.01 & [8.59, 8.65] & 619.77 & < .001 \\
1\% & -1.62 & 0.03 & [-1.68, -1.55] & -49.53 & < .001 \\
5\% & -1.57 & 0.03 & [-1.63, -1.51] & -48.13 & < .001 \\
10\% & -1.55 & 0.03 & [-1.62, -1.49] & -47.63 & < .001 \\
25\% & -1.52 & 0.03 & [-1.59, -1.46] & -46.74 & < .001 \\
50\% & -1.77 & 0.03 & [-1.83, -1.70] & -53.95 & < .001 \\
\cmidrule{1-6}
\end{tabular}
\smallskip
\scriptsize
\begin{tablenotes}
\item[]Note: SE: Standard Error, CI: Confidence Interval, $t(18350)$ = t-statistic with 18350 degrees of freedom
\end{tablenotes}
\end{threeparttable}
\end{table}

% \setlength{\tabcolsep}{2pt}
% \begin{table}[h]
% \caption{Regression analysis of hallucination probability over accident rate and minimum distance}
% \label{tab:regression_hallucination_probability}
% \centering

% \begin{threeparttable}
% % \small
% \scriptsize
% \begin{tabular}{clccccc}
% \cmidrule{1-7}
% Variable & Parameter & Coefficient & SE & 95\% CI [LL, UL] & $t(199)$ & $p$ \\
% % \hline
% \cmidrule{1-7}

% \multirow{2}{*}{\centering AR}& (Intercept) & $4.01e{-3}$ & $6.75e{-3}$ & [-0.01, 0.02] & 0.59 & 0.553 \\
% & HP & 0.19 & 0.03 & [ 0.13, 0.24] & 7.06 & $< .001$ \\
% % \hline
% \cmidrule{1-7}

% \multirow{2}{*}{\centering MD}& (Intercept) & 7.05 & 0.05 & [6.95, 7.14] & 144.34 & $< .001$ \\
% & HP & -0.33 & 0.19 & [-0.71, 0.05] & -1.73 & 0.086 \\

% \cmidrule{1-7}
% \end{tabular}
% \smallskip
% \scriptsize
% \begin{tablenotes}
%     \RaggedRight
%     % \item[]* Type: C = Categoric, N = Numeric, B = Binary
%     \item[]Note: AR: Accident Rate; MD: Minimum Distance; HP: Hallucination Prabability; SE = standard error; 95\% CI = 95\% confidence interval [lower limit, upper limit]; $t(197)$ = t-statistic with 197 degrees of freedom; $p$ = significance level for the hypothesis test that the coefficient equals zero
% \end{tablenotes}
% \end{threeparttable}
% \end{table}

    \textbf{The effect of hallucination persistence on system safety} (\bm{$H_6$}) (Tables \ref{tab:anova_hi_accident_probability} and \ref{tab:anova_hi_minimum_distance}, row 6, and Tables \ref{tab:odds_ratio_hallucination_persistence} and \ref{tab:regression_hallucination_persistence}). Hallucination persistence significantly affected system safety. It affected both accident likelihood (\chg{LR} $\chi^2 = 148.59, p < 0.001$, Table \ref{tab:anova_hi_accident_probability}) ($H_{6.1}$ accepted) and minimum distance \chg{($F(2, 18353) = 3{,}556$}, $p<.001$, partial $\eta^2=0.28$, Table \ref{tab:anova_hi_minimum_distance}) ($H_{6.2}$ accepted). While intermittent hallucinations increased the odds of an accident by $2.34$ times ($p<0.001$) compared to the baseline, permanent hallucinations increased them by $3.86$ times ($p<0.001$) (Table \ref{tab:odds_ratio_hallucination_persistence}). This finding was consistent with the effect of persistence on minimum distance (Table \ref{tab:regression_hallucination_persistence}), as permanent hallucinations caused a larger reduction in \chg{minimum distance} \chg{($1.73$\,m)} than intermittent hallucinations \chg{($1.48$\,m)}. Since both $H_{6.1}$ and $H_{6.2}$ were accepted, $H_6$ was accepted.

    % FROM APPENDIX SUPPLEMENTARY STATISTIC TABLES - ODDS RATIO
    \setlength{\tabcolsep}{3pt}
% \begin{table}[h]
\begin{table}[H]
\caption{Odds ratio results for predictor Hallucination Persistence (Hypothesis $H_{6.1}$)}
\label{tab:odds_ratio_hallucination_persistence}
\centering
\scriptsize
\begin{threeparttable}
\begin{tabular}{lccccc}
\cmidrule{1-6}
Parameter & Odds Ratio & SE & 95\% CI & z & p \\
\cmidrule{1-6}
(Intercept) & 0.01 & $1.24 \times 10^{-3}$ & [0.01, 0.02] & -45.62 & < .001 \\
Inter & 2.34 & 0.30 & [1.83, 3.01] & 6.67 & < .001 \\
Perm & 3.86 & 0.45 & [3.08, 4.87] & 11.59 & < .001 \\
\cmidrule{1-6}
\end{tabular}
\smallskip
\scriptsize
\begin{tablenotes}
\item[]Note: SE: Standard Error, CI: Confidence Interval
\end{tablenotes}
\end{threeparttable}
\end{table}

    % FROM APPENDIX SUPPLEMENTARY STATISTIC TABLES - MINIMUM DISTANCE
    \setlength{\tabcolsep}{3pt}
% \begin{table}[h]
\begin{table}[H]
\caption{Linear model results for predictor Hallucination Persistence (Hypothesis $H_{6.2}$)}
\label{tab:regression_hallucination_persistence}
\centering
\scriptsize
\begin{threeparttable}
\begin{tabular}{lccccc}
\cmidrule{1-6}
Parameter & Coefficient & SE & 95\% CI & t(18353) & p \\
\cmidrule{1-6}
(Intercept) & 8.62 & 0.01 & [8.59, 8.65] & 620.61 & < .001 \\
Inter & -1.48 & 0.02 & [-1.53, -1.43] & -63.63 & < .001 \\
Perm & -1.73 & 0.02 & [-1.77, -1.68] & -74.43 & < .001 \\
\cmidrule{1-6}
\end{tabular}
\smallskip
\scriptsize
\begin{tablenotes}
\item[]Note: SE: Standard Error, CI: Confidence Interval, $t(18353)$ = t-statistic with 18353 degrees of freedom
\end{tablenotes}
\end{threeparttable}
\end{table}

% \setlength{\tabcolsep}{2pt}
% \begin{table}[h]
% \caption{Regression analysis of hallucination persistence over accident rate and minimum distance}
% \label{tab:regression_hallucination_persistence}
% \centering

% \begin{threeparttable}
% % \small
% \scriptsize
% \begin{tabular}{clccccc}
% \cmidrule{1-7}
% Variable & Parameter & Coefficient & SE & 95\% CI [LL, UL] & $t(198)$ & $p$ \\
% % \hline
% \cmidrule{1-7}
% \multirow{3}{*}{\rotatebox[origin=c]{90}{\parbox{0.6cm}{\centering \tiny{Accident Rate}}}}& (Intercept) & $8.36e{-16}$ & 0.07 & [-0.15, 0.15] & $1.12e{-14}$ & $> .999$ \\
% & Inter & 0.03 & 0.07 & [-0.12, 0.18] & 0.38 & 0.706 \\
% & Perm & 0.05 & 0.07 & [-0.10, 0.20] & 0.64 & 0.524 \\
% % \hline
% \cmidrule{1-7}

% \multirow{3}{*}{\rotatebox[origin=c]{90}{\parbox{0.7cm}{\centering \tiny{Minimum Distance}}}}& (Intercept) & 7.80 & 0.48 & [ 6.86, 8.74] & 16.29 & $< .001$ \\
% & Inter   & -0.72 & 0.48 & [-1.67, 0.22] & -1.50 & 0.134 \\
% & Perm & -0.91 & 0.48 & [-1.86, 0.04] & -1.89 & 0.061 \\

% \cmidrule{1-7}
% \end{tabular}
% \smallskip
% \scriptsize
% \begin{tablenotes}
%     \RaggedRight
%     % \item[]* Type: C = Categoric, N = Numeric, B = Binary
%     \item[]Note: SE = standard error; 95\% CI = 95\% confidence interval [lower limit, upper limit]; $t(197)$ = t-statistic with 197 degrees of freedom; $p$ = significance level for the hypothesis test that the coefficient equals zero
% \end{tablenotes}
% \end{threeparttable}
% \end{table}

\section{Discussion}\label{sec:discussion}

    This study evaluated the impact of HI on AV safety. HI operates at the perception--control interface rather than within a specific sensor or ML module. By injecting observable distortions into perception outputs, it supports a safety analysis process that is less dependent on sensor technologies or perception architecture. In this sense, HI complements sensor-level FI by linking perception-output distortions to unsafe control behavior.

    The results show that hallucinations affect both \chg{collision odds and minimum distance}. These metrics capture different failure manifestations. Collision odds reflect discrete failures, while minimum distance captures the continuous erosion of safety margins. Table \ref{tab:comparative_ranking} ranks hallucination types by Odds Ratio (OR) and by the linear effect on minimum distance ($\beta$). The rank difference, $\Delta$ Rank ($Rank_{MinDist} - Rank_{Collision}$), shows that some hallucinations primarily increase \chg{collision odds}, while others primarily reduce minimum distances. \textit{Missed Detection} is collision-dominant, with $OR = 5.20$, but it ranks $5^{th}$ in minimum distance degradation with \chg{$\beta = -1.42$\,m}. \textit{Perception Linear Drift} shows the opposite pattern. Its effect on collision \chg{odds} is the smallest, with $OR = 1.46$, but its \chg{reduction in minimum distance} is significant, with \chg{$\beta = -1.52$\,m}, placing it $4^{th}$ in minimum distance reduction. This is why evaluations based only on collisions may fail to detect latent degradation that can become critical under tighter interactions.

    % \begin{table*}[ht]
% \centering
% \caption{Comparative Ranking of Hallucination Effects (Collision Odds vs. Minimum Distance)}
% \label{tab:comparative_ranking}
% \begin{tabular}{lcccccccc}
% \toprule
% Hallucination & Rank & OR & Sig & Rank & $\beta$ & Sig & $\Delta$ & Risk \\
% Type & (Coll.) &  & Coll. & (MinDist) & (m) & Dist. & Rank & Pattern \\
% \midrule
% Missed Detection & 1 & 5.20 & Yes & 5 & -1.42 & Yes & +4 & Collision-heavy \\
% Blind Region & 2 & 4.92 & Yes & 1 & -1.85 & Yes & -1 & Dual-impact (severe) \\
% Angular Drift & 3 & 2.52 & Yes & 2 & -1.72 & Yes & -1 & Dual-impact \\
% Phantom & 4 & 2.23 & Yes & 3 & -1.60 & Yes & -1 & Dual-impact \\
% Latency & 5 & 1.81 & Yes & 6 & -1.11 & Yes & +1 & Collision-leaning \\
% Linear Drift & 6 & 1.46 & No & 4 & -1.52 & Yes & -2 & Near-miss dominant \\
% \bottomrule
% \end{tabular}
% \end{table*}

\setlength{\tabcolsep}{2pt}
\begin{table}[H]
\caption{Comparative ranking of hallucination effects across collision odds and minimum distance}
\label{tab:comparative_ranking}
\centering
\scriptsize
% \begin{threeparttable}
\begin{tabular}{lcccccc}
\cmidrule{1-7}
Type & $Coll_{Rank}$ & OR & $Dist_{Rank}$ & $\beta$ (m) & $\Delta_{Rank}$ & Profile \\
\cmidrule{1-7}
Missed Detection & 1 & 5.20 & 5 & -1.42 & +4 & \chg{CD} \\
Blind Region & 2 & 4.92 & 1 & -1.85 & -1 & DI \\
Angular Drift & 3 & 2.52 & 2 & -1.72 & -1 & DI \\
Phantom & 4 & 2.23 & 3 & -1.60 & -1 & DD \\
Latency & 5 & 1.81 & 6 & -1.11 & +1 & LB \\
Linear Drift & 6 & 1.46 & 4 & -1.52 & -2 & LB \\
\cmidrule{1-7}
\end{tabular}
% \smallskip
% \scriptsize
% \begin{tablenotes}
% \item[]Note. $Coll_{Rank}$ ranks collision odds ratios. $Dist_{Rank}$ ranks $|\beta|$ for minimum distance. $\Delta_{Rank}$ equals $Dist_{Rank} - Coll_{Rank}$. Profile codes follow the quadrant map using median thresholds on OR and $|\beta|$. DI is dual impact. CD is collision dominant. MD is margin dominant. MI is moderate impact.
% \end{tablenotes}

\vspace{2pt}

% --- notas com largura da coluna (IEEE), não da tabela ---
\begin{minipage}{\columnwidth}
\scriptsize
\raggedright
\setlength{\parindent}{0pt}
Note. $Coll_{Rank}$ ranks collision odds ratios. $Dist_{Rank}$ ranks $|\beta|$ for minimum distance. $\Delta_{Rank}$ equals $Dist_{Rank} - Coll_{Rank}$. Profile codes follow the quadrant map using median thresholds on OR and $|\beta|$. \chg{DI: Dual Impact. CD: Collision Dominant. DD: Distance Dominant. LB: Low on Both.}\\
\end{minipage}

% \end{threeparttable}
\end{table}

\begin{figure}[h]
        \centering
        \chgfig{\includegraphics[scale=0.08]{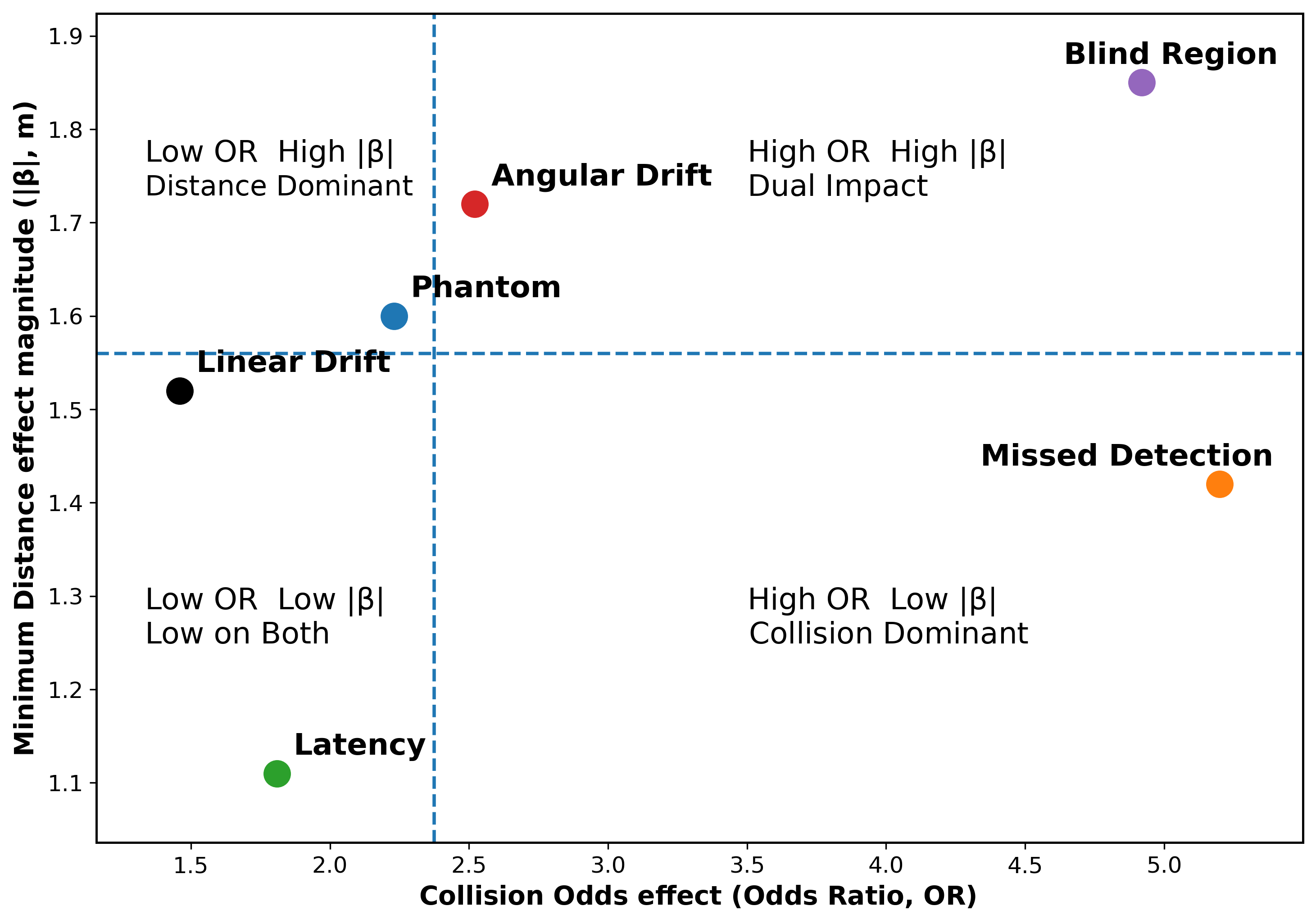}}
        \caption{Hallucination type impact on collision odds and minimum distance degradation}
        \label{fig:quad_type}
    \end{figure}

    To support multi-dimensional visual analysis of hallucination effects, Figure~\ref{fig:quad_type} presents a $2\times2$ quadrant map that positions each hallucination type according to its impact on OR and safety-margin degradation ($|\beta|$ for minimum distance). This visualization is intended as an interpretative summary of independently estimated regression effects rather than as an additional inferential model. Its purpose is to compare asymmetries in impact magnitude across the two safety metrics, not to infer a statistical relationship between them. \chg{Collision occurrence} and minimum distance are conceptually related, since collisions represent extreme cases of safety-margin erosion, but they were analyzed using separate statistical frameworks (logistic and linear regression, respectively). The quadrant map does not assume statistical independence between these metrics and does not introduce new hypothesis testing. Instead, it provides a structured way to compare effect magnitudes across complementary safety indicators. This improves interpretative transparency by making explicit whether a hallucination mechanism primarily increases \chg{collision odds}, \chg{reduces the minimum distance}, or affects both dimensions simultaneously.

    Median-based thresholds were used to define the quadrants and distinguish high and low magnitudes along each axis, enabling a data-driven classification of risk profiles. The upper-right quadrant (high OR, high $|\beta|$) represents dual-impact hallucinations that simultaneously amplify collision \chg{odds} and erode minimum distance, reflecting structurally severe distortions of situational awareness. This is the quadrant in which \textit{Perception Angular Drift} and \textit{Blind Region} are positioned. The upper-left quadrant (low OR, high $|\beta|$) captures \chg{distance-dominant} mechanisms, which primarily \chg{reduce the minimum distance} without proportionally increasing discrete collision events. \textit{Phantom Perception} is positioned in this quadrant. The lower-left quadrant (low OR, low $|\beta|$) reflects comparatively moderate or scenario-constrained effects. This is where \textit{Perception Linear Drift} and \textit{Perception Latency} are positioned. Finally, the lower-right quadrant (high OR, low $|\beta|$) represents collision-dominant effects, in which abrupt failure events are amplified more than continuous safety degradation. \textit{Missed Detection} appears in this quadrant. This visual representation complements the ranking analysis by showing that hallucination mechanisms cluster into qualitatively distinct safety behaviors rather than forming a single continuum of “more or less dangerous”. By jointly visualizing acute and latent risk dimensions, the quadrant map reinforces the importance of multi-metric evaluation for robust AV safety assessment.

    The divergence observed for \textit{Perception Linear Drift} illustrates this distinction: although it does not significantly increase \chg{the odds of an accident} in the tested scenario, it consistently reduces \chg{the minimum distance}, suggesting latent risk accumulation that may become critical under different operational conditions. Because minimum distance reduction can be interpreted as latent risk, it is possible that the narrow experimental setup used in this study, including a single scenario with specific configurations, stressed the system sufficiently to increase latent risk (reduce minimum distance) under \textit{Perception Linear Drift}, but not enough to produce a statistically significant increase in collisions.

    %  These findings reinforce the importance of evaluating multiple complementary safety metrics. They also reinforce the need for future studies to test HI across a broader range of scenarios.

\begin{figure}[h]
            \centering
            \chgfig{\includegraphics[width=233pt]{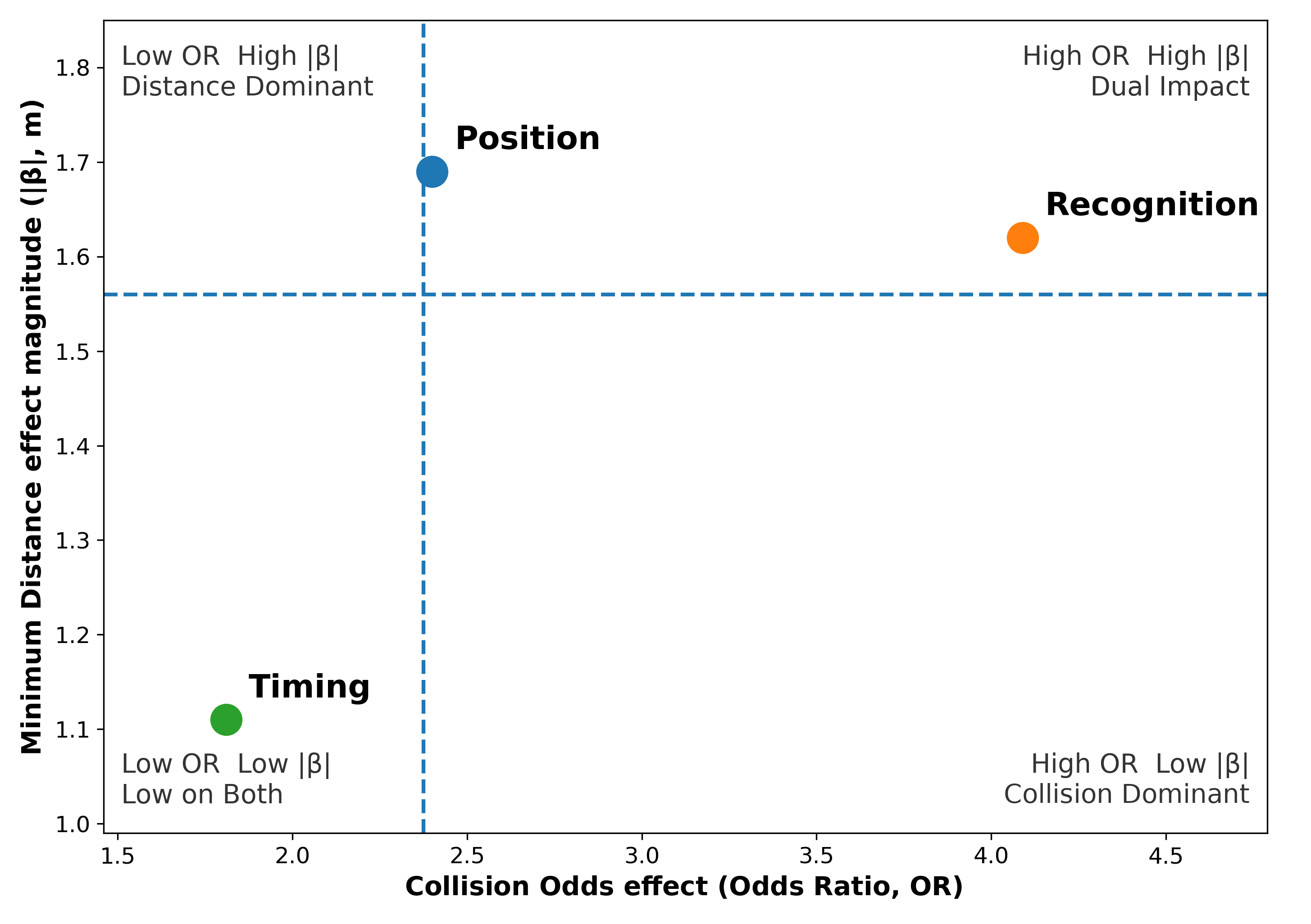}}
            \caption{Affected-domain impact on collision odds and minimum distance degradation}
            \label{fig:quad_affected}
        \end{figure}

    Analogously, Table~\ref{tab:affected_domain_comparison} was compiled to support the same type of analysis for Affected Domain, and the corresponding $2\times2$ quadrant plot in Figure~\ref{fig:quad_affected} provides a visual synthesis of these effects across safety dimensions. \chg{As observed in the figure, the Recognition and Position domains are positioned in the upper-right (dual-impact) quadrant, whereas Timing is positioned in the lower-left quadrant, with smaller but still statistically significant effects on both collision odds and minimum distance reduction. Recognition errors extend farther along the horizontal axis, reflecting a stronger influence on collision odds ($OR = 4.09$, compared with $2.40$ for Position). Position lies close to the median collision-odds threshold, and its minimum-distance reduction ($\beta = -1.69$) is similar to that of Recognition ($\beta = -1.62$). Therefore, at the domain level, recognition-related hallucinations are the main driver of collision odds, whereas position- and recognition-related hallucinations reduce the minimum distance to a similar extent.} Therefore, the quadrant distribution indicates that, unlike certain hallucination types that show metric-dependent divergence, domain-level disturbances manifest more uniformly across both acute failure events (collisions) and continuous \chg{reductions in minimum distance}. This visual and statistical alignment supports the interpretation that affected domains represent systemic vulnerability layers whose impact propagates consistently across multiple safety indicators.

    % VERSÃO 11/03/26
    % All domains show significant increases in collision odds and significant reductions in minimum distance. Recognition has the largest collision odds effect with OR $3.68$. Position has the largest reduction in minimum distance with beta $-1.73$. Timing remains significant but smaller with OR $1.81$ and beta $-1.11$. Table \ref{tab:affected_domain_comparison} and Figure \ref{fig:quad_affected} summarize these effects.

    % \begin{table}[htbp]
% \caption{Comparative Effects of Affected Domain on Collision Odds and Minimum Distance}
% \centering
% \begin{tabular}{lccccccc}
% \toprule
% Affected Domain 
% & OR 
% & 95\% CI (OR) 
% & Sig Coll? 
% & $\beta$ (m) 
% & 95\% CI ($\beta$) 
% & Sig Dist? 
% & Risk Profile \\
% \midrule
% Position    
% & 3.02 
% & [2.40, 3.81] 
% & Yes 
% & -1.73 
% & [-1.78, -1.69] 
% & Yes 
% & Dual-impact (buffer-heavy) \\

% Recognition 
% & 3.68 
% & [2.85, 4.76] 
% & Yes 
% & -1.51 
% & [-1.57, -1.46] 
% & Yes 
% & Dual-impact (collision-leaning) \\

% Timing      
% & 1.81 
% & [1.11, 2.81] 
% & Yes 
% & -1.11 
% & [-1.19, -1.02] 
% & Yes 
% & Moderate dual-impact \\
% \bottomrule
% \end{tabular}
% \label{tab:affected_domain_comparison}
% \end{table}

\setlength{\tabcolsep}{3pt}
\begin{table}[H]
\caption{Affected domain effects across collision odds and minimum distance}
\label{tab:affected_domain_comparison}
\centering
% \scriptsize
% \begin{threeparttable}
\begin{tabular}{lccc}
\cmidrule{1-4}
Domain & OR & $\beta$ (m) & Profile \\
\cmidrule{1-4}
\chg{Position} & \chg{2.40} & \chg{-1.69} & \chg{DI} \\
\chg{Recognition} & \chg{4.09} & \chg{-1.62} & \chg{DI} \\
\chg{Timing} & \chg{1.81} & \chg{-1.11} & \chg{LB} \\
\cmidrule{1-4}
\end{tabular}

% \smallskip
% \scriptsize
% \begin{tablenotes}
% \item[]Note. The quadrant map uses OR on the horizontal axis and $|\beta|$ on the vertical axis. Profile codes match the quadrant map using median thresholds. DI means dual impact. LB means low on both.
% \end{tablenotes}
% \end{threeparttable}

\vspace{2pt}

% --- notas com largura da coluna (IEEE), não da tabela ---
\begin{minipage}{\columnwidth}
\scriptsize
\raggedright
\setlength{\parindent}{0pt}
Note. Profile codes match the quadrant map using median thresholds. \chg{DI: Dual Impact. LB: Low on Both.}\\
\end{minipage}

\end{table}

    Importantly, this finding also underscores the methodological value of employing multiple hallucination types when evaluating safety at the domain level. Each hallucination mechanism (e.g., drift, latency, missed detection) perturbs perception in structurally distinct ways and therefore stresses different aspects of Position, Recognition, or Timing processing. When combined, these heterogeneous perturbations exercise the underlying domain more comprehensively, revealing both collision-triggering vulnerabilities and margin-eroding tendencies. In other words, domain-level risk does not emerge from a single failure mode, but from the combined effect of diverse perceptual distortions. Consequently, evaluating multiple hallucination types approximates the compounded stress real-world perception failures may impose on each affected domain.
        
    Overall, the results of this study demonstrate that \chg{the} HI module and all of its properties significantly affected the AV system safety by stressing the system in the experimental use case. By abstracting perception failures as high-level hallucinations, the HI framework provides a component-agnostic method for revealing safety issues and quantifying their consequences. The acceptance of all hypotheses \chg{except $H_{5.2}$} supports the view that HI can serve both as an effective perturbation mechanism and as a statistically grounded tool for AV safety analysis. The experiments showed that HI, as a general condition, more than tripled the \chg{odds} of an accident ($OR = 3.09$) and substantially degraded operational safety by reducing the mean minimum distance by \chg{$1.60$\,m.} This establishes a clear causal link between perception-level hallucinations and system-level risk.

    Moreover, the HI module provides a flexible framework for exploring new potential \chg{threats} in AV systems, even before research on the fault modes of new sensors or fault mechanisms has been consolidated. In practice, a researcher only needs to define and implement new hallucination types or configurations and execute simulations to evaluate how critical they are to AV safety. This can help prioritize research on the most critical mechanisms and guide the development of the most important protection mechanisms, potentially accelerating the AV development cycle.    
    
    The results indicate that not all hallucinations contribute equally to safety degradation. Failures causing complete information loss produced the most severe impacts. Specifically, \textit{Missed Detection} and \textit{Blind Region} increased accident odds by approximately five times. This result was corroborated by the minimum distance analysis, in which \textit{Blind Region} \chg{caused the largest reduction in minimum distance, by} an average of \chg{$1.85$\,m.} This is consistent with previous evidence showing that physical degradation of LiDARs or cameras can significantly increase accident risk \cite{10.1145/3728910}. Although less severe, hallucinations that distort information without removing it still represent a meaningful safety threat. For example, phantom objects more than doubled the odds of an accident, confirming that these hallucinations should not be ignored. This distinction is important because it reinforces the need to prioritize the most impactful hallucinations in both design-time validation and runtime monitoring.

    \chg{Reported real-world incidents are presented here only as context and do not validate the simulation-based ranking of hallucination risks observed in this study.} \chg{Missed-detection failures have been reported across multiple AV operators.} Reported cases include Waymo’s failures to detect low-profile obstacles such as chains, sidewalks, and poles \cite{Shepardson2025Waymo, NHTSA_SGO}, Cruise’s collisions with stationary or partially occluded pedestrians \cite{BrownTheDrive, StoneGovtech}, Zoox’s intermittent tracking errors that resulted in recalls \cite{ZooxRecall2025, Shepardson2025ZooxRecall}, and Tesla's diverse accidents \cite{BeeneBloomberg2025, BBCNewsTeslaCrash2018}. \chg{Phantom-object events have also been reported, particularly} in relation to Tesla’s camera-based driver-assistance system, where phantom-braking events have contributed to several incidents \cite{InceBusinessInsider2023, BBCNewsTeslaRecall2024}. \chg{These reports suggest that the hallucination types observed in practice may depend} in part on the underlying MP architecture \cite{tang2023multi}. \chg{In the simulated scenario, \textit{Missed Detection} and \textit{Blind Region} showed the highest odds ratios. The reported incidents show that missed detections also occur in deployed systems, but they are not exposure-normalized and therefore cannot be compared quantitatively with these simulation-based estimates. These observations suggest that} AV safety assessments should consider both the likelihood and severity of each hallucination type when guiding the development of mitigation strategies.

    Another important finding is that the affected perception domain matters. Hallucinations that affect object recognition and object position increased accident odds more than \chg{timing-related hallucinations, and recognition-related hallucinations showed the largest increase ($OR = 4.09$)}. Their effect on minimum distance was also severe, \chg{with position- and recognition-related hallucinations causing similar average reductions in minimum distance ($1.69$\,m and $1.62$\,m, respectively)}. This observation has direct implications for the design of AI-based perception models. Achieving high accuracy in object localization and classification should remain a priority, \chg{as also illustrated by} recent safety incidents related to perception failures \cite{carter2025Tesla, master2025xiaomi}. Architectural choices, algorithms, training datasets, and test and \chg{validation} procedures should reflect the disproportionate safety impact of these perceptual dimensions.

    Some operational configurations can create especially high risk in the use case evaluated here. One of the most dangerous conditions identified was an angular drift of \chg{25\textdegree}\ to the right, which increased accident odds by a factor of seven and was also among the most damaging configurations in terms of minimum-distance reduction ($1.89$\,m). This highlights how miscalibrations can produce disproportionate safety impacts and illustrates the value of the HI method for systematically identifying high-risk corner cases. This approach is consistent with other frameworks that use systematic perception-error injection for virtual safety validation \cite{piazzoni2023pem}. Furthermore, identifying these specific high-impact failures is a prerequisite for developing real-time safety monitors capable of assessing the danger of a given failure relative to the vehicle's current plan \cite{antonante2023task}.

    The analysis also showed that both the probability and persistence of hallucinations significantly modify their safety impact. This finding contributes to a growing body of research showing that perception failures cannot be treated as isolated events, because their safety implications depend strongly on their dynamic and temporal characteristics. A more holistic understanding of component health requires reasoning about diagnostic information as it evolves over time \cite{antonante2023monitoring}. The goal of a monitoring system is not merely to detect every error, but to identify task-relevant failures that pose genuine risk to the vehicle’s current plan, and to do so quickly enough to enable safe recovery \cite{chakraborty2025system}. Achieving this requires continuous real-time monitoring that can provide timely alerts and support rapid response \cite{hou2023fault}. Therefore, the present results highlight the need for continuous health-monitoring and predictive fault-detection systems that account for both the frequency and duration of failures rather than treating them as isolated events.

    The relationship between hallucination probability and accident risk was complex. \chg{At a very low fault rate ($1\%$), the estimated odds of an accident were below the baseline ($OR = 0.64$), but this difference was not statistically significant ($p = 0.097$), and a $5\%$ rate also did not show a significant difference. This pattern may indicate that the AV control reacts to minimal perceptual noise by adopting more cautious behavior, but it should be interpreted with caution.} However, any such protective effect disappeared as the fault rate increased. As hallucination probability increased, \chg{the odds of an accident grew substantially, reaching more than eight times the baseline odds at the $50\%$ probability level} (OR = $8.53$). Hallucination persistence was also directly related to safety impact. \chg{The Permanent persistence condition had a stronger safety impact than the Intermittent condition}, increasing accident odds more strongly (OR $=$ $3.86$ vs. $2.34$) and causing a greater reduction in minimum distance \chg{($1.73$\,m} vs. \chg{$1.48$\,m)}.

    These findings illustrate the value of an effect-centric strategy for AV safety. By focusing on observable system-level outcomes rather than on the internal mechanics of specific sensors or algorithms, the HI framework provides a reusable, sensor-agnostic method for safety analysis that is less dependent on specific AV architectures. The proposed framework, which incorporates five analysis dimensions (hallucination type, affected domain, configuration, probability, and persistence), offers a standardized taxonomy that can support \chg{simulation-based} testing across AV platforms.

    The implications extend beyond research. Regulators and standards bodies could use hallucination-based tests as part of safety assurance pipelines, complementing performance benchmarks with explicit evaluation of resilience to perception failures. This would move AV testing closer to the safety infrastructures established in aviation and other high-reliability industries. However, unlike aviation, where a relatively small number of actors operate under unified and enforceable regulatory frameworks, the AV ecosystem remains fragmented and competitive, often shaped by proprietary incentives \cite{shimanuki2025navigating, leon2022industry}. This fragmentation, reinforced by what has been termed the \chg{``AV-IP} \chg{problem''}, in which proprietary AV stacks limit independent safety evaluation \cite{tahir2021intersection}, has slowed the development of interoperable safety infrastructure and left systematic safety evaluation lagging behind the pace of AI innovation \cite{nascimento2018concerns, bengio2024managing}. The HI framework can help narrow this gap by offering a scalable and statistically validated toolkit that can support both industry practice and regulatory oversight.    
    
    The experimental validation in this study adopted two deliberate abstractions: (i) a \chg{ground-truth-based} MP layer and (ii) a simplified, interpretable MC policy based on gap acceptance. These choices enable controlled isolation of the causal link between perceptual distortions and unsafe control outcomes, but they do not capture the full stochasticity, uncertainty propagation, and error structure of modern perception pipelines or complex planning stacks.

    % Acho que isso endereça diretamente o ponto pedido pelo revisor 1:
    % "I’d still ask them to expand this section and be very explicit about what parts generalize (the abstraction/taxonomy) vs. what might not (single scenario, one simulator/control stack)."
    Therefore, the quantitative risk magnitudes reported in this study should be interpreted in relation to the evaluated scenario and implementation. In contrast, the main contributions expected to generalize are: (a) the hallucination taxonomy, (b) the abstraction of failures at the perception-output level, and (c) the injection paradigm as a reusable safety evaluation mechanism that is independent of sensor modality or ML architecture.

    % Future work will integrate the HI framework with simulated sensors and learned perception models to evaluate how perception uncertainty, detection confidence, occlusions, and temporal filtering interact with HI and propagate through more realistic AV stacks.

    %It is noteworthy that some of the results might be not generalizable to other configurations and use cases. Thus, additional investigations are needed. Changing the initial condition and other conditions in the specific scenario could be a first step to understand how generalizable those results are. Moreover, testing with various use cases systematically is also important to validate the generalization of the findings.

\section{Concluding Remarks} \label{sec:conclusions}

    This study presents a simulation-based HI framework for systematically evaluating how perceptual distortions (hallucinations) affect AV safety. HI complements sensor-level and model-level FI by providing a system-level abstraction of how perceptual failures affect AV behavior. By modeling six sensor-agnostic hallucination types and performing $18{,}356$ simulations in a high-risk intersection scenario, the analysis showed that all HI-module properties significantly influenced accident risk and minimum distances. The results quantify how different hallucination types and related properties affect safety-critical metrics, extending a previously developed framework.

    The findings show that some hallucinations pose disproportionately severe risks. These hallucinations can originate from both software components \chg{(e.g.,} machine learning (ML) perception models) and the hardware layer \chg{(e.g.,} sensors, GPUs), reinforcing the value of the framework's component-agnostic design. The results also provide useful information for AV developers, regulators, and safety engineers by identifying high-impact failure scenarios that require careful testing and monitoring.    
        
    %This study provides a foundational analysis in a controlled environment, a design choice that also defines the boundaries of the findings. The experiments were conducted in a single, unsignalized crossing with an experimental motion controller. 
    The current experimental validation adopts simplified perception and control abstractions to isolate causal safety effects. Thus, the present study evaluates interface-level hallucinations in the information delivered from perception to control, rather than sensor-realistic perception faults. Moreover, all the experiments were conducted in a single, unsignalized crossing scenario. Therefore, the quantitative risk magnitudes reported in this study, such as odds ratios and minimum-distance reductions, should be interpreted in relation to the evaluated scenario and implementation. However, the results demonstrate that the HI framework can systematically expose safety-relevant effects of distorted perception information. The proposed HI framework is independent of specific perception implementations and can be integrated in future work with realistic sensor models and learning-based perception pipelines.

    Future work should evaluate the HI framework across a broader range of driving scenarios, from complex urban intersections to highways, and across different AV architectures, prioritizing scenarios according to the risk and incident frequencies described in Appendix~\ref{app:sgo}. It should also include varied traffic situations, road geometries, traffic conditions, and more sophisticated AV control systems. In addition, future work should integrate HI with simulated sensors and learning-based perception models to evaluate how perception uncertainty, detection confidence, occlusions, and temporal filtering interact with hallucination effects in more realistic AV stacks. Future evaluations should also compare HI with related approaches, particularly PEM, considering the error types common to both frameworks while further exploring the additional hallucination types and combinations supported by HI. These studies should report the actual number of fault activations, particularly for low-probability permanent faults, and quantify the relationship between the number of injected or activated hallucinations and their effects on safety outcomes. Statistical comparisons across error types should also incorporate additional safety measures, such as time-to-collision (TTC) and unsuccessful-crossing rates.

    % Extending the investigation in this way will be necessary to build a broader map of safety vulnerabilities, support the design of more robust AV safety systems, and contribute to the development of validation protocols and future regulatory standards.

\appendices
\section{\break Exploratory Analysis of NHTSA SGO Crash Reports (2021--2025)}
\label{app:sgo}

To support scenario selection, the following analysis summarizes roadway-type frequencies in NHTSA SGO \cite{SGO_21_25} crash reports for ADS and ADAS vehicles from 2021 to 2025 under progressively stricter filters. These results are used only as contextual evidence. SGO reports reflect reported incidents and are not exposure-normalized (e.g., miles driven or ODD distribution), so they should not be interpreted as population-level risk estimates.

\subsection{Data preparation}
    This study computes \textit{Roadway Type} frequencies from NHTSA SGO crash reports (ADS and ADAS, 2021--2025) using the following procedure:

    \begin{enumerate}
        \item Merge ADS and ADAS records into a single table with a consistent schema.
        
        \item Remove records with empty or \textit{(blank)} values in \textit{Roadway Type}.
        
        \item Compute roadway-type counts and shares (Table~\ref{tab:sgo_table1}).
        
        \item Apply a \textit{CP Pre-Crash Movement} exclusion filter by removing records labeled as \textit{Backing}, \textit{Parked}, \textit{Parking Maneuver}, \textit{Stopped}, \textit{Unknown}, \textit{(blank)}, and \texttt{empty}. Then recompute roadway-type counts and shares (Table~\ref{tab:sgo_table2}).        
        
        \item Apply an additional \textit{Highest Injury Severity Alleged} exclusion filter by removing records labeled as \textit{No Injuries Reported}, \textit{Unknown}, \textit{(blank)}, and \texttt{empty}. Then recompute roadway-type counts and shares (Table~\ref{tab:sgo_table3}).
   
    \end{enumerate}

\subsection{Roadway-type summaries and filters}
    Tables~\ref{tab:sgo_table1} to \ref{tab:sgo_table3} report roadway-type counts and shares under progressively stricter filters. Without additional filtering beyond non-empty \emph{Roadway Type} (Table~\ref{tab:sgo_table1}), intersections account for 18.0\% of reports, while \emph{Unknown} remains substantial. Restricting the dataset to non-stationary pre-crash movements (Table~\ref{tab:sgo_table2}) reduces the share of \emph{Unknown} and increases the share of intersections to 34.8\%, close to streets (36.2\%). When the dataset is further restricted to reports alleging injuries (Table~\ref{tab:sgo_table3}), intersections become the most prevalent roadway type (34.8\%).

    \setlength{\tabcolsep}{5pt} % Ajuste o espaçamento entre as colunas conforme necessário
\begin{table}[H]
\caption{SGO (ADS+ADAS, 2021--2025): Roadway Type distribution after removing empty/blank \textit{Roadway Type} entries}
\label{tab:sgo_table1}
\centering

% --- tabela estreita (como você já tem) ---
\begin{tabular}{lcc}
\cmidrule{1-3}
\textbf{Roadway Type} & \textbf{Count} & \textbf{Share} \\
\cmidrule{1-3}
Highway / Freeway & 1977 & 31.3\% \\
Street            & 1561 & 24.7\% \\
Unknown           & 1366 & 21.6\% \\
\textbf{Intersection} & \textbf{1137} & \textbf{18.0\%} \\
Parking Lot       & 156  & 2.5\% \\
Rural Road        & 108  & 1.7\% \\
Traffic Circle    & 12   & 0.2\% \\
Unpaved Road      & 3    & 0.0\% \\
\cmidrule{1-3}
\end{tabular}

\vspace{2pt}

% --- notas com largura da coluna (IEEE), não da tabela ---
\begin{minipage}{\columnwidth}
\scriptsize
\raggedright
\setlength{\parindent}{0pt}
\textit{Notes:} Counts and shares are computed over reports with non-empty \textit{Roadway Type}.
\end{minipage}

\end{table}
    % -------------------------
% Table SGO-2
% -------------------------
\begin{table}[H]
\caption{SGO (ADS+ADAS, 2021--2025): Roadway Type distribution after filtering to non-stationary \textit{CP Pre-Crash Movement} categories}
\label{tab:sgo_table2}
\centering

% --- tabela estreita (como você já tem) ---
\begin{tabular}{lcc}
\cmidrule{1-3}
\textbf{Roadway Type} & \textbf{Count} & \textbf{Share} \\
\cmidrule{1-3}
Street            & 1013 & 36.2\% \\
\textbf{Intersection} & \textbf{972} & \textbf{34.8\%} \\
Highway / Freeway & 686  & 24.5\% \\
Parking Lot       & 68   & 2.4\% \\
Rural Road        & 31   & 1.1\% \\
Unknown           & 19   & 0.7\% \\
Traffic Circle    & 6    & 0.2\% \\
\cmidrule{1-3}
\end{tabular}

\vspace{2pt}

% --- notas com largura da coluna (não da tabela) ---
\begin{minipage}{\linewidth}
\scriptsize
\raggedright
\setlength{\parindent}{0pt}
\textit{Filter:} \textit{CP Pre-Crash Movement} excludes empty, \textit{Backing}, \textit{Parked}, \textit{Parking Maneuver}, \textit{Stopped}, \textit{Unknown}, and \textit{(blank)}.\\
\end{minipage}

\end{table}
    % -------------------------
% Table SGO-3
% -------------------------
\setlength{\tabcolsep}{5pt} % Ajuste o espaçamento entre as colunas conforme necessário
\begin{table}[H]
\caption{SGO (ADS+ADAS, 2021--2025): Roadway Type distribution for injury-alleging reports after movement filtering}
\label{tab:sgo_table3}
\centering

% --- tabela estreita (como você já tem) ---
\begin{tabular}{lcc}
\cmidrule{1-3}
\textbf{Roadway Type} & \textbf{Count} & \textbf{Share} \\
\cmidrule{1-3}
\textbf{Intersection} & \textbf{146} & \textbf{34.8\%} \\
Street            & 139 & 33.1\% \\
Highway / Freeway & 116 & 27.6\% \\
Rural Road        & 7   & 1.7\% \\
Unknown           & 7   & 1.7\% \\
Traffic Circle    & 4   & 1.0\% \\
Parking Lot       & 1   & 0.2\% \\
\cmidrule{1-3}
\end{tabular}

\vspace{2pt}

% --- notas com largura da coluna (IEEE), não da tabela ---
\begin{minipage}{\columnwidth}
\scriptsize
\raggedright
\setlength{\parindent}{0pt}
\textit{Filter:} \textit{Highest Injury Severity Alleged} excludes empty, \textit{No Injuries Reported}, \textit{Unknown}, and \textit{(blank)}.\\
\end{minipage}

\end{table}

    These trends support the use of an intersection crossing maneuver as a high-relevance testbed for system-level safety evaluation.

%%%%%%%%%%%%%%%%%%%%%

\section{How to Reproduce this Study} \label{app:repro_guideline}

\raggedbottom

    This appendix provides a concise reproduction checklist for the study. It summarizes the simulation environment, scenario configuration, HI settings, repetitions, run termination and validity criteria, recorded data, and statistical analysis. The steps below describe how to reproduce the study, and Table \ref{tab:reproducibility_checklist} summarizes the main settings and where they are defined in the paper.

\begin{table*}[t]
\centering
\caption{Reproduction checklist for the study}
\label{tab:reproducibility_checklist}
\scriptsize
\setlength{\tabcolsep}{4pt}
\begin{tabularx}{\textwidth}{@{}p{0.23\textwidth} X p{0.19\textwidth}@{}}
\toprule
\textbf{Item} & \textbf{Settings for reproduction} & \textbf{Where defined in the paper} \\
\midrule

Scenario configuration &
Unsignalized intersection crossing with five crossing vehicles that have the right of way. The AV starts about $250\,m$ from the intersection. The other vehicles start about $270\,m$ from the intersection and accelerate from rest up to $54\,km/h$. The AV identifies candidate windows (CWs) to cross. &
Section~\ref{sub-sec:experimental_use_case} and Figure~\ref{fig:fig-23} \\

\addlinespace
Simulation environment and integration &
USP54 ReTS integrating OpenDS and MATLAB. The traffic scenario is executed in OpenDS. MC and HI are executed in MATLAB. Data is exchanged between OpenDS and MATLAB through a TCP/IP socket interface. &
Section~\ref{sub-sec:simulation_environment} and Figure~\ref{fig:fig-11} \\

\addlinespace
Perception output used by HI &
Simulator ground truth filtered by the virtual FOV is used as the perception output used by HI. \chg{The sensor-model uncertainty and error module is enabled in all runs, including the baseline, which differs from HI-enabled runs only in that HI is disabled.} HI is applied at the perception--control interface. &
Section~\ref{sub-sec:machine_perception_module} and Figure~\ref{fig:fig-37} \\

\addlinespace
HI properties &
HI is defined by ModuleActivation, HallucinationType, AffectedDomain, HallucinationConfiguration, HallucinationProbability, and HallucinationPersistence. &
Section~\ref{sub-sec:hallucination_injection_module} and Table~\ref{tab:investigated_components} \\

\addlinespace
Hallucination types &
PerceptionLinearDrift, PhantomPerception, MissedDetection, PerceptionAngularDrift, BlindRegion, and PerceptionLatency. &
Section~\ref{sub-sec:hallucination_injection_module} and Table~\ref{tab:failuretypes_labels} \\

\addlinespace
Affected domains &
ObjectPosition, ObjectRecognition, and InformationTiming. &
Section~\ref{sub-sec:hallucination_injection_module} and Table~\ref{tab:dimensionaffected_labels} \\

\addlinespace
Hallucination configurations &
Location; Car1, Car2, Car3; Ang05L, Ang10L, Ang20L, Ang25L; Ang05R, Ang10R, Ang20R, Ang25R; Blind40L, Blind50L, Blind60L; Lat20, Lat40. \chg{Location replaces the distance $d_i$ of every detected crossing vehicle to the intersection by $d_i' = d_i(1+u)$, with $u \sim U(-1,1)$ drawn once per run and applied to all vehicles. One execution cycle lasts 14.5\,ms on average, so Lat20 and Lat40 correspond to approximately 0.29\,s and 0.58\,s.} &
Section~\ref{sub-sec:hallucination_injection_module} and Table~\ref{tab:failureconfiguration_labels} \\

\addlinespace
Probability and persistence &
Hallucination probabilities of 1\%, 5\%, 10\%, 25\%, and 50\%. Persistence categories: Baseline, Intermittent, and Permanent. &
Section~\ref{sub-sec:hallucination_injection_module} \\

\addlinespace
Conditions and repetitions &
201 conditions were defined: 1 baseline condition with HI disabled and 200 HI conditions. Approximately 50 valid runs were collected for each HI condition and approximately 9000 runs for the baseline condition. &
Section~\ref{sub-sec:experimental_procedures} \\

\addlinespace
Run termination and valid executions &
Each run ends with safe crossing, collision, or indefinite halt. Simulator crashes and runtime failures are treated as invalid executions. Invalid logs are discarded and the condition is rerun until the target number of valid runs is reached. &
Sections~\ref{sub-sec:experimental_use_case} and \ref{sub-sec:experimental_procedures} \\

\addlinespace
Randomness and seeds &
Randomness was generated internally by MATLAB's pseudo-random routines, but per-run seeds were not archived. &
Appendix~\ref{app:repro_guideline}, Step~6 \\

\addlinespace
Recorded data &
Simulation logs include time, AV position and speed, steering, throttle, brake, and the position and speed of the crossing vehicles. The consolidated dataset includes Accident, MinimalDistance, and the HI variables for each run. &
Section~\ref{sub-sec:experimental_procedures} and Tables~\ref{tab:information_collected} and \ref{tab:dataset_fields_summary} \\

\addlinespace
Statistical analysis &
Logistic regression for Accident, linear regression for Minimum Distance, ANOVA for overall factor significance, significance threshold $\alpha=0.05$, and implementation in RStudio. &
Section~\ref{sub-sec:analysis_procedures} \\

\addlinespace
Execution environment &
The experiments were executed on two Windows 10 laptops. Simulation runs and log processing were performed on a Windows 10 laptop with Intel Core i7, 8 GB RAM, and NVIDIA GeForce MX150 GPU. Statistical analyses in RStudio were performed on a second Windows 10 laptop. &
Section~\ref{sub-sec:experimental_hardware_infrastructure} \\

\bottomrule
\end{tabularx}
\end{table*}

    \begin{enumerate}
        \item \textbf{Set up the simulation environment.} Use the real-time component of the simulation framework (ReTS), which integrates MATLAB and OpenDS. Execute the traffic scenario in OpenDS. Execute the MC and HI modules in MATLAB. Exchange data between OpenDS and MATLAB through a TCP/IP socket interface.

        \item \textbf{Load the scenario configuration.} Use an unsignalized intersection crossing scenario such as the one shown in Figure \ref{fig:fig-23}. The AV starts approximately $250$\,m from the intersection. Five crossing vehicles start approximately $270$\,m from the intersection, accelerate from rest up to $54$\,km/h, and have the right of way. The AV identifies candidate windows (CWs) to decide when to cross.

        \item \textbf{Configure the perception output used by HI.} Use simulator ground truth from MP \chg{as the source of the} perception output stream used by MC. \chg{Keep the sensor-model uncertainty and error module enabled for all experimental runs. For baseline runs, disable only the HI module. For HI-enabled runs, retain the same sensor uncertainty and error process and additionally apply the configured hallucination at the perception--control interface.} \chg{For \textit{PerceptionLinearDrift}, apply a proportional longitudinal perturbation to all detected crossing vehicles. For each vehicle $i$, replace its true distance to the intersection $d_i$ by $d_i' = d_i(1+u)$, where $u \sim U(-1,1)$. Sample $u$ once per simulation run and apply the same sampled value to all detected crossing vehicles throughout that run.}

        \item \textbf{Define the HI settings and experimental conditions.} Create one baseline condition with HI disabled and HI-enabled conditions defined by combinations of hallucination type, configuration, probability, and persistence, as specified in Section \ref{sub-sec:hallucination_injection_module}. In this study, the full design is composed of 201 conditions: 1 baseline condition and 200 HI conditions.

        \item \textbf{Execute the runs using the validity criteria.} End each simulation when the AV (i) crosses safely, (ii) causes a collision, or (iii) halts indefinitely because no safe crossing opportunity is available. Treat simulator crashes and runtime failures as invalid executions. Discard those logs and repeat the condition until the target number of valid runs is reached.

        \item \textbf{Repeat runs and store condition metadata.} \chg{Execute approximately 50 valid runs for each HI condition and approximately $9{,}000$ runs for the baseline condition.} Store the run identifier and condition metadata for each valid execution. Randomness was generated internally by MATLAB's pseudo-random routines, but per-run seeds were not archived.

        \item \textbf{Build the dataset and run the statistical analysis.} Record the simulation log variables listed in Table \ref{tab:information_collected}. Consolidate the valid runs into a dataset in which each row corresponds to one run and includes the HI variables and the safety outcomes summarized in Table \ref{tab:dataset_fields_summary}. Import the dataset into RStudio and run the logistic regression, linear regression, and ANOVA procedures described in Section \ref{sub-sec:analysis_procedures}.

    \end{enumerate}

    % \subsubsection{General parameters checklist}

        % Table \ref{tab:reproducibility_checklist} presents the general parameters used throughout all experiments.
    \noindent

\bibliographystyle{IEEEtran}
\bibliography{references}

% \raggedbottom

\begin{IEEEbiography}[{\includegraphics[width=1in,height=1.25in,clip,keepaspectratio]{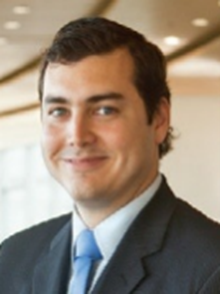}}]{Alexandre M. Nascimento} is an advisor and associate of the FRL (Future Realities Lab) at the University of Miami, and a researcher at Beyond the Hype, focusing on developing novel AI and safety approaches for safety-critical autonomous systems. He has conducted research at Stanford University and the University of Massachusetts. He holds four Ph.D. degrees with emphasis on AI: Computer Engineering (University of São Paulo), Computer Science (Federal University of São Paulo), Electronic Engineering and Informatics (Aeronautics Institute of Technology), and Business Administration (Fundação Getulio Vargas). He also holds two M.Sc. degrees (MIT and University of São Paulo).  Dr. Nascimento has authored more than 50 scientific papers, 100 media articles, 10 books, and 10 patents, and has received over a dozen awards. Previously, he served as International Product Engineering Director at Samsung Research America, leading the development of the SmartThings Internet-of-Things (IoT) platform in Silicon Valley.
\end{IEEEbiography}

\begin{IEEEbiography}[{\includegraphics[width=1in,height=1.25in,clip,keepaspectratio]{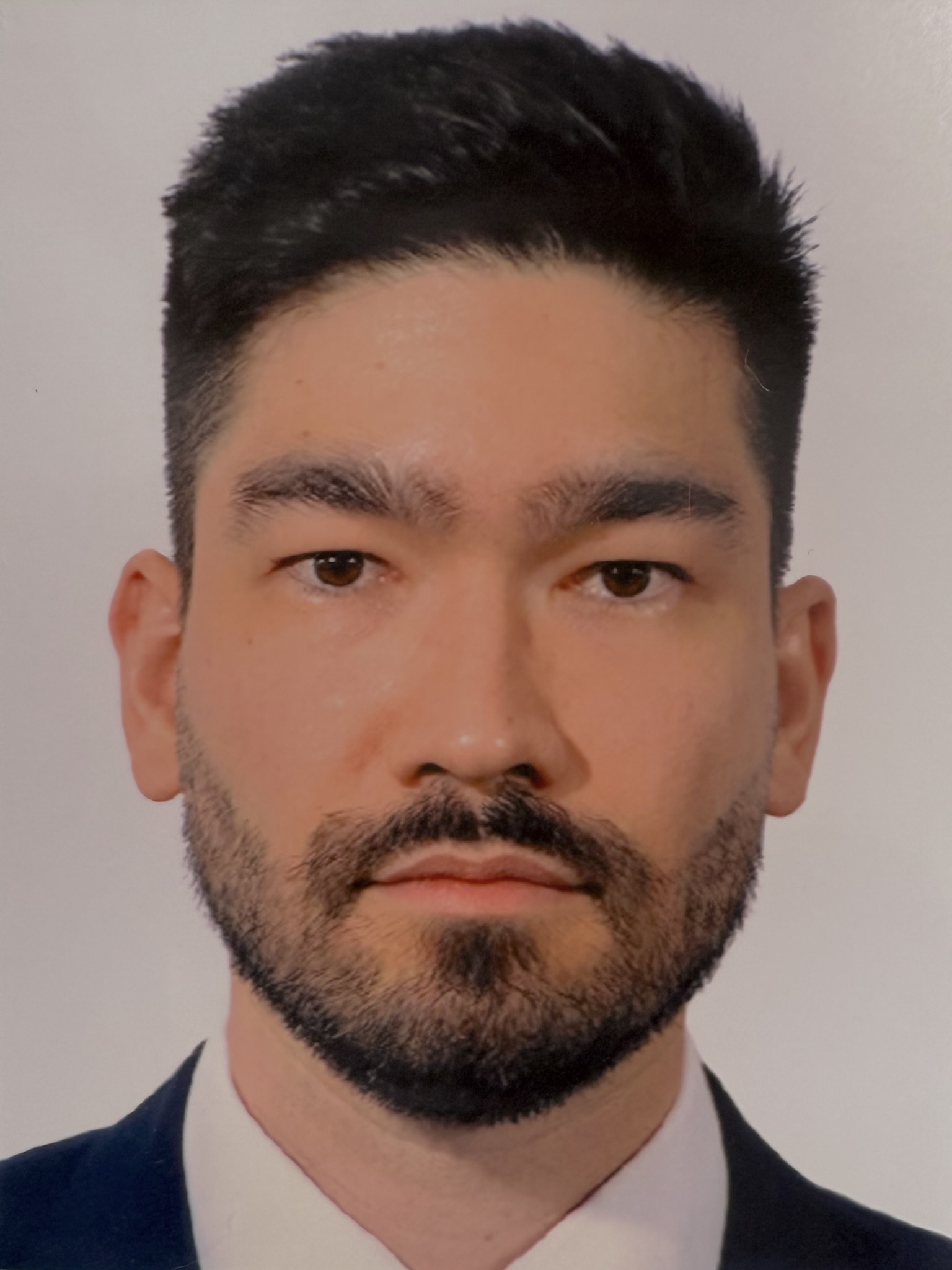}}]{Gabriel K. G. Shimanuki} received the B.Eng. degree (with honors) in Computer Engineering from the University of São Paulo (USP), Brazil, in 2023, where he later received the M.Eng. degree in Computer Engineering in 2026. Since 2020, he has been a member of the Safety Analysis Group at Poli-USP, focusing on autonomous vehicle safety, fault injection, and optimization for embedded systems. His research has led to several peer-reviewed publications, award-winning projects, and the development of simulation-based safety assessment frameworks. He is a recipient of multiple research fellowships and was nominated for a Top Senior Project at USP.
\end{IEEEbiography}

\begin{IEEEbiography}[{\includegraphics[width=1in,height=1.25in,clip,keepaspectratio]{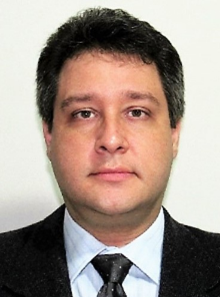}}]{Lúcio F. Vismari} received his B.Eng. (2001), M.Eng. (2007), and D.Sc. (2023) degrees in electrical engineering from the School of Engineering at the University of São Paulo (Poli-USP). Since 2002, he has been with the Safety Analysis Group (GAS) at Poli-USP as a senior researcher in the system safety field. His current research interests are the safety assurance challenges in complex engineered systems, mainly the application of Cyber-Physical Systems and new technological trends (such as AI, 5G communication and XR) in high-risk critical domains. He has more than 60 published scientific works in periodicals, conferences and books. Besides, he has more than 20 years of experience as an independent safety assessor, working in dozens of large-scale public and private projects in railway, subway, Air Traffic Management and Defense domains.\end{IEEEbiography}

\begin{IEEEbiography}[{\includegraphics[width=1in,height=1.25in,clip,keepaspectratio]{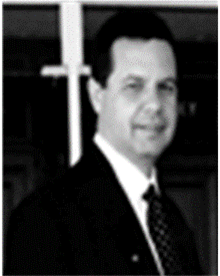}}]{João B. Camargo Jr.} was an IEEE Member from 2002 until 2014. He received his Bachelor’s Degree in electronic engineering and the M.Sc. and Ph.D. degrees from the School of Engineering of the University of São Paulo (Poli-USP), São Paulo, Brazil, in 1981, 1989, and 1996, respectively. He is currently an Associate Professor with the Department of Computer and Digital Systems Engineering (PCS) at Poli-USP, where he is also the coordinator of the Safety Analysis Group (GAS). Professor Camargo has 30 articles in Scientific Journals, 5 Organized Books, 6 Published Book Chapters, and 115 complete works published in proceedings of conferences. He is a reviewer for different Scientific Journals, such as IEEE Transactions on Intelligent Transportation Systems, Journal of Intelligent \& Robotic Systems, Risk Analysis, Journal of Advanced Transportation, Reliability Engineering \& System Safety, Transportation Research - Part C and IEEE Software. \end{IEEEbiography}

\begin{IEEEbiography}[{\includegraphics[width=1in,height=1.25in,clip,keepaspectratio]{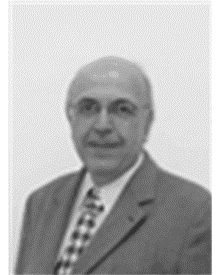}}]{Jorge R. de Almeida Jr.} is an Electronic Engineer (1981) from the School of Engineering at the University of São Paulo – Poli-USP (Brazil), with an MSc from Poli-USP (1989) and a PhD from Poli-USP (1995). He is a Full Professor in the Computer and Digital Systems Engineering Department at Poli-USP, University of São Paulo, Brazil; he is a member of the Safety Analysis Group at Poli-USP. His research is in the area of reliable and safe computational systems for critical applications. \end{IEEEbiography}

\begin{IEEEbiography}[{\includegraphics[width=1in,height=1.25in,clip,keepaspectratio]{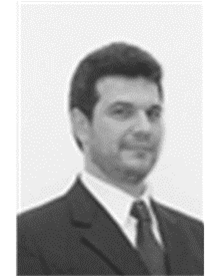}}]{Paulo S. Cugnasca} is an Electronic Engineer (1987) from the School of Engineering at the University of São Paulo – Poli-USP (Brazil), with an MSc from Poli-USP (1993) and a PhD from Poli-USP (1999). He is an Assistant Professor in the Computer Engineering and Digital Systems Department at Poli-USP, University of São Paulo, Brazil, and is a member of the Safety Analysis Group at Poli-USP. His research is in the area of reliable and safe computational systems for critical applications. \end{IEEEbiography}

\begin{IEEEbiography}[{\includegraphics[width=1in,height=1.25in,clip,keepaspectratio]{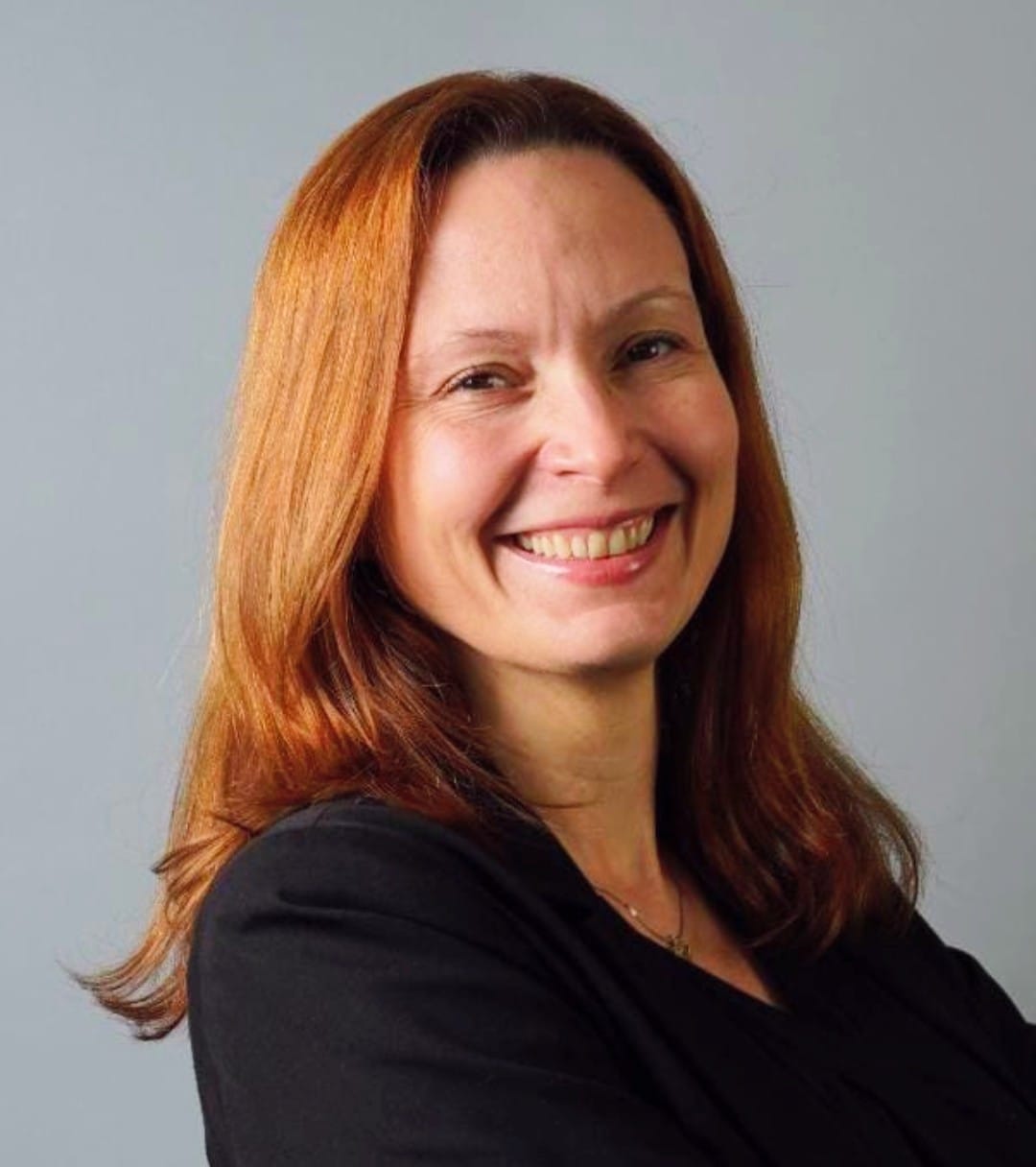}}]{Anna C. M. Queiroz} is an Associate Professor of Interactive Media at the University of Miami and a member of the Frost Institute for Data Science and Computing. She is the founding director of the Future Realities Research Lab at the University of Miami. Dr. Queiroz is an internationally recognized expert on the impacts of immersive media affordances. Her work combines lab, field, and large-scale studies to understand human behavior using immersive technologies. She has a postdoc in Communication and Education from Stanford University, a Ph.D., and a master's in Cognitive Psychology from the University of São Paulo (USP). Her project implementing virtual reality with 33,000 students was recently awarded by the King Center on Global Development at Stanford University. \end{IEEEbiography}

\begin{IEEEbiography}[{\includegraphics[width=1in,height=1.25in,clip,keepaspectratio]{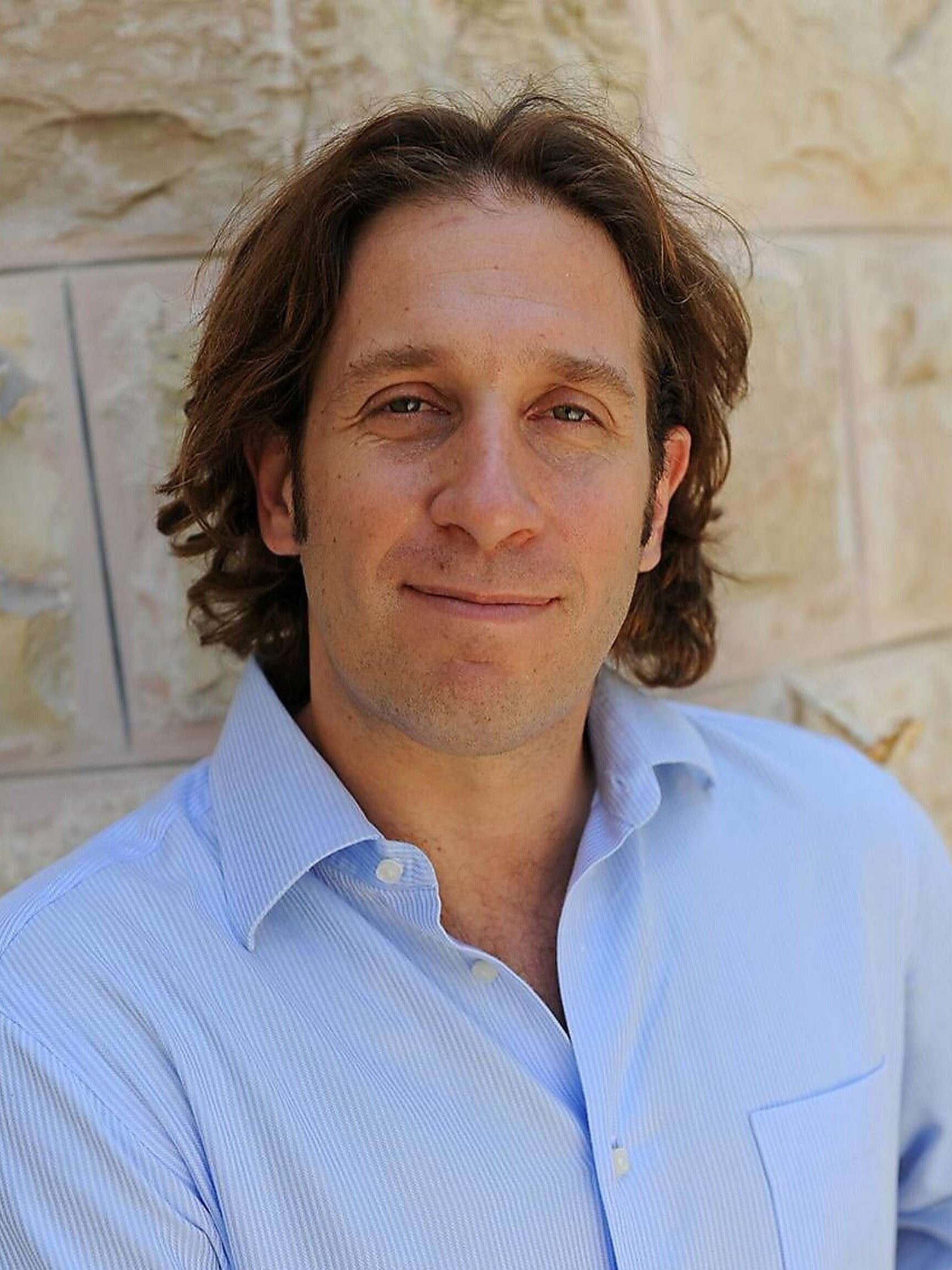}}]{Jeremy N. Bailenson} studies virtual reality's impacts on perception and social interaction. He directs Stanford's Virtual Human Interaction Lab and has published extensively on VR's applications in education, environment, empathy, and health. With more than 250 publications, his work has been awarded multiple times, and continuously funded by the National Science Foundation for over 25 years. He has produced or directed six Virtual Reality documentary experiences which were official selections at the Tribeca Film Festival. \end{IEEEbiography}

% \par

\EOD

% \raggedbottom

% \newpage

\end{document}